\documentclass[preprint,12pt]{elsarticle}

\usepackage[a4paper,
            bindingoffset=0.2in,
            left=0.6in,
            right=0.6in,
            top=0.6in,
            bottom=0.6in,
            footskip=.25in]{geometry}

\usepackage{graphicx}
\usepackage{amsmath}
\usepackage{amssymb}
\usepackage{float}
\usepackage{tikz}
\usepackage{placeins}

\usepackage{multicol}
\usepackage{longtable}

\usepackage{caption}
\usepackage{subcaption}

\usepackage[colorlinks=true, allcolors=blue]{hyperref}

\begin{document}

\title{Extremization to Fine Tune Physics Informed Neural Networks for Solving Boundary Value Problems}

\author[1,3]{Abhiram Anand Thiruthummal\corref{cor1}}
\ead{thiruthuma@uni.coventry.ac.uk}
\author[2,3]{Sergiy Shelyag}
\ead{sergiy.shelyag@flinders.edu.au}
\author[1]{Eun-jin Kim}
\ead{ejk92122@gmail.com}
\cortext[cor1]{Corresponding author}
\affiliation[1]{organization={Centre for Fluids and Complex Systems, Coventry University},
city={Coventry},
country={United Kingdom}}
\affiliation[2]{organization={College of Science and Engineering, Flinders University},
city={Adelaide},
country={Australia}}
\affiliation[3]{organization={School of Information Technology, Deakin University},
city={Melbourne},
country={Australia}}

\begin{abstract}
   We propose a novel method for fast and accurate training of physics-informed neural networks (PINNs) to find solutions to boundary value problems (BVPs) and initial boundary value problems (IBVPs). By combining the methods of training deep neural networks (DNNs) and Extreme Learning Machines (ELMs), we develop a model which has the expressivity of DNNs with the fine-tuning ability of ELMs. We showcase the superiority of our proposed method by solving several BVPs and IBVPs which include linear and non-linear ordinary differential equations (ODEs), partial differential equations (PDEs) and coupled PDEs. The examples we consider include a stiff coupled ODE system where traditional numerical methods fail, a 3+1D non-linear PDE, Kovasznay flow and Taylor-Green vortex solutions to incompressible Navier-Stokes equations and pure advection solution of 1+1 D compressible Euler equation.
   \\
   The Theory of Functional Connections (TFC) is used to exactly impose initial and boundary conditions (IBCs) of (I)BVPs on PINNs. We propose a modification to the TFC framework named Reduced TFC and show a significant improvement in the training and inference time of PINNs compared to IBCs imposed using TFC. Furthermore, Reduced TFC is shown to be able to generalize to more complex boundary geometries which is not possible with TFC.  We also introduce a method of applying boundary conditions at infinity for BVPs and numerically solve the pure advection in 1+1 D Euler equations using these boundary conditions.
\end{abstract}

\maketitle

\section{Introduction}

Differential equations are used to mathematically model various phenomena in the fields of engineering, physics, chemistry, economics and biology. A tiny fraction of such differential equations admit an analytic closed-form solution. Therefore the study of solutions of differential equations requires the use of a variety of computational methods. There exist several numerical methods to find the solutions to ordinary differential equations (ODEs) and partial differential equations (PDEs). Traditionally, such methods are based on discretization of geometric or frequency spatial and temporal domains. Also, some numerical approximation of the derivatives involved in the differential equations is used on the grid defined by the discretization. Recently, however, a novel family of numerical methods for ODE and PDE solution began to emerge based on utilisation of artificial neural networks. These methods exploit the universal function approximation capabilities of the neural networks and have several properties, unusual for numerical methods, such as differentiability of the provided solution.

Physics-informed neural networks (PINNs) utilise physical laws written in the form of differential equations to facilitate the learning process to find the network weights, which can then be combined with other information to achieve various objectives. One of the first uses of PINN can be traced to \cite{lagaris1998artificial} where the authors used neural networks along with unsupervised learning to solve boundary value problems (BVPs) of ODEs and PDEs. The term `physics informed neural network' was coined much later in \cite{raissi2017physics} where a semi-supervised learning paradigm was introduced to find solutions to PDEs in a data-driven manner from sparse and noisy experimental data. The idea of PINN was later extended for data-driven discovery of PDEs \cite{raissi2019physics}. For data-driven discovery of PDEs and PDE solutions, PINN outperforms traditional methods, which requires computing the solution of PDEs multiple times iteratively to estimate the parameters.

In the case of solving IBVPs PINNs have some advantages over traditional methods like Finite Difference, Finite Volume or Finite Element Methods (FD, FVM and FEM). A PINN uses neural networks as an analytic approximation of the solution whereas FD/FVM/FEM discretize the spatial domain into elements, within which the solution is approximated using finite differences instead of derivatives, averaging and interpolation over a control volume, or fitting an arbitrary function within a control volume and minimizing the error. This generally works satisfactorily, however the problems represented by hyperbolic PDEs are often numerically unstable and require some additional numerical stabilisation techniques, inevitably leading to unphysically large and uncontrollable numerical diffusion. Spatial discretization necessitates the need for a numerically stable interpolation scheme to compute the value of the solution between two elements, whereas it is straightforward to compute the values from the analytic expression of the neural network. Furthermore, the number of elements scales exponentially with the number of dimensions. 

Nevertheless, for BVPs of low-dimensional PDEs PINN was still significantly slower and less accurate than an efficient spatial grid-based solver in computing solutions. To address this problem in \cite{dwivedi2020physics} the authors make use of the Extreme Learning Machine (ELM) \cite{huang2006extreme} algorithm, which significantly reduces training time of PINNs. ELM used a shallow neural network to approximate the solution and used Gauss-Newton Extremization (GNE) (Sec. \ref{sec:gne}) to train only the weights in the final layer of the neural network. In \cite{dong2021local} by combining ELM with domain decomposition, PINNs are shown to outperform FEM in terms of solution accuracy and computational time on a wide variety of BVPs. 

Another drawback of PINNs was the approximate loss function based method of imposing initial conditions (ICs) and boundary conditions (BCs). This required additional hyperparameter choice in determining the relative importance of loss functions. In \cite{lagaris1998artificial} a trial function was constructed from the neural network output which exactly satisfy the ICs and BCs. But this method was difficult to generalize due to a lack of a general prescription to generate the trial functions for arbitrary ICs and BCs. Recently, based on the Theory of Functional Connections (TFC) \cite{mortari2019multivariate, leake2020multivariate} a neural network method of solving IBVPs while exactly imposing the ICs and BCs was introduced \cite{leake2020deep}. This method was further developed to use ELM \cite{schiassi2021extreme} producing solutions orders of magnitude more accurate than grid-based methods in some instances. A fundamental limitation of shallow networks used in ELM is its lack of expressivity compared to a deep neural network (DNN) \cite{poole2016exponential}. In \cite{schiassi2021extreme} this results in DNN outperforming ELM in terms of solution accuracy in the context of Burgers equation and Navier-Stokes equations. In \cite{dong2021local} the authors could improve the expressivity by using multiple ELMs to represent the solution.

In this work, by combining the ideas of ELM and deep networks, we propose a framework wherein a DNN is initially trained using standard methods and then GNE is used to significantly improve the solution accuracy. By combining these methods we get the expressive power of a DNN \cite{poole2016exponential} combined with the fine tuning ability of shallow networks trained with ELMs. We also present a modification to the TFC framework used in \cite{schiassi2021extreme, leake2020deep} to significantly reduce the computational time required for training and inference.

The rest of this manuscript is structured as follows:
Sec. \ref{sec:pinn_intro} provides a concise introduction to PINN and discusses the need for TFC framework to impose constraints in BVPs.
Sec. \ref{sec:tfc_intro} introduces the TFC framework and demonstrates the application of TFC to impose constraints. In Sec. \ref{sec:rtfc} a modification to the TFC framework is proposed to speed up computation. In Sec. \ref{sec:tfc_remarks} we discuss the drawbacks of TFC and how to generalize to complex boundary geometries.
Sec. \ref{sec:pinn_training} details the neural network architectures and the training methods used for the examples provided in the section that follows.
Sec. \ref{sec:pinn_results} uses various known analytic solutions to linear and non-linear ODEs (Sec. \ref{sec:pinn_ode}), PDEs (Sec. \ref{sec:pinn_pde}) and Coupled PDEs (Sec. \ref{sec:pinn_cpde}) to show the superiority of our proposed method and discuss its drawbacks. Section 6 contains discussion. Some supplementary information is also provided in \ref{app:lstsq} - \ref{app:tfc_expressions}.

\section{Physics Informed Neural Networks}
\label{sec:pinn_intro}
Neural networks are considered as universal function approximators \cite{hornik1989multilayer, kidger2020universal}. Let $\mathcal{N}_{\{\theta\}}$ be an arbitrary function represented by a neural network, where $\{\theta\}$ denotes the parameters of the neural network. The idea of PINN is to make this neural network `physics informed' by imposing $\mathcal{N}_{\{\theta\}}$ to satisfy some given differential equation(s). 

Let $\mathcal{D}\left(\boldsymbol{x},f(\boldsymbol{x})\right)$ denote a general differential operator of an $m$-dimensional variable $\boldsymbol{x}$ acting on function $f(\boldsymbol{x})$. Any differential equation can be written as $\mathcal{D}\left(\boldsymbol{x},f(\boldsymbol{x})\right)=0$. In order to impose that $\mathcal{N}_{\{\theta\}}$ satisfy this differential equation, the differential equation itself can be used as a loss function while training. The derivatives in the differential equation can be accurately computed to an arbitrary precision using auto-differentiation \cite{margossian2019review}. Note that since neural networks approximate a function, in practice this differential equation is never exactly satisfied by $\mathcal{N}_{\{\theta\}}$. This deviation of the differential operator acting on $\mathcal{N}_{\{\theta\}}$ from identically being equal to zero is called residual, denoted by $\mathcal{R}$:
\begin{equation}
\label{eq:pinn_residual}
    \mathcal{R}(x; {\{\theta\}}) = \mathcal{D}\left(\boldsymbol{x},\mathcal{N}_{\{\theta\}}(\boldsymbol{x})\right).
\end{equation}
For a neural network training batch with $N$ samples $\boldsymbol{x}_1, \cdots, \boldsymbol{x}_n$, the loss function can now be defined as a mean squared residual (MSR) of the differential equation:
\begin{equation}
\label{eq:de_loss}
    \mathcal{L}_{DE}(\{\theta\}) = \frac{1}{N}\sum_{i=1}^{N}\|\mathcal{R}(\boldsymbol{x}_i, {\{\theta\}})\|^2.
\end{equation}
Minimizing the value of $\mathcal{L}_{DE}$ results in $\mathcal{N}_{\{\theta\}}$ more closely satisfying the differential equation $\mathcal{D}\left(\boldsymbol{x},f(\boldsymbol{x})\right)=0$. Note that, in the most general case we can have neural networks with multiple outputs $\mathcal{N}_1, \cdots, \mathcal{N}_l$ informed by a system of differential equations $\boldsymbol{\mathcal{D}} \equiv (\mathcal{D}_1, \cdots, \mathcal{D}_k) = 0$. Then the same Eq. \ref{eq:de_loss} holds with $\boldsymbol{\mathcal{R}} \equiv (\mathcal{R}_1, \cdots, \mathcal{R}_k)$.

In literature \cite{raissi2017physics, raissi2019physics, dwivedi2020physics, cai2022physics, lv2021hybrid, jagtap2021extended} $\mathcal{L}_{DE}$ can be found paired with other loss functions during training to achieve mainly three objectives:
\begin{itemize}
    \item \textbf{Data Driven Solution Discovery}\\
    The objective of this framework is to find the solution of a differential equation based on sparse experimental data. Let $v_E(\boldsymbol{y})$ denote some noisy experimental measurement of the state of the system at point $\boldsymbol{y}$. We can approximate this state using the neural network $\mathcal{N}^v_{\{\theta\}}$. For M experimental data points $\boldsymbol{y}_1, \cdots, \boldsymbol{y}_M$, the data driven loss can be defined as:
    \begin{equation}
        \mathcal{L}_{DD}(\{\theta\}) = \frac{1}{M}\sum_{i=1}^{M}\|\mathcal{N}^v_{\{\theta\}}(\boldsymbol{y}_i) - v_E(\boldsymbol{y}_i)\|^2
    \end{equation}
    The solution can be discovered by minimizing the combined loss function, which in this case will be a weighted sum of $\mathcal{L}_{DE}$ and $\mathcal{L}_{DD}$.
    \begin{equation}
    \label{eq:ddsd_loss}
        \mathcal{L}(\{\theta\}) = \omega_{DE} \mathcal{L}_{DE}(\{\theta\}) + \omega_{DD} \mathcal{L}_{DD}(\{\theta\})
    \end{equation}
    Here $\omega_{DE}$ and $\omega_{DD}$ are weights which needs to be carefully chosen for the optimal training of the neural network.
    \item \textbf{Data Driven Parameter Discovery}\\
    For a differential equation $\mathcal{D}_{\{\gamma\}}\left(\boldsymbol{x},f(\boldsymbol{x})\right)=0$ parameterized by a set of parameters $\{\gamma\}$. Experimental measurement of the state $v_E(\boldsymbol{y_i})$ at points $\{\boldsymbol{y_i}\}$ can be used to find the parameters of the differential equation. The loss function to minimize in this case will be the same as in Eq. \ref{eq:ddsd_loss} but with an additional set of parameters $\{\gamma\}$ to optimize.
    \begin{equation}
    \label{eq:ddpd_loss}
        \mathcal{L}(\{\theta\}, \{\gamma\}) = \omega_{DE} \mathcal{L}_{DE}(\{\theta\}, \{\gamma\}) + \omega_{DD} \mathcal{L}_{DD}(\{\theta\}, \{\gamma\})
    \end{equation}
    
    \item \textbf{Solving Boundary Value Problems}\\
    PINN framework can also be used to numerically solve the differential equation based on ICs and BCs. If $f(x)$ denotes a solution to the differential equation $\mathcal{D}\left(\boldsymbol{x},f(\boldsymbol{x})\right)=0$, we can denote its $n$ ICs and BCs applied on the respective boundaries $\partial \Omega_1, \cdots, \partial \Omega_n$ as a set of constraints: 
    \begin{equation}
    \begin{aligned}
        \mathcal{B}_1(f(\boldsymbol{x})) &=0, \quad \forall \boldsymbol{x}\in \partial \Omega_1\\
        \cdots&\\
        \mathcal{B}_n(f(\boldsymbol{x})) &=0, \quad \forall \boldsymbol{x} \in \partial \Omega_n
    \end{aligned}
    \end{equation}
    
        These constraints can be of any form including Dirichlet, Neumann and Robin BCs. The loss function for a single constraint can be written as
    \begin{equation}
        \mathcal{L}_{BC,k}(\{\theta\}) = \frac{1}{|\Omega_k|}\sum_{\boldsymbol{x}_i \in \Omega_k} \|\mathcal{B}_i(f(\boldsymbol{x}_i))\|^2
    \end{equation}
    where $|\Omega_k|$ denotes the number of datapoints used in the summation. The total loss for the ICs and BCs can then be the weighted sum of the individual losses, where the weights $\omega_{BC, i}, \cdots, \omega_{BC, n}$ need to be carefully chosen for optimal training.
    \begin{equation}
    \begin{aligned}
        \label{eq:bc_loss}
        \mathcal{L}_{BC}(\{\theta\}) &= \sum_{i=1}^{n}\omega_{BC, i} \mathcal{L}_{BC, i}(\{\theta\})
    \end{aligned}
    \end{equation}
    The solution of a BVP can then be found by minimizing the loss
    \begin{equation}
        \label{eq:bvp_loss}
        \mathcal{L}(\{\theta\}) = \mathcal{L}_{BC}(\{\theta\}) + \omega_{DE} \mathcal{L}_{DE}(\{\theta\})
    \end{equation}
\end{itemize}

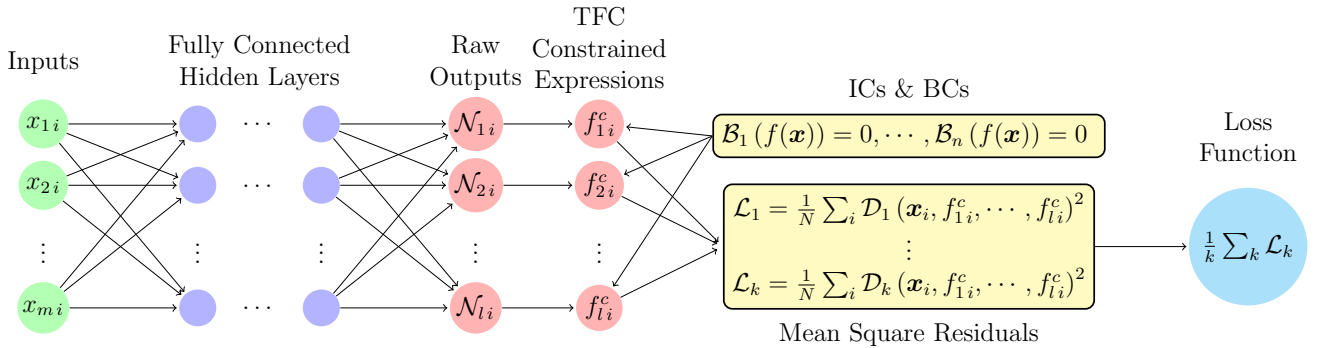
\begin{figure}[H]
    \centering
\resizebox{\columnwidth}{!}{
\begin{tikzpicture}

\foreach \s [count=\i] in {$x_{1 \, i}$, $x_{2 \, i}$,1, $x_{m \, i}$}
{
    \ifnum \i=3
    \node at (0,-\i)  {\vdots};
    \else
    \node[circle, minimum size = 8mm,fill=green!30, inner sep =0.5mm] (Input-\i) at (0,-\i) {\s};
    \fi
}

\foreach \i in {1,2,3,4}
{
    \ifnum \i=3
    \node at (2.5,-\i)  {\vdots};
    \else
    \node[circle, minimum size = 6mm,fill=blue!30] (Hidden1-\i) at (2.5,-\i) {};
    \fi
}

\node at (3.5,-1)  {\dots};
\node at (3.5,-2)  {\dots};
\node at (3.5,-4)  {\dots};

\foreach \i in {1,2,3,4}
{
    \ifnum \i=3
    \node at (4.5,-\i)  {\vdots};
    \else
    \node[circle, minimum size = 6mm,fill=blue!30] (Hidden2-\i) at (4.5,-\i) {};
    \fi
}

\foreach \s [count=\i] in {$\mathcal{N}_{1 \, i}$, $\mathcal{N}_{2 \, i}$,1, $\mathcal{N}_{l \, i}$}
{
    \ifnum \i=3
    \node at (7,-\i)  {\vdots};
    \else
    \node[circle, minimum size = 3mm,fill=red!30, inner sep = 0.5mm] (Output-\i) at (7,-\i) {\s};
    \fi
}

\foreach \s [count=\i] in {$f^c_{1 \, i}$, $f^c_{2 \, i}$,1, $f^c_{l \, i}$}
{
    \ifnum \i=3
    \node at (9,-\i)  {\vdots};
    \else
    \node[circle, minimum size = 3mm,fill=red!30, inner sep = 0.5mm] (Output1-\i) at (9,-\i) {\s};
    \fi
}

\foreach \j in {1,2,4}
    {
        \draw[->, shorten >=1pt]  (Output-\j) -- (Output1-\j);
    }

\foreach \i in {1,2,4}
    {
        \foreach \j in {1,2,4}
    {
        \draw[->, shorten >=1pt]  (Input-\j) -- (Hidden1-\i);
        \draw[->, shorten >=1pt]  (Hidden2-\j) -- (Output-\i);
    }
    }
\node[draw,thick,minimum width=2cm,minimum height=2cm, rounded corners, fill=yellow!30,align=center, anchor=west] (rect) at (11,-3) {
            $\mathcal{L}_1 = \frac{1}{N}\sum_i \mathcal{D}_1\left(\boldsymbol{x}_i, f^c_{1 \, i},  \cdots, f^c_{l \, i} \right)^2$\\
            \vdots \\
            $\mathcal{L}_k = \frac{1}{N}\sum_i \mathcal{D}_k\left(\boldsymbol{x}_i, f^c_{1 \, i}, \cdots, f^c_{l \, i} \right)^2$
            };

\node[draw,thick,minimum width=2cm,minimum height=0cm, rounded corners, fill=yellow!30,align=center, anchor=south, yshift=+1em] (rect2) at (rect.north) {
$\mathcal{B}_1\left(f(\boldsymbol{x})\right)=0, \cdots, \mathcal{B}_n\left(f(\boldsymbol{x})\right)=0$
};

%\node[draw, dashed,minimum width=1.5cm,minimum height=4.3cm, yshift = -1mm] (rect3) at (Output1-2.south) {};
%\draw[->, shorten >=1pt]  (rect2.west) -- (rect3.east|-rect2);

\foreach \j in {1,2,4}
    {
        \draw[->, shorten >=1pt]  (rect2.west) -- (Output1-\j);
    }

\foreach \j in {1,2,4}
    {
        \draw[->, shorten >=3pt]  (Output1-\j) -- (rect.west);
    }
    
\node[circle, minimum size = 6mm,fill=cyan!30, xshift = 6em] (Loss) at (rect.east) {$\frac{1}{k}\sum_k \mathcal{L}_k$};

\draw[->, shorten >=1pt]  (rect.east) -- (Loss);
    
\node [text centered] at (0,0) {Inputs};

\node [text width=8em, text centered] at (3.5,0) {Fully Connected Hidden Layers}; 
\node [text width=4em, text centered] at (Output-1|-,0) {Raw Outputs};
\node [text width=6em, text centered] at (Output1-1|-,0.2) {TFC Constrained Expressions};
\node [text width=16em, text centered, yshift = -1em] at (rect.south) {Mean Square Residuals};
\node [text width=16em, text centered, yshift = +1em] at (rect2.north) {ICs \& BCs};
\node [text width=4em, text centered, yshift = 2em] at (Loss.north) {Loss Function};

\end{tikzpicture}
}
    \caption{Structure of a fully connect neural network solving a BVP with $k$ coupled differential equations $\mathcal{D}_1, \cdots, \mathcal{D}_k$ and $n$ initial and boundary conditions $\mathcal{B}_1, \cdots, \mathcal{B}_n$ acting on $l$ functions dependent on $m$ variables $x_1, \cdots, x_m$}
    \label{fig:pinn}
\end{figure}

In this work we will focus our attention solely on BVP. Since this is a multi-objective optimization problem (Eqs. \ref{eq:bc_loss} \& \ref{eq:bvp_loss}), the accuracy of the solutions will be sensitive to the hyperparameters $\omega_{DE}, \omega_{BC,1}, \cdots, \omega_{BC,n}$. Tuning of these hyperparameters becomes challenging and time consuming especially if a large number of boundary conditions is given. In \cite{leake2020deep, schiassi2021extreme} this issue was overcome by using a mathematical framework called Theory of Functional Connections (TFC) to exactly satisfy the ICs and BCs without using a loss function. We provide a concise introduction to TFC in Sec. \ref{sec:tfc_intro}. Using TFC the neural network ouput $\mathcal{N}(x)$ is converted to a constrained function $f^c(x)$ which always satisy the ICs and BCs. The training then becomes a single objective optimization problem as shown in Fig. \ref{fig:pinn}.

\section{Theory of Functional Connections}
\label{sec:tfc_intro}
TFC is a mathematical framework for converting constrained problems into unconstrained problems. The idea of TFC is to write down the most general function which satisfies the constraints of the problem. In the case of PINNs, the constraints are in the form of initial conditions and boundary conditions of the differential equations. In the following subsections we will provide a brief self-contained exposition of TFC. See \cite{mortari2019multivariate, leake2020multivariate} for a rigorous introduction.

\subsection{Univariate Case}
Let's consider an arbitrary univariate function $f(x)$ subject to the constraints $\mathcal{C}_1[f] = c_1, \mathcal{C}_2[f] = c_2, \cdots, \mathcal{C}_n[f] = c_n$, where $\mathcal{C}_1, \cdots, \mathcal{C}_n$ are linear functionals. Functional in this context refers to a mapping from the space of all functions to the real/complex numbers. The term linear implies that $\mathcal{C}[\alpha\, f + \beta \, g] = \alpha\, \mathcal{C}[f] + \beta\, \mathcal{C}[g]$ for any two functions $f$ and $g$, and any two real/complex numbers $\alpha$ and $\beta$. Some examples of linear functionals are: $\mathcal{C}[f] = f(x_1)$, $\mathcal{C}[f] = f'(x_1)$, $\mathcal{C}[f] = \int_a^b f(x) dx$. The idea of TFC is to write down a constrained expression $f^c(x)$ for $f(x)$, such that $f^c(x)$ always satisfy the constraints:
\begin{equation}
\label{eq:tfc_1d_general}
    f^c(x) = \mathcal{N}(x) + \sum_{i=1}^{n}\eta_i s_i(x).
\end{equation}
Here $\mathcal{N}(x)$ is some arbitrary function known as the free function, which in the context of PINN will be represented by a neural network. $\eta_1, \cdots, \eta_n$ are some unknown coefficients and $\{s_1(x), \cdots, s_n(x)\}$ is a set of linearly independent functions we choose, called the support functions. An example of a set of support functions is the polynomial basis $\{1, x, \cdots, x^{n-1}\}$. Since by definition $f^c(x)$  satisfy all the constraints, Eq. \ref{eq:tfc_1d_general} can be written as the following set of equations:
\begin{equation}
    \label{eq:tfc_1d_solution}
    \begin{aligned}
    c_1 &= \mathcal{C}_1[f] = \mathcal{C}_1[\mathcal{N}] + \sum_{i=1}^{n}\eta_i \mathcal{C}_1[s_i],\\
    &\vdots\\
    c_n &= \mathcal{C}_n[f] = \mathcal{C}_n[\mathcal{N}] + \sum_{i=1}^{n}\eta_i \mathcal{C}_n[s_i].
    \end{aligned}
\end{equation}

%The linearity of the functionals was used to split the RHS of Eq. \ref{eq:tfc_1d_eta_solution} into multiple terms. Once a set of support functions is chosen, we can solve for $\eta_i$ to get an exact expression for $f^c(x)$ in terms of the free function $\mathcal{N}(x)$. Note that in some rare cases the support functions will need to be carefully chosen to make the matrix in Eq. \ref{eq:tfc_1d_eta_solution} invertible. For most commonly occuring ICs and BCs the polynomial basis can be used as the support functions. 

Once a set of support functions is chosen, we can solve for $\eta_i$ to get an exact expression for $f^c(x)$ in terms of the free function $\mathcal{N}(x)$:

\begin{equation}
\label{eq:tfc_1d_eta_solution}
    \begin{bmatrix}
    \eta_1 \\
    \vdots \\
    \eta_n
    \end{bmatrix}
    =
    \begin{bmatrix}
    \mathcal{C}_1[s_1] & \cdots & \mathcal{C}_1[s_n]\\
    \vdots & \ddots & \vdots\\
    \mathcal{C}_n[s_1] & \cdots & \mathcal{C}_n[s_n]
    \end{bmatrix}^{-1}
    \begin{bmatrix}
    c_1 - \mathcal{C}_1[\mathcal{N}] \\
    \vdots\\
    c_n-\mathcal{C}_n[\mathcal{N}]
    \end{bmatrix}.
\end{equation}
The linearity of the functionals was used to split the RHS of Eq. \ref{eq:tfc_1d_solution} into multiple terms.  Note that in some rare cases the support functions will need to be carefully chosen to make the matrix in Eq. \ref{eq:tfc_1d_eta_solution} invertible. For most commonly occuring ICs and BCs the polynomial basis can be used as the support functions. 

When solving a differential equation with PINN, $\mathcal{N}(x)$ represents the neural network output, $\mathcal{C}_1, \cdots, \mathcal{C}_n$ the initial and boundary conditions and $f^c(x)$ the solution of the differential equation which satisfy the initial and boundary conditions.

\subsection{Multivariate Case}
For a multi-variate function $f(x_1, x_2, \cdots x_m)$, this procedure of finding the constrained expression can be applied iteratively on one variable at a time to get the general constrained expression. To do so we first define the  operator $\mathcal{C}^i$ which is a mapping from the space of multivariate functions which depend on $x_1, x_2, \cdots x_m$, to a space of multivariate functions which depend on $x_1, \cdots x_{i-1}, x_{i+1}, \cdots, x_m$. In short the operator $\mathcal{C}^i$ acts only on the i\textsuperscript{th} variable of the multivariate function. Now we can represent $n_i$ constraints on each of the $i^{th}$ dimension as follows:
\begin{equation}
\label{eq:multivariate_constraints}
    \begin{aligned}
    \mathcal{C}^i_1[f] &= c^i_1(x_1, \cdots x_{i-1}, x_{i+1}, \cdots, x_m),\\
    \mathcal{C}^i_2[f] &= c^i_2(x_1, \cdots x_{i-1}, x_{i+1}, \cdots, x_m),\\
    &\vdots \\
    \mathcal{C}^i_{n_i}[f] &= c^i_{n_i}(x_1, \cdots x_{i-1}, x_{i+1}, \cdots, x_m),
    \end{aligned}
\end{equation}

where $\mathcal{C}^i_1, \cdots, \mathcal{C}^i_{n_i}$ are linear operators acting only on the variable $x_i$. For example the Dirichlet BC $f(x_1, 5, x_3, \cdots, x_n) = \sin (x_1+x_3+x_4+\cdots+x_n)$ has the form in Eq. \ref{eq:multivariate_constraints} with $\mathcal{C}^2 = f(x_1, 5, x_3, \cdots, x_n)$ and $c^2(x_1,x_3,x_4, \cdots,x_n) = \sin (x_1+x_3+x_4+\cdots+x_n)$. Similarly we can represent Neumann and Robin BCs. Note that this form of constraints which are represented by linear operators which act on a single variable limits the shape of domain boundaries on which BCs can be applied. BCs can only be applied on boundaries of the form $x_i = constant$. This is a fundamental limitation of the TFC framework. See \ref{sec:tfc_remarks} for further discussion and generalization.

Now starting with a free function $\mathcal{N}(x_1, \cdots, x_n)$, represented by the neural network, we can iteratively apply the constraints to derive the general constrained expression as follows:

\begin{equation}
\label{eq:tfc_nd_general}
\begin{aligned}
    f^c_1(x_1, \cdots, x_m) &= \mathcal{N}(x_1, \cdots, x_n) + \sum_{i=1}^{n_1}\eta^1_i s_i(x_1)\\
    f^c_2(x_1, \cdots, x_m) &= f^c_1(x_1, \cdots, x_n) + \sum_{i=1}^{n_2}\eta^2_i s_i(x_2)\\
    &\vdots \\ 
    f^c_m(x_1, \cdots, x_m) &= f_{m-1}^c(x_1, \cdots, x_n) + \sum_{i=1}^{n_m}\eta^m_i s_i(x_m)
    \end{aligned}.
\end{equation}
Here $f^c_1$ satisfies the $\mathcal{C}^1$ constraints, $f^c_2$ satisfies the $\mathcal{C}^1$ and $\mathcal{C}^2$ constraints, and so on. Using the same method for the one-dimensional case (Eq. \ref{eq:tfc_1d_general} \& Eq. \ref{eq:tfc_1d_solution}), at each iteration we can compute the $\eta_i^j$. $f^c \equiv f^c_m$ now satisfy all the constraints. Note that at each step of the iteration we can use different set of support functions. In this work we exclusively use polynomial basis functions as support functions. For high-dimensional functions, the algebra of this iterative application of constraints becomes cumbersome, therefore we make use of symbolic computation capabilities of Mathematica to get the constrained expression.

\subsection{Reduced TFC}
\label{sec:rtfc}
For the univariate case of TFC, in order to get the exact expression for $f^c(x)$ (Eq. \ref{eq:tfc_1d_general}), we need to derive the expression for $\eta_i$ using Eq. \ref{eq:tfc_1d_eta_solution}. This solution contains the quantities $\mathcal{C}_1[\mathcal{N}], \cdots, \mathcal{C}_n[\mathcal{N}]$. In the case of PINN, since $\mathcal{N}(x)$ is represented by a neural network, computing these quantities requires additional evaluation of the neural network. The number of additional evaluations depends on the type of constraints. If $\mathcal{C}$ is a value constraint like $\mathcal{C}[\mathcal{N}] = \mathcal{N}(\tilde{x})$ where $\tilde{x}$ is some constant, computing $\mathcal{C}_n[\mathcal{N}]$ requires an evaluation of the neural network for the value $\tilde{x}$, a derivative constraint $\mathcal{C}[\mathcal{N}] = \mathcal{N}'(\tilde{x})$ requires a more expensive auto-differentiation to evaluate, and an integral constraint $\mathcal{C}_n[\mathcal{N}] = \int\mathcal{N}(x) dx$ requires multiple evaluations. For $n$ constraints, computing of $f^c(x)$ in the univariate case therefore requires atleast $n+1$ evaluations of the neural network.

For the case of the multivariate functions, for the sake of simplicity we will assume all the constraints are value constraints. In Eq. \ref{eq:tfc_nd_general}, following similar arguments for the univariate case, evaluation of $f^c_1$ requires $n_1+1$ evaluation of the neural network. The evaluation of $f_2^c$ requires $n_2+1$ evaluations of $f^c_1$, hence $(n_2+1)\times(n_1+1)$ evaluations of the neural network. Using inductive logic we get that evaluation of $f^c_m$ requires $(n_1+1)\times\cdots\times(n_m+1)$ evaluations of the neural network. 

In this work, we propose the idea of Reduced TFC to reduce the number of evaluations of the neural network, which can be computationally expensive. To achieve this we note that majority of the boundary value problems we encounter in practice have value or derivative constraints. 

Suppose for the univariate case we have a $k^{th}$ derivative constraint $\mathcal{C}[f] := f^{(k)}(\tilde{x}) = c$ where $\tilde{x}$ and $c$ are some constants, we can modify the free function in Eq. \ref{eq:tfc_1d_general}, $\mathcal{N}(x) \to \mathcal{N}(x)(x-\tilde{x})^{k+1}$, so that $\mathcal{C}[\mathcal{N}(x)(x-\tilde{x})^{k+1}] = 0$. Therefore an additional evaluation of the $\mathcal{N}(x)$ represented by the neural network is not required. To generalize this, if we have $n$ derivative constraints $f^{(k_1)}(\tilde{x}_1) = c_1, \cdots, f^{(k_n)}(\tilde{x}_n) = c_n$, the Reduced TFC constraint expression can be written as
\begin{equation}
\label{eq:rtfc_1d_general}
    f^c(x) = \mathcal{N}(x)(x-\tilde{x}_1)^{k_1+1}\cdots(x-\tilde{x}_n)^{k_n+1}+ \sum_{i=0}^{n}\eta_i s_i(x).
\end{equation}
The solution for $\eta_i$ will then be given by a simplified version of Eq. \ref{eq:tfc_1d_eta_solution}:
\begin{equation}
\label{eq:rtfc_1d_eta_solution}
    \begin{bmatrix}
    \eta_1 \\
    \vdots \\
    \eta_n
    \end{bmatrix}
    =
    \begin{bmatrix}
    s_1^{(k_1)}(x_1) & \cdots & s_n^{(k_1)}(x_1)\\
    \vdots & \ddots & \vdots\\
    s_1^{(k_n)}(x_n) & \cdots & s_n^{(k_n)}(x_n)
    \end{bmatrix}^{-1}
    \begin{bmatrix}
    c_1 \\
    \vdots\\
    c_n
    \end{bmatrix}
\end{equation}

In the multivariate case, for the set of constraints 
\begin{equation}
\mathcal{C}^i_j[f] := \left. \partial_{x_i}^{k_j}f(x_1,\cdots,x_m)\right|_{x_i = \tilde{x}^i_j} = c^i_j
\end{equation}
the following modification is required to the free function:
\begin{equation}
    \mathcal{N}(x_1,\cdots,x_m) \to \mathcal{N}(x_1,\cdots,x_m) \prod_{i,j} (x_i - \tilde{x}^i_j)^{k_j+1}.
\end{equation}
This can then be substituted in Eq.~\ref{eq:tfc_nd_general} and all $\eta^i_j$ can be computed. The resulting $f^c(x_1,\cdots,x_m)$ requires only a single evaluation of $\mathcal{N}(x_1,\cdots,x_m)$ which is represented by the neural network. As we will see in Sec. \ref{sec:heat_pde}, Sec. \ref{sec:3p1_nl_pde} and Sec. \ref{sec:kovas}, compared to TFC, Reduced TFC will lead to a significant speedup in computing the solution of PDEs.

\subsection{Remarks}
\label{sec:tfc_remarks}
For the multivariate TFC while deriving the expression for the constrained expression $f^c$ (Eq. \ref{eq:tfc_nd_general}), we assume that the constraints are of the form $\mathcal{C}_{j}^i[f]=c_{j}^i\left(x_1, \cdots x_{j-1}, x_{j+1}, \cdots, x_m\right)$, where $\mathcal{C}_{j}^i$ is an operator acting only on the variable $x_j$. This limits the geometry of the hypersurface on which ICs and BCs can be applied. The hypersurface should always be of the form $\partial \Omega = \{\boldsymbol{x} \mid x_j = \text{const.}\}$ for some coordinate $x_j$ or some combination of it. The class of BVPs where ICs and BCs are applied to the boundary of the domain $[a_1,b_1]\times\cdots\times[a_m,b_m]$ satisfy this condition. In real world applications a large class of BVPs are defined with BCs defined on more complex geometries. The unit circle $\partial\Omega = \{(x,y) \mid x^2 + y^2 = 1\}$ is an example of a boundary on which the TFC framework fails since it cannot be expressed in the form $\{\boldsymbol{x} \mid x_j = \text{const.}\}$. While for this specific case we can still use TFC after a coordinate change to the polar coordinates $(r, \theta)$ and the boundary can be represented by the equation $r=1$, such coordinate changes are not possible in general.

For complex boundary geometries in 2D an extension of TFC using bijective mapping can be used \cite{mortari2020bijective}. For more complicated geometries in arbitrary dimensions, a method based on approximate distance functions has been proposed in \cite{sukumar2022exact}. The form of Reduced TFC expression (Eq. \ref{eq:rtfc_1d_general}) can also be generalized to complex geometries. Note that the form of Eq. \ref{eq:rtfc_1d_general} is such that the 1\textsuperscript{st} term in the RHS is identically zero on the boundaries and the 2\textsuperscript{nd} term satisfies the values at the boundary. This can in principle be extended to more complex boundaries where Dirichlet, Neumann or mixed boundary conditions are imposed. Let $b_1(\boldsymbol{x})=0, b_2(\boldsymbol{x})=0, \cdots, b_n(\boldsymbol{x})=0$ be the $n$ boundaries where BCs are applied and $k_1, k_2, \cdots, k_n$ be the order of the derivative constraint applied at the boundary. Note that $k=0$ denotes a Dirichlet BC. Then the constrained expression can be written as
\begin{equation}
\label{eq:rtfc_g}
    f^c(\boldsymbol{x}) = \mathcal{N}(\boldsymbol{x}) b_1(\boldsymbol{x})^{k_1+1} \cdots b_n(\boldsymbol{x})^{k_n+1} + \mathcal{G}(\boldsymbol{x}).
\end{equation}
Here the 1\textsuperscript{st} term in the RHS is identically zero at the boundaries after applying the appropriate number of derivative operations and the 2\textsuperscript{nd} term $\mathcal{G}(\boldsymbol{x})$ satisfies the values at the boundaries. For a combination of boundary conditions it is not straightforward to derive a function $\mathcal{G}(\boldsymbol{x})$ to satisfy the values at the boundaries for all boundary conditions. For rectangular boundaries this is solved using TFC. For more complicated BCs a neural network can be used to represent $\mathcal{G}(\boldsymbol{x})$ and the neural network can be trained to learn the values at the boundaries using the loss function in Eq. \ref{eq:bc_loss}. Once trained, the neural network $\mathcal{G}(\boldsymbol{x})$ can be used to impose the boundary conditions and only the neural network $\mathcal{N}(\boldsymbol{x})$ needs to be trained to solve the differential equation. Further investigation is needed in this direction to find better ways to derive $\mathcal{G}(\boldsymbol{x})$ for the case of multiple BCs with complex boundary geometries.

For the sake of simplicity, in this work we mostly use ICs and BCs on rectangular boundaries where TFC is applicable. Note that TFC and Reduced TFC cannot be used to impose periodic BCs. For PINNs periodic BCs can be imposed by changing the neural network architecture \cite{dong2021method}. This is explored in Sec. \ref{sec:taylor_pde}. Also see Sec. \ref{sec:pure_advection} for a case where BCs are imposed at infinity.

\section{Neural Network \& Training}
\label{sec:pinn_training}
In this work, we use a fully connected neural network (FCNN) to represent the function(s) which are solution(s) of differential equation(s). FCNNs have been previously used in the context of PINNs \cite{lagaris1998artificial, dong2021local, dong2021method, leake2020deep}. In this work, we mainly use two different network architectures: a single hidden layer (shallow) FCNN \cite{huang2006extreme} and a multi-layer (deep) FCNN. Based on previous studies \cite{dwivedi2020physics, schiassi2021extreme}, a shallow FCNN, trained with extreme learning machine (ELM) \cite{huang2006extreme} algorithm is found to outperform DNNs in terms of solution accuracy and training speed in different scenarios. The ELM algorithm uses Guass-Newton Extremization (GNE) to train the weights in the final layer of the neural network.

\subsection{Gauss-Newton Extremization}
\label{sec:gne}
GNE is a popular algorithm for non-linear least squares optimization. Let $\mathcal{N}_{\{\lambda\}}(\boldsymbol{x})$ be a non-linear fitting function parameterized by a set of parameters $\{\lambda\}$. If $\mathcal{N}$ represents a neural network $\{\lambda\}$ can be a subset of parameters of the neural network. Now for an input variable $x_i$ and its corresponding output $y_i$, the residual of the fitting function $\mathcal{R}_i(\{\lambda\})$ is defined as 
\begin{equation}
    \mathcal{R}_i(\{\lambda\}) := y_i - \mathcal{N}_{\{\lambda\}}(\boldsymbol{x})
\end{equation}
Note that for solving BVPs using PINNs the residual $\mathcal{R}$ will be given by Eq. \ref{eq:pinn_residual}. The Gauss-Newton algorithm is then used to iteratively minimize the sum of squares of the residual, $\sum_i \mathcal{R}_i(\{\lambda\})^2$.  Starting from an initial guess $\boldsymbol{\lambda}_0$ for the parameter vector, GNE prescribes the following iteration to minimize the sum of squares of $\mathcal{R}_i$ :
\begin{equation}
\label{eq:gauss_newton}
    \boldsymbol{\lambda}_{t+1} - \boldsymbol{\lambda}_{t}=  - \left(\mathbf{J}_{t}^{\top} \mathbf{J}_{t}\right)^{-1} \mathbf{J}_{t}^{\top} \boldsymbol{\mathcal{R}}(\boldsymbol{\lambda}_t).
\end{equation}
Here $\boldsymbol{\mathcal{R}} \equiv (\mathcal{R}_i, \cdots )$ is the residual vector and $\mathbf{J}$ is the Jacobian matrix defined as
\begin{equation}
\label{eq:jacobian}
    \left(\mathbf{J}_{t}\right)_{ij} := \frac{\partial \mathcal{R}_i\left(\boldsymbol{\lambda}_t\right)}{\partial \left(\boldsymbol{\lambda}_t\right)_j}.
\end{equation}
Note that Eq. \ref{eq:gauss_newton} involves a matrix inversion, which is numerically unstable. An alternate way to compute the quantity on the RHS of Eq. \ref{eq:gauss_newton} is to note that for the linear least squares (LLS) problem of minimizing $\|\mathbf{J}_{t}-  \boldsymbol{\mathcal{R}}(\boldsymbol{\lambda}_t)\boldsymbol{\beta}\|^2$ w.r.t. $\boldsymbol{\beta}$, the analytic solution is given by $\boldsymbol{\beta} = \left(\mathbf{J}_{t}^{\top} \mathbf{J}_{t}\right)^{-1} \mathbf{J}_{t}^{\top} \boldsymbol{\mathcal{R}}(\boldsymbol{\lambda}_t)$. Therefore the RHS of Eq. \ref{eq:gauss_newton} can be computed in a numerically stable way using LLS. In this work we use the PyTorch function \textit{torch.linalg.lstsq} for LLS which utilizes the standard implementation in LAPACK library \cite{lapack}. LAPACK provides several different algorithms for computing the LLS solution. Out of these methods, singular value decomposition (SVD) is found to provide consistent results throughout different problems we have considered in this work. See \ref{app:lstsq} for details.

\subsection{Neural Network Architecture}
\textbf{The ELM architecture} consists of a single hidden layer FCNN, with randomly initialized constant input weights and biases, and trainable weights and no bias in the final output layer. The ELM architecture with $m$ inputs $(x_1, \cdots, x_m)$, $l$ outputs $(\mathcal{N}_1, \cdots, \mathcal{N}_l)$ and $H$ nodes in the hidden layer can be written as
\begin{equation}
    \mathcal{N}_k = \sum_{j=1}^H \beta_{kj} \sigma\left(\sum_{i=1}^m w_{ji} x_i+b_j\right), \quad 1 \leq k \leq l.
\end{equation}
Here $w_{ji}$ and $b_j$ are the untrainable weights and biases respectively, and $\beta_{kj}$ are the trainable weights. $\sigma$ is a non-linear function known as the activation function. We use $\tanh$ activation throughout this work. The values of $\beta_{kj}$ are computed using GNE during training.

\textbf{The deep FCNN architecture} we use in this study consists of $M$ hidden layers. All hidden layers except the last has $H$ nodes with biases and the last hidden layer has $H_E$ nodes without bias. This choice was made to replicate the structure of ELM for the last layer. The mathematical description of the neural network can be written as:

\begin{equation}
\label{eq:fcnn_eq}
\begin{aligned}
    h^{(1)}_{j}&= \sigma\left(\sum_{i=1}^m w^{(1)}_{ji} x_i+b_j^{(1)}\right), \\
    h^{(2)}_{j}&= \sigma\left(\sum_{i=1}^H w^{(2)}_{ji} h^{(1)}_{i}+b_j^{(2)}\right), \\
    & \vdots \\
    \mathcal{N}_k&=\sum_{j=1}^{H_{E}}\beta_{kj} h^{(M)}_{j}, \quad 1 \leq k \leq l.
\end{aligned}
\end{equation}
Here $h^{(k)}_{j}$ are the values of nodes in the $k^{th}$ hidden layer and $w^{(k)}_{ji}$ and $b_j^{(1)}$ the corresponding weights and biases respectively. $\beta_{kj}$ are the weights of the final output layer. If we choose $M = 1$ we get the ELM architecture.

See Fig. \ref{fig:pinn} for the complete layout of the PINN.
\subsection{Training}
\label{sec:training}
The deep FCNN used in the context of PINNs are typically trained with stochastic gradient descent method like Adam \cite{kingma2014adam} or quasi-Newton method like BFGS \cite{fletcher2013practical}. In most cases BFGS optimizer is found to provide the best solution accuracy for PINN. An exception can be found in Sec. \ref{sec:sin_eq}, where we used Adam. In all other examples considered in this work a limited memory variant of the BFGS algorithm (L-BFGS) \cite{liu1989limited} is used for optimizing the parameters of the deep network. The specific implementation of the optimizers we use are from the PyTorch library
\cite{pytorch}. In this work we propose that after training the deep FCNN with the desired optimizer, the $\beta_{kj}$ parameters be further optimized using GNE. In Sec. \ref{sec:pinn_results} we show that combining optimizers in this manner results in orders of magnitude improvement in the solution accuracy.

Note that any linear differential equation $\mathcal{D}\left(\boldsymbol{x}, f(\boldsymbol{x})\right)$ can be written as $\tilde{\mathcal{D}}\left(\boldsymbol{x}, f(\boldsymbol{x})\right) + g(\boldsymbol{x})$, where $\tilde{\mathcal{D}}$ is the linear homogeneous part. If one of the outputs $\mathcal{N}_k$ of the neural network satisfies a linear differential equation, using Eq. \ref{eq:fcnn_eq} we can write
\begin{equation}
    \mathcal{D}\left(\boldsymbol{x}, \mathcal{N}_k(\boldsymbol{x})\right) = \sum_{j=1}^{H_{E}}\beta_{kj}\tilde{\mathcal{D}}\left(\boldsymbol{x}, h^{(M)}_{j}(\boldsymbol{x})\right) + g(\boldsymbol{x}).
\end{equation}
Finding the minimum of the residual of $\mathcal{D}\left(\boldsymbol{x}, \mathcal{N}_k(\boldsymbol{x})\right)$ w.r.t. $\beta_{kj}$ now becomes a LLS problem which can be minimized with a single GNE iteration. The same argument applies to a coupled linear system of differential equations and the weights $\beta_{kj}$ can be computed using a single GNE iteration. The arguments still hold when using TFC constrained expression, since TFC and Reduced TFC use linear transformations to generate the constrained expressions. For a non-linear differential equation, we need to perform multiple iterations in GNE and the iteration is stopped when the MSR (Eq. \ref{eq:de_loss}) stops decreasing.

We train the neural network by uniformly sampling points from the domain and compute the loss function (Eq. \ref{eq:de_loss}). For ELM the loss function is minimized w.r.t. $\beta_{kj}$ using GNE. For the deep network, the loss function is first minimized w.r.t. all the parameters using Adam or L-BFGS. Then after the desired number of iterations is reached, the loss function is minimized w.r.t. $\beta_{kj}$ using GNE. New random samples are generated for each iteration of Adam or L-BFGS. This prevents overfitting even if the number of samples is small. For GNE we use a larger number of sample points to prevent overfitting. Throughout this study the number of sample points for Adam or L-BFGS and GNE is chosen after a crude manual hyperparameter search. 

Performing GNE requires computing the Jacobian. In this work for the ease of implementation, Jacobian is computed using forward mode auto-differentiation implemented in PyTorch as \textit{torch.autograd.functional.jacobian()}. The specific vectorized implementation of Jacobian in PyTorch is memory intensive and while performing GNE using a given set of sample points, the Jacobian needs to be computed in batches and the batch size depends on the the neural network architecture and the operations performed to compute the residual. Note that it is possible to significantly speed up the computation of Jacobian by deriving an analytic expression for it. A drawback is that it needs to be done on a case by case basis. This process can be automated using symbolic computation packages like Mathematica but is not pursued in this work.

During training, the progress is quantified using the root mean squared residual (RMSR) of the differential equation or the system of differential equations. Since the actual training curves of the neural network will have significant fluctuations, the training curves shown in Sec. \ref{sec:pinn_results} represent the values of the best model which has the lowest RMSR obtained until that particular training step. Therefore the training curves will be monotonically decreasing. Throughout this work, unless stated otherwise, all the parameters of the neural networks are initialized using Xavier uniform initialization \cite{glorot2010understanding}. We use $\tanh$ activation function in this study for simplicity. See \cite{jagtap2020locally, gnanasambandam2022self} for trainable activation functions.

\section{Results}
\label{sec:pinn_results}
In this section by solving different BVPs we show the superiority of the proposed extremization method for DNNs. The ICs and BCs are applied using TFC or Reduced TFC. The exact expressions can be found in \ref{app:tfc_expressions}. Some examples considered in this section are taken from \cite{lagaris1998artificial, schiassi2021extreme}. Note that ELMs and DNNs trained after applying boundary conditions using TFC is referred to as X-TFC and Deep-TFC respectively in \cite{schiassi2021extreme}. All computations in this section were performed using double-precision arithmetic on a workstation with AMD EPYC 7552 CPU, Nvidia Titan RTX GPU and 64GB of RAM. The general outline of this section is given in Table \ref{tab:result_summary}:

    \begin{longtable}{|p{0.07\linewidth}|p{0.23\linewidth}|p{0.6\linewidth}|}
        \hline
         Section &  Problem & Remarks \\
        \hline
         \multicolumn{3}{|c|}{\textbf{ODE}}\\
        \hline
          \ref{sec:sin_eq} & $y'(t) = -\sin{t}$ & Showcase limitation of ELM when learning complex solutions and necessity of deep networks. GNE solution is demonstrated to improve with increase in number of neurons in the final layer. Adam used since L-BFGS fails to converge.\\
        \hline
         \ref{sec:stiff_eq} & Non-linear coupled ODEs & An example where traditional numerical methods fail. Incremental training method introduced. PINN fails without incremental training.\\
        \hline
         \multicolumn{3}{|c|}{\textbf{PDE}}\\
        \hline
         \ref{sec:l_v_nl} & 2D linear \& non-linear PDEs with same solution & ELM \& L-BFGS + GNE give similar solution accuracy for linear PDE. L-BFGS + GNE outperforms ELM for non-linear PDE.\\
        \hline
         \ref{sec:burgers_eq} & 1+1 D Burgers Equation & GNE fails if large gradients are present. L-BFGS + GNE shown to work over a wider range of gradients compared to ELM. Drawback of TFC discussed.\\
        \hline
         \ref{sec:heat_pde} & 2+1 D Heat Equation & Smaller solution error with L-BFGS compared to ELM but significantly slower. L-BFGS + GNE improves upon L-BFGS by more than 2 orders of magnitude with a small increase in computation time over ELM. 4x speedup with Reduced TFC compared to TFC. \\
        \hline
         \ref{sec:3p1_nl_pde} & 3+1 D Non-linear PDE & 20x - 40x speedup with Reduced TFC compared to TFC.\\
        \hline
         \multicolumn{3}{|c|}{\textbf{Coupled PDEs}}\\
        \hline
         \ref{sec:kovas} & Kovasznay flow & Steady state solution to 2D incompressible stationary Navier-Stokes equations. L-BFGS + GNE significantly outperforms traditional as well as other neural network based methods.\\
        \hline
         \ref{sec:taylor_pde} & Taylor-Green vortex  & Unsteady solution to 2D incompressible Navier-Stokes equations. Periodic BCs applied by changing the neural network architecture and using sinusoidal activation function. Showcase benefits of domain decomposition on large domains.\\
        \hline
         \ref{sec:pure_advection} & Pure advection in 1+1 D Euler Equation & ELM fails to solve. Incremental training method used. Imposing vanishing BCs at infinity.\\
        \hline
    \caption{A summary of different problems considered in the Results section. The Remarks column outlines important observations for each of the considered problems.}
    \label{tab:result_summary}
    \end{longtable}

\subsection{ODE}
\label{sec:pinn_ode}
\subsubsection{$y'(t) = -\sin{t}$}
\label{sec:sin_eq}
\FloatBarrier
In this subsection we will use a simple example $y'(t) = -\sin{t}$ to show the limitation of ELM in learning complicated functions. Since this is a linear ODE, a single Gauss-Newton iteration is required to find the solution using ELM. In this example we adjust the complexity of the solution by changing the size of the solution domain $[0,t_{max}]$. Throughout this example neural networks are trained using 1000 random uniformly sampled points from the domain for Adam and 2000 random uniformly sampled points for GNE. For uniformity all networks using Adam were trained for 100,000 steps. Note that this training method is extremely slow and the incremental training introduced in Sec. \ref{sec:stiff_eq} will significantly speed up the process, but is not used here for simplicity.

\begin{figure}[h]
    \centering
    \includegraphics[width = 0.49 \linewidth]{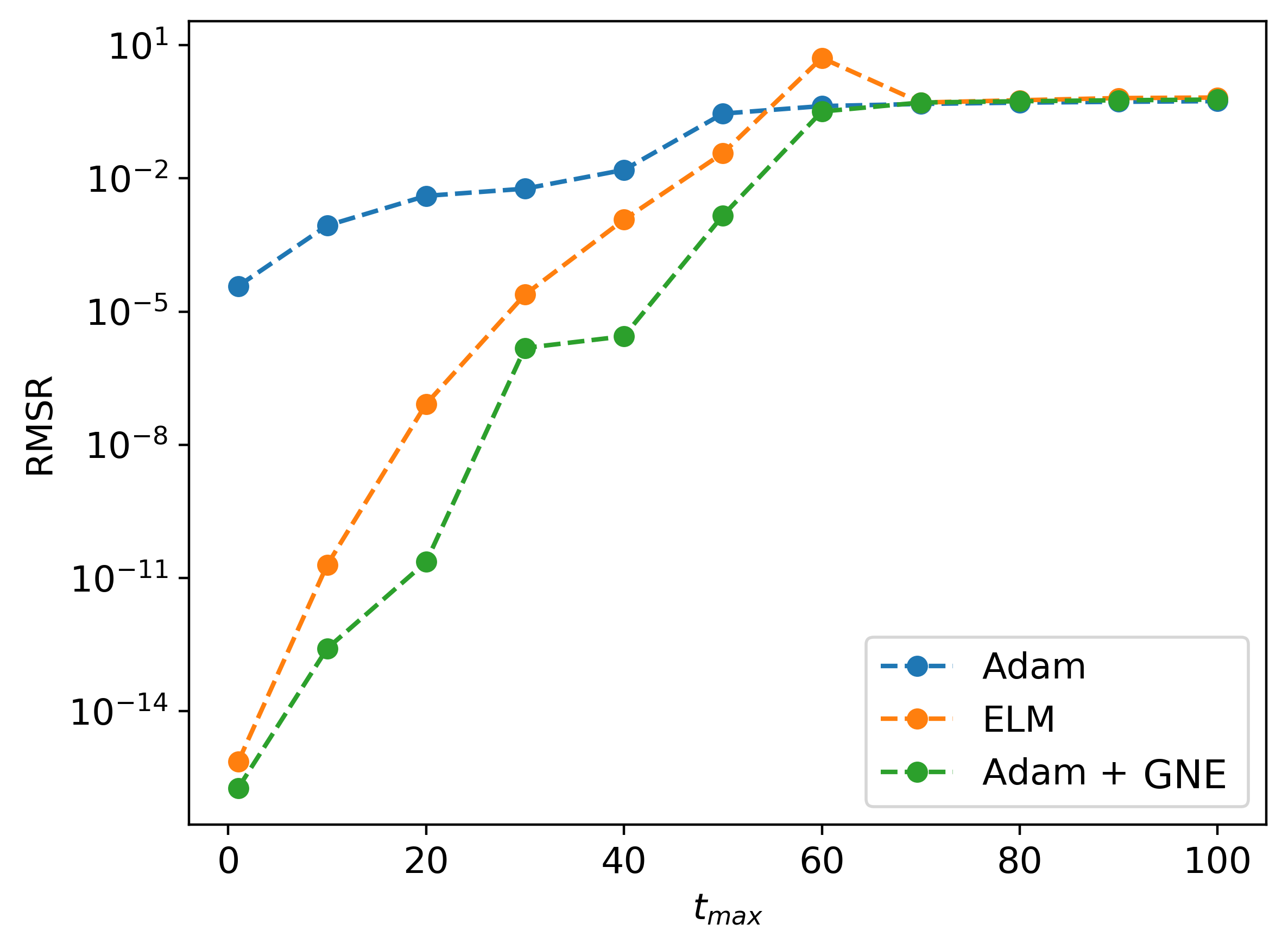}
    \includegraphics[width = 0.49 \linewidth]{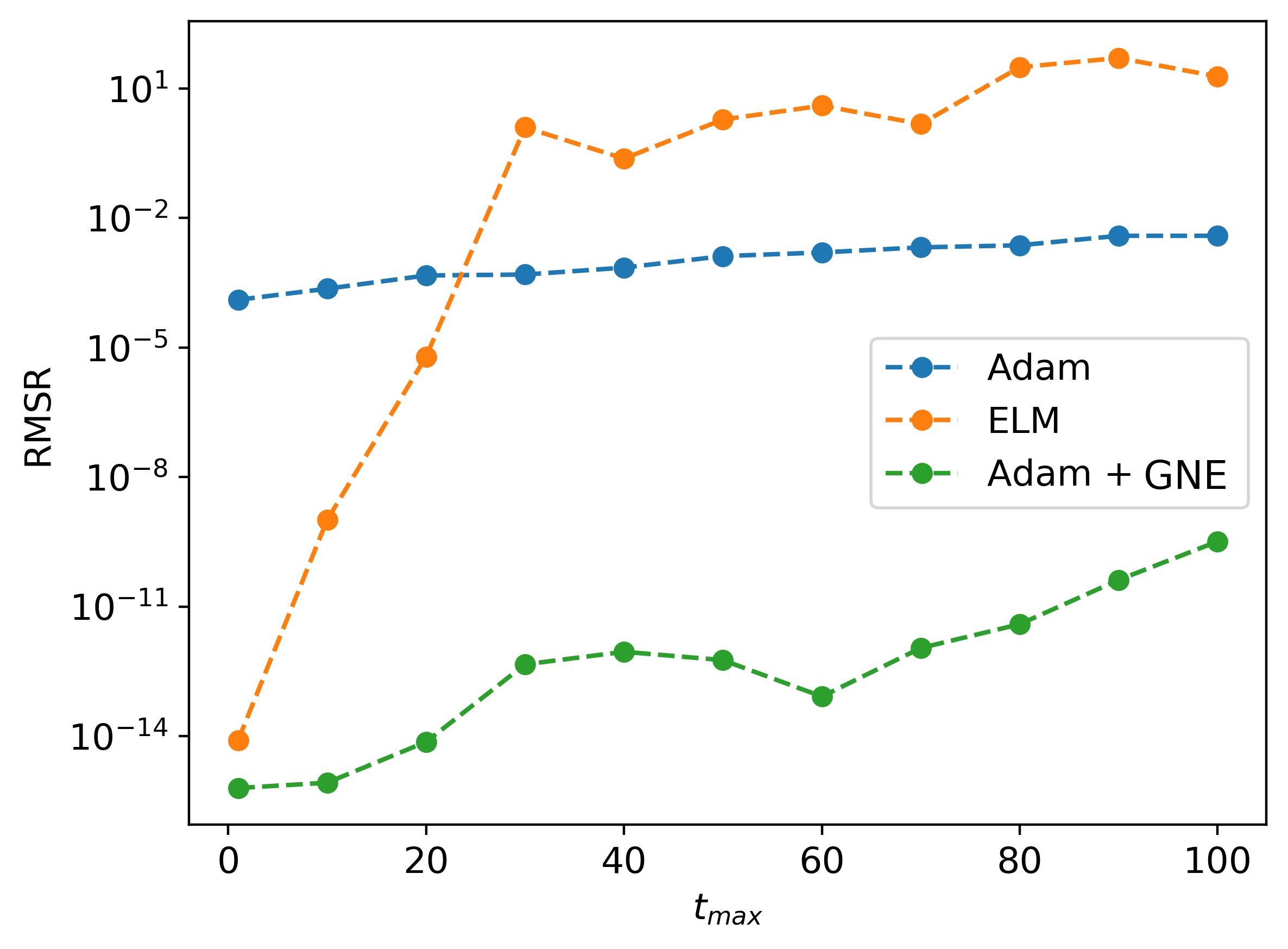}
    \caption{Root mean square of the residual of the ODE for neural networks with \textbf{(left)} 1 hidden layer with 400 neurons and \textbf{(right)} 4 hidden layers with (32, 32, 32, 400) neurons, trained on the interval $t \in [0, t_{max}]$.}
    \label{fig:sin_l2_v_l5}
\end{figure}

\begin{figure}[h]
    \centering
    \includegraphics[width = 0.49 \linewidth]{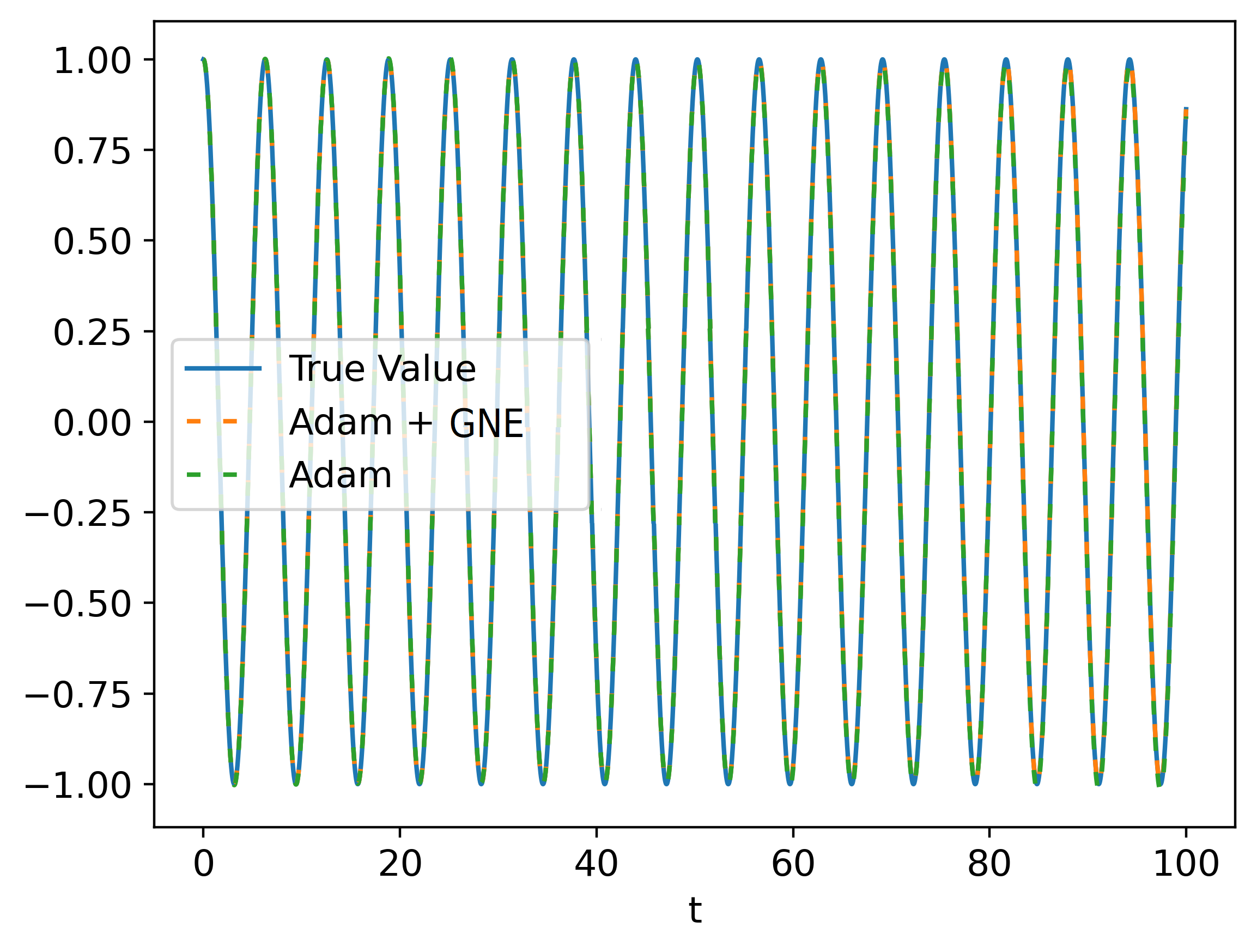}
    \includegraphics[width = 0.49 \linewidth]{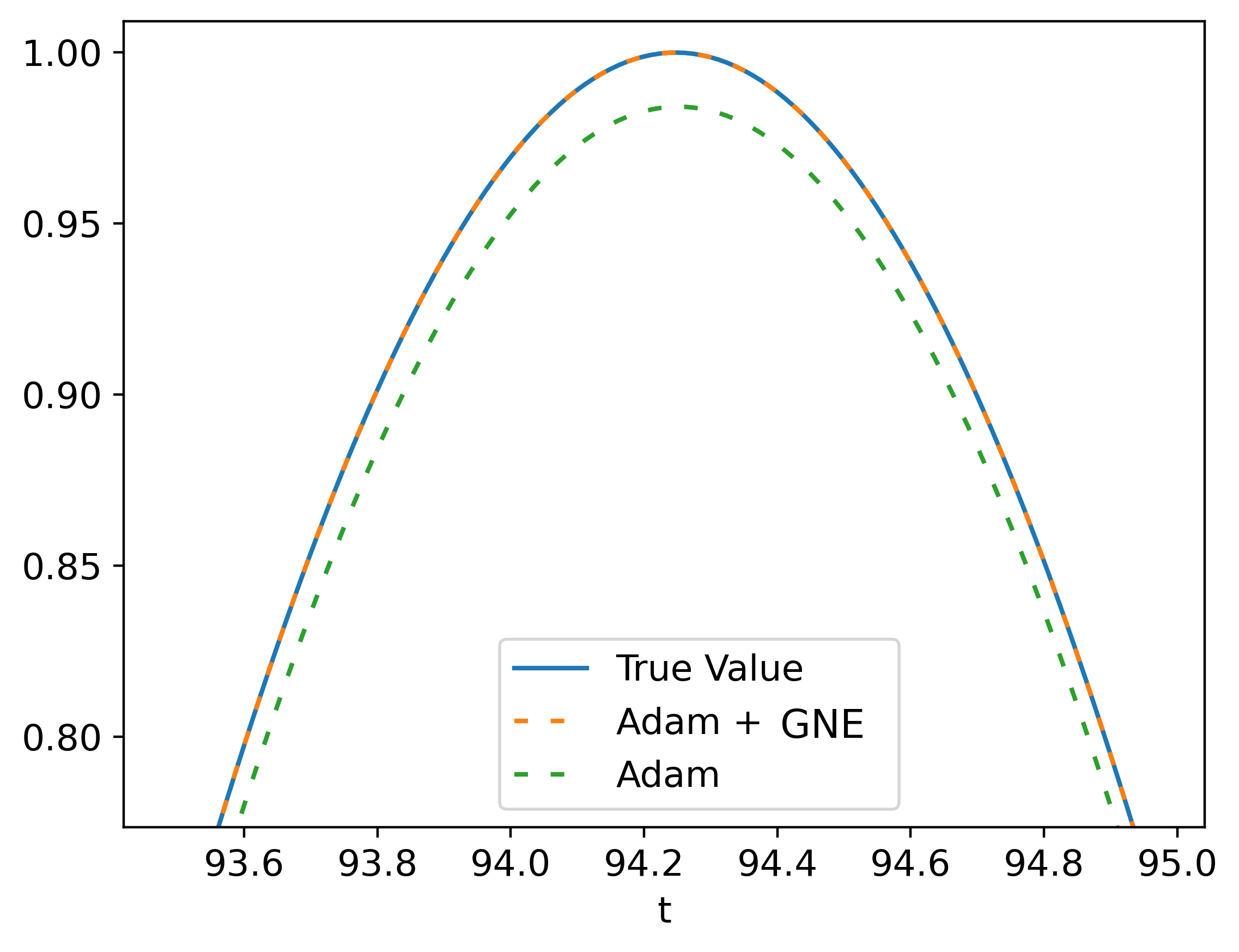}
    \caption{Comparison of the exact solution with the neural network approximation computed using a different training method on the domain $ t \in [0,100]$. The neural network had 4 hidden layers with (32, 32, 32, 400) neurons. Note that the error in the Adam solution is apparent when we zoom into a small section of the domain (right panel).}
    \label{fig:sin_compare}
\end{figure}

\begin{figure}[h]
    \centering
    \includegraphics[width = 0.49 \linewidth]{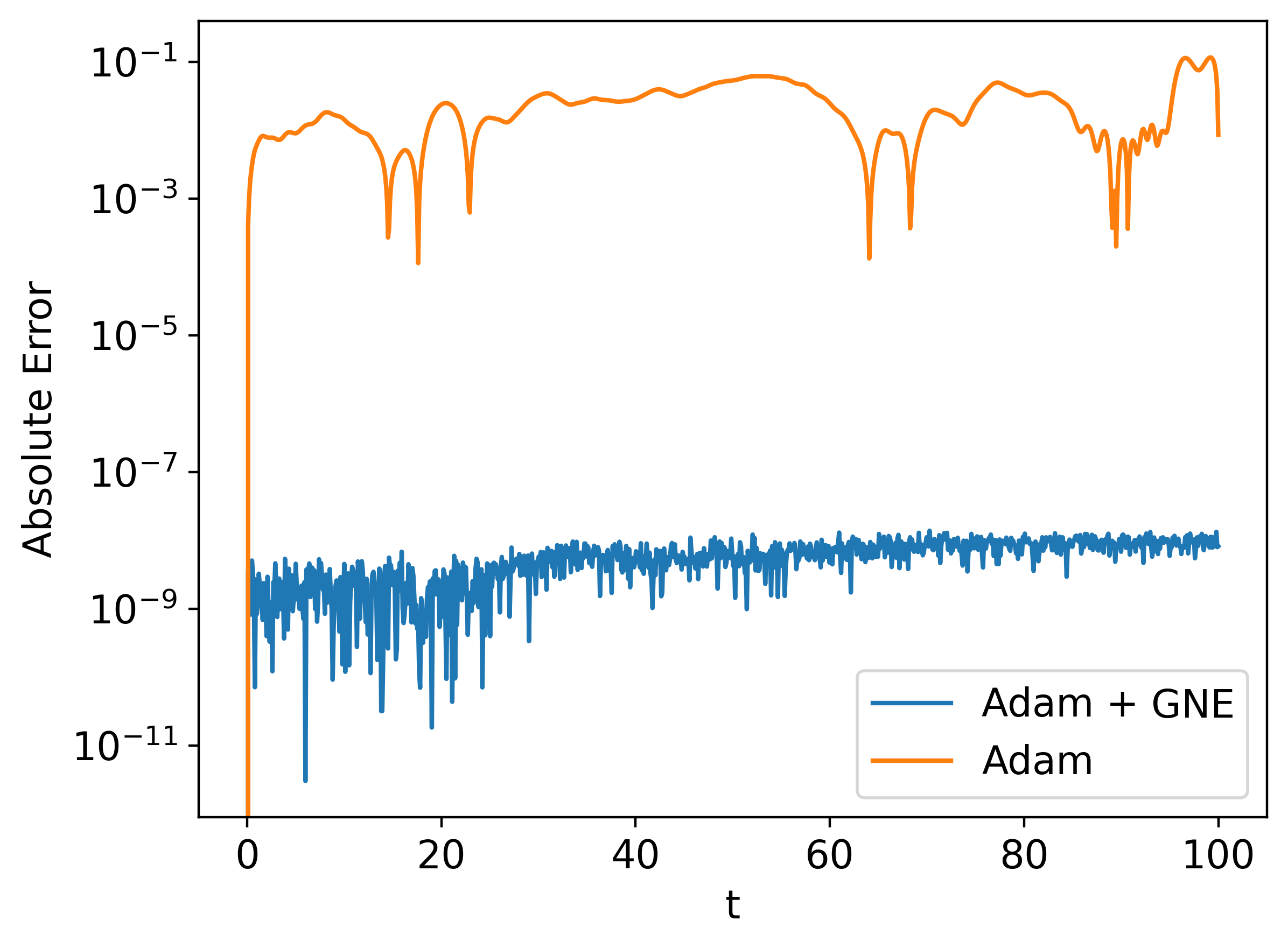}
    \caption{The absolute error in the solution after training a 4 hidden layer neural network with (32, 32, 32, 400) neurons for 100,000 steps on the domain $t \in [0,100]$.}
    \label{fig:sin_err}
\end{figure}

In Fig. \ref{fig:sin_l2_v_l5} \textbf{(left)} it can be seen that the single hidden layer neural network fails to learn the solution irrespective of the training algorithm, when the domain size becomes large. Up to 1200 neurons in the hidden layer was tried but yielded similar results. For the deep network in Fig. \ref{fig:sin_l2_v_l5} \textbf{(right)}, Adam starts performing better than ELM at around $t_{max} = 20$. This is because ELM is only optimizing the weights of the final layer of the neural network whereas with Adam we are optimizing the entire network. Now by combining Adam with GNE we create a training method which can efficiently search the large parameter space of the DNN and extremize w.r.t. the final layer to fine-tune the solution. This results in orders of magnitude improvement in the RMSR of the ODE as seen in Fig. \ref{fig:sin_l2_v_l5} \textbf{(right)}. This improvement in RMSR is reflected in the error of the solution as can be seen in Fig. \ref{fig:sin_compare} and Fig. \ref{fig:sin_err}.

\begin{figure}[h]
    \centering
    \includegraphics[width = 0.49 \linewidth]{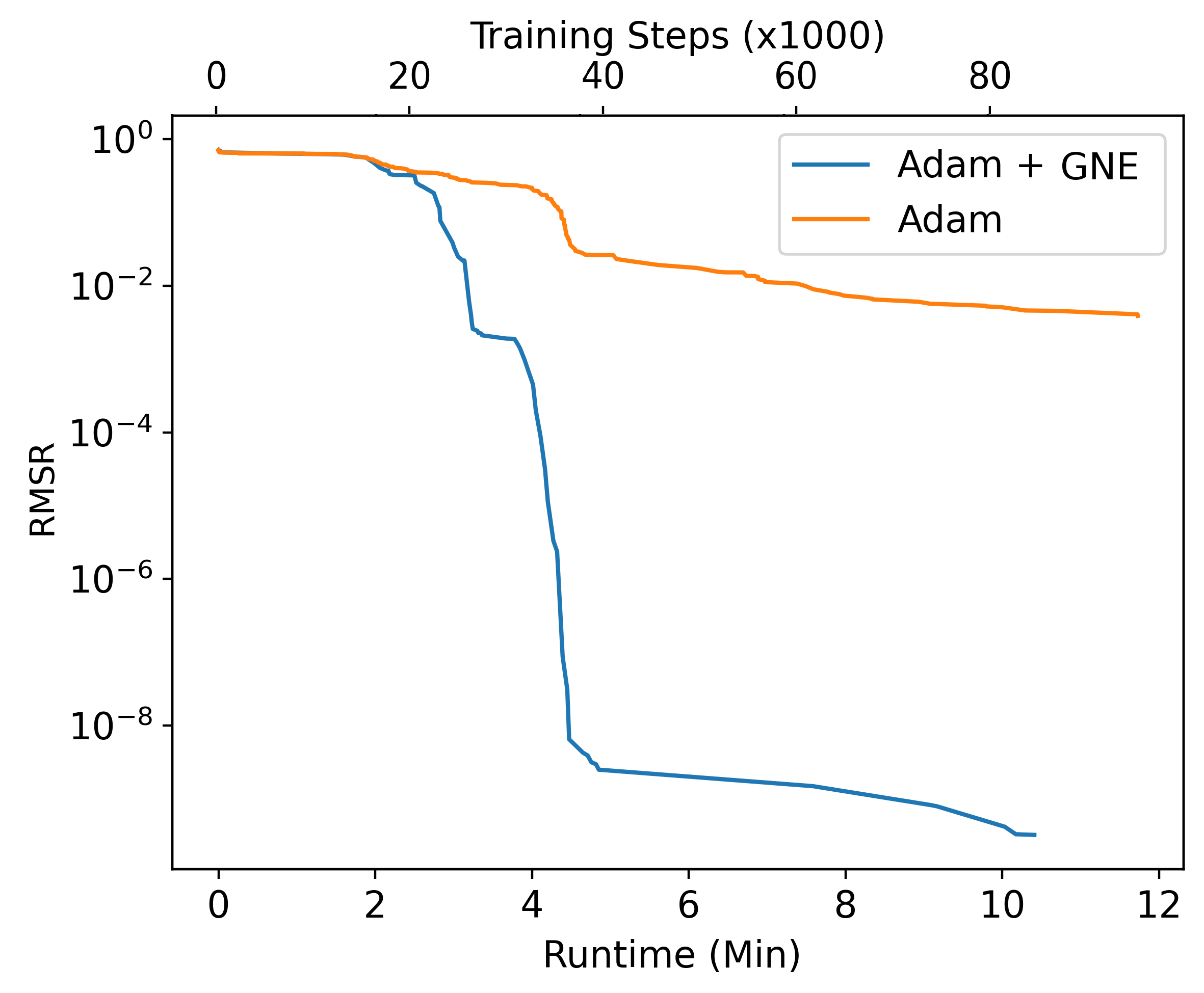}
    \caption{Learning curve of a DNN with 4 hidden layers with (32, 32, 32, 400) neurons trained on the domain $t \in [0,100]$.}
    \label{fig:sin_training}
\end{figure}

The learning curve in Fig. \ref{fig:sin_training} shows the benefit of using Adam before extremization w.r.t. the final layer. The plot was generated from the same training process where a copy of the current best model from Adam training was made at regular intervals and GNE was done on this copy. Note that the sudden drop in RMSR value for Adam at around 35,000 steps corresponds to a similar but more significant drop in RMSR for the case of Adam + GNE. The error in the numerical solution before and after using GNE is shown in Fig. \ref{fig:sin_err}.  Adam in this case helped us to efficiently search the parameter space of the DNN. After Adam learned the essential features of the solution, using GNE we were able to fine-tune the solution to significantly improve the accuracy. There is nothing special about Adam in this training process. Any optimizing algorithm that can efficiently search a high dimensional parameter space can be paired with the GNE. 

\begin{figure}[h]
    \centering
    \includegraphics[width = 0.49 \linewidth]{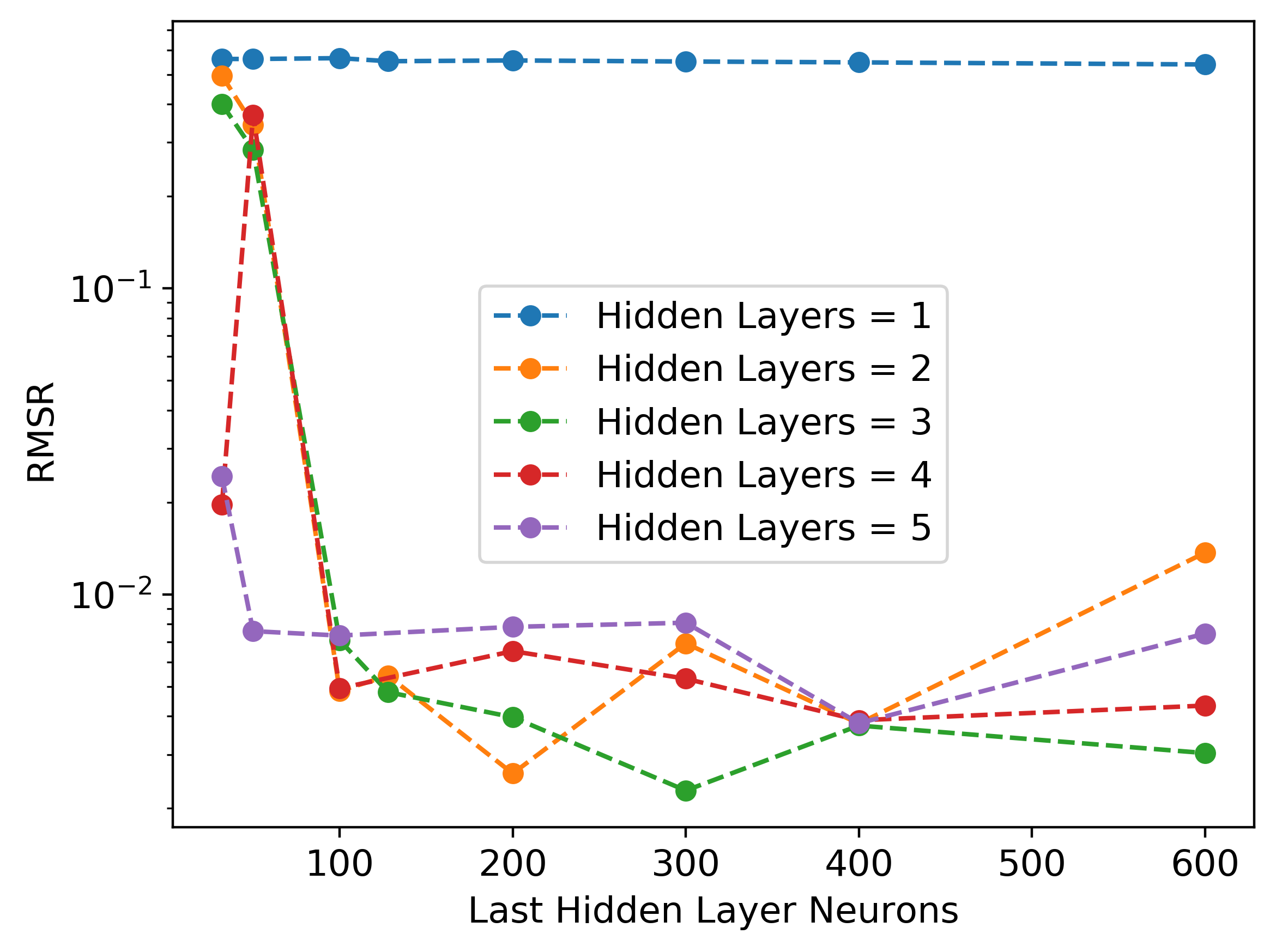}
    \includegraphics[width = 0.49 \linewidth]{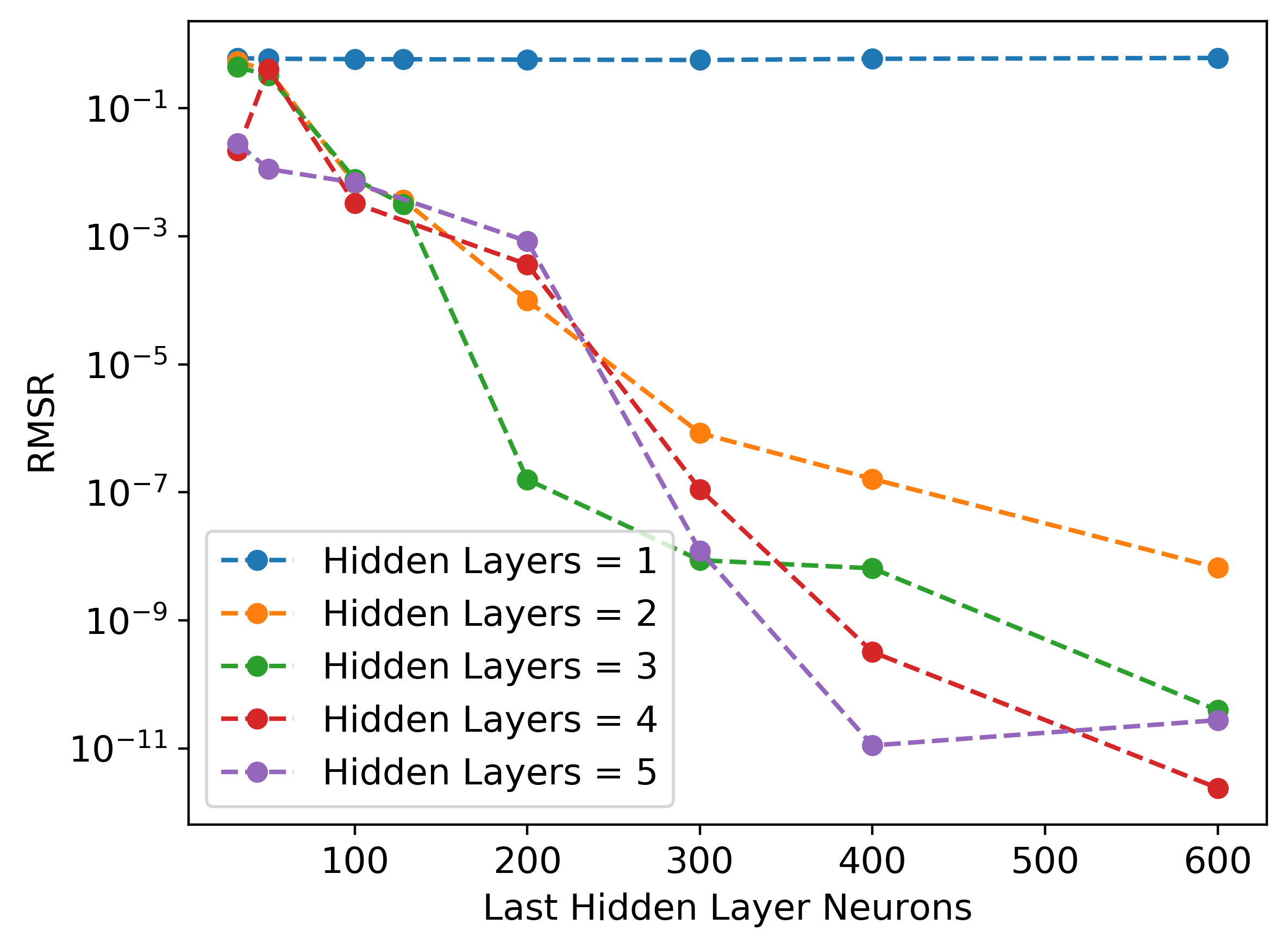}
    \caption{Root mean square of residual of ODE trained in the interval $t \in [0,100]$ with different neural networks using \textbf{(left)} Adam and \textbf{(right)} Adam + GNE. All hidden layers of neural network except the last had 32 neurons.}
    \label{fig:sin_config_compare}
\end{figure}

Fig. \ref{fig:sin_config_compare} shows the effect of the configuration of the neural network on its ability to learn the solution of the ODE. A single hidden layer network is unable to learn the solution of the ODE irrespective of the number of neurons in the hidden layer. However, adding an extra hidden layer Adam is more capable to learn the solution and GNE further refines and improves this solution. Beyond a threshold of around 100 neurons, the number of neurons in the final layer had no significant effect on the accuracy of the solution when Adam was used. As for GNE an increase in the number of neurons was associated with an increase in the accuracy of the solution. This is mainly due to the fact that with an increase in the number of neurons the extremization step has more degrees of freedom to fine-tune the solution.

Note that for PINNs, L-BFGS optimizer is generally found to be faster. It also leads to the lower RMSR compared to Adam. However, for this example L-BFGS fails to converge when the domain size becomes too large. Even though incremental training described in the next subsection can be used to mitigate this, we use Adam in this example for simplicity. Throughout the rest of this study we will exclusively use L-BFGS.

\FloatBarrier
\subsubsection{Stiff Coupled Non-linear ODEs}
\label{sec:stiff_eq}
\FloatBarrier
A system of ODEs
\begin{align}
\label{eq:stiff_ode_1}
\frac{d}{d t} u &=\cos (t)+u^2+v-\left(1+t^2+\sin ^2(t)\right), \\
\label{eq:stiff_ode_2}
\frac{d}{d t} v &=2 t-\left(1+t^2\right) \sin (t)+u v
\end{align}
has an analytical solution for the initial conditions $u(0) = 0$ and $v(0) = 1$, given by

\begin{align}
u(t) = \sin(t) \\
v(t) = 1 + t^2.
\end{align}

%I still have a problem with calling this system stiff, but let's see what referees write.
\begin{figure}[h]
    \centering
    \includegraphics[width = 0.5 \linewidth]{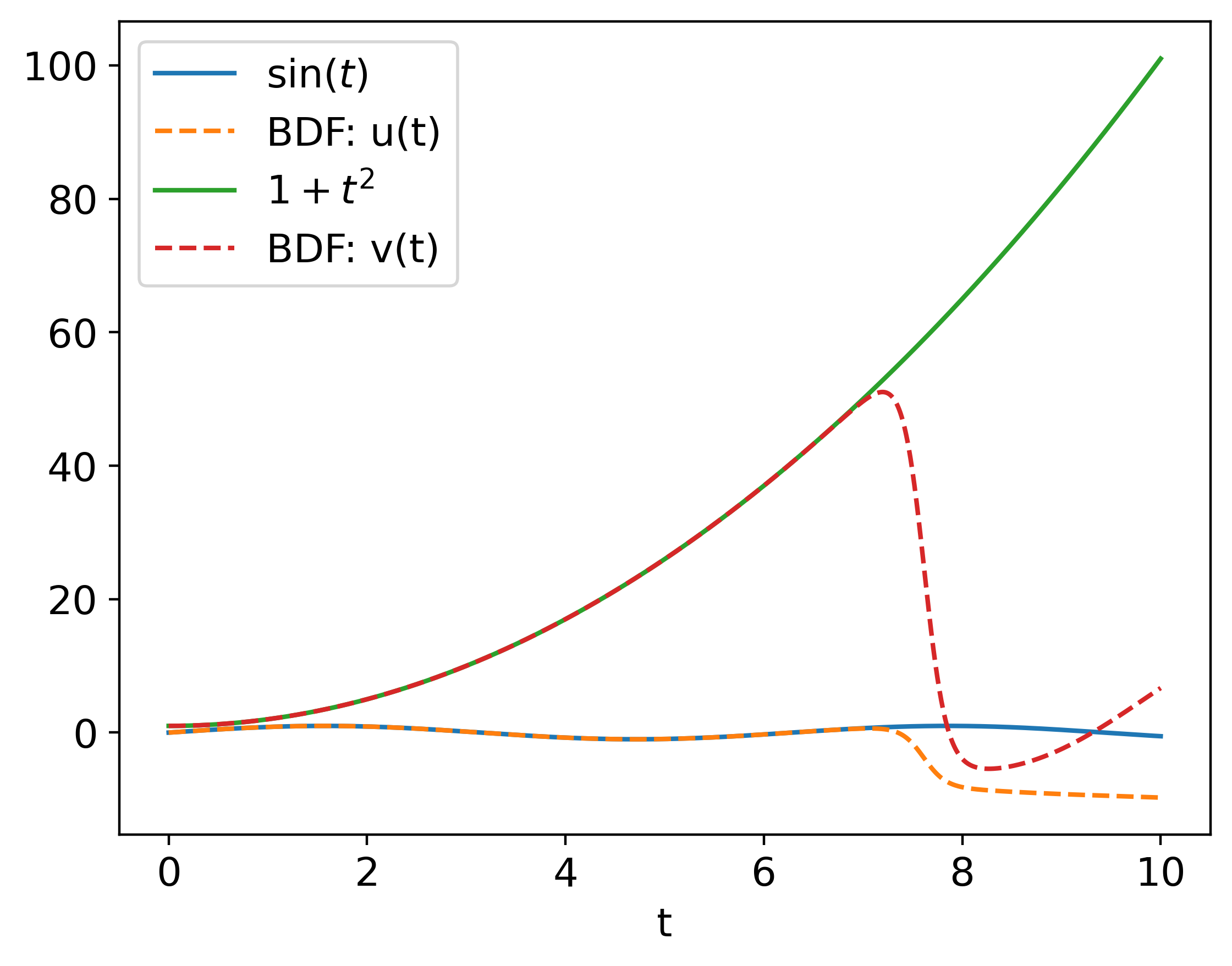}
    \caption{Comparison of exact solution with numerical solution found using BDF method for the coupled ODEs Eq. \ref{eq:stiff_ode_1} and Eq.  \ref{eq:stiff_ode_2}. The relative and absolute tolerance for BDF was set at $10^{-15}$.}
    \label{fig:bdf_fail}
\end{figure}

This system of ODEs was numerically solved in \cite{lagaris1998artificial} using a neural network on the domain $[0,3]$. On a larger domain this system exhibits stiffness and numerical instability. We tried solving this system of ODEs on the domain $[0,10]$ using several standard and commonly used ODE solvers like LSODA \cite{petzold1983automatic}, RK45 \cite{dormand1980family}, DOP853 \cite{hairer1993solving} and BDF \cite{shampine1997matlab} implemented in \textit{solve\_ivp} function of \textit{scipy} package. All these solvers failed to solve the system of ODEs in different ways. For instance in the case of LSODA the step size became too small irrespective of the prescribed tolerance values. RK45 finds an incorrect solution when absolute and relative tolerances are greater than $10^{-10}$. For smaller tolerance values RK45 step size effectively becomes 0 and integration fails. A similar effect, although with smaller tolerance values, is observed with DOP853 method. BDF, which is capable of providing numerical solutions for some stiff problems, with relative and absolute tolerance set at $10^{-15}$ gave an incorrect solution as shown in Fig. \ref{fig:bdf_fail}. 

Different training algorithms for PINNs also failed to converge when directly trained on the full domain $[0,10]$. Instead we start with a smaller domain $[0,0.5]$ and increase the domain size during training when certain conditions are met. For ELM, since the equation under consideration is non-linear, GNE iterations are repeated until the RMSR stops decreasing, then we increase the domain size by $0.5$. As for L-BFGS we start with $[0,0.5]$ domain and increase the domain size by $0.5$ when the RMSR becomes smaller than $5 \times 10^{-2}$. This process is repeated until the full domain size $[0,10]$ is reached. Then the training continues without any domain size increments. We call this incremental training. 1000 random uniformly sampled points were chosen from the domain for each L-BFGS training step irrespective of the domain size. 4000 random uniformly sampled points were used for GNE in ELM and after L-BFGS.

\begin{figure}[h]
    \centering
    \includegraphics[width = 0.5 \linewidth]{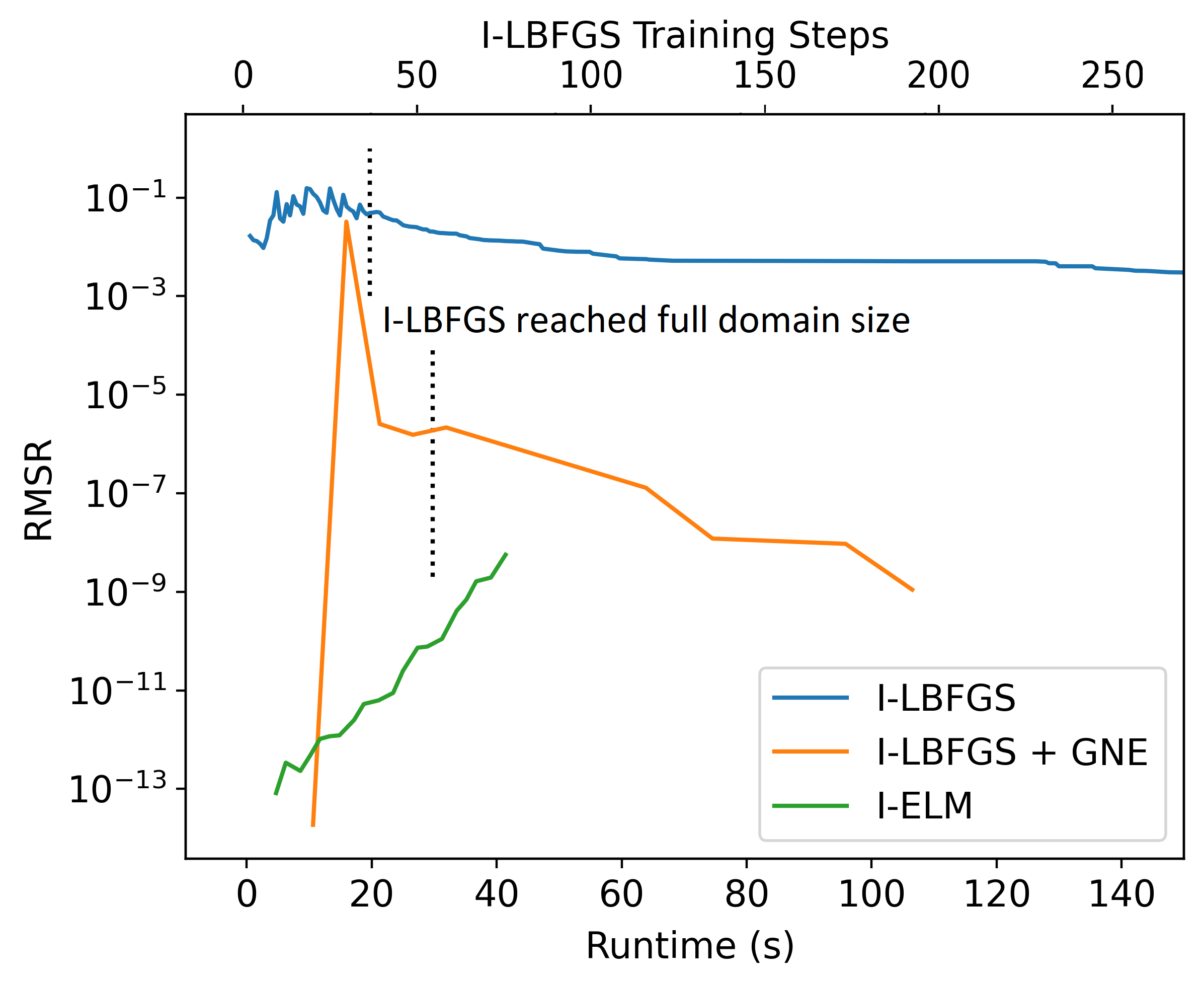}
    \caption{Evolution of RMSR with incremental training for different training algorithms. The black vertical dotted lines denote the point when L-BFGS reaches full domain size $[0,10]$ with incremental training. This line appears at different locations on the I-LBFGS and I-LBFGS + GNE curves since the latter requires additional time to perform GNE after I-LBFGS reaches full domain size. ELM had a single hidden layer with 400 neurons. L-BFGS optimized a neural network with 2 hidden layers with (32,400) neurons. }
    \label{fig:stiff_rmsr}
\end{figure}

\begin{figure}[h]
    \centering
    \includegraphics[width = 0.49 \linewidth]{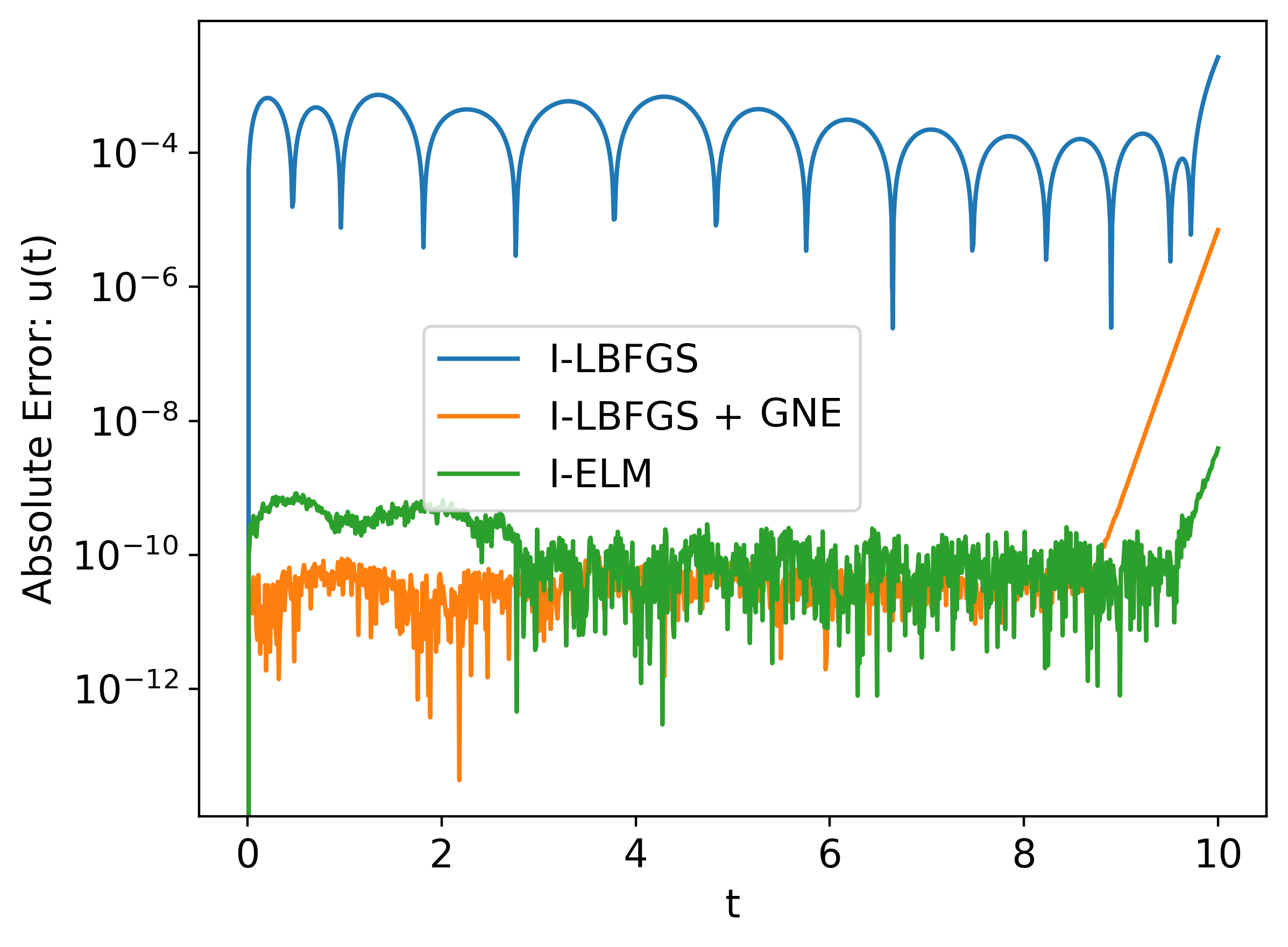}
    \includegraphics[width = 0.49 \linewidth]{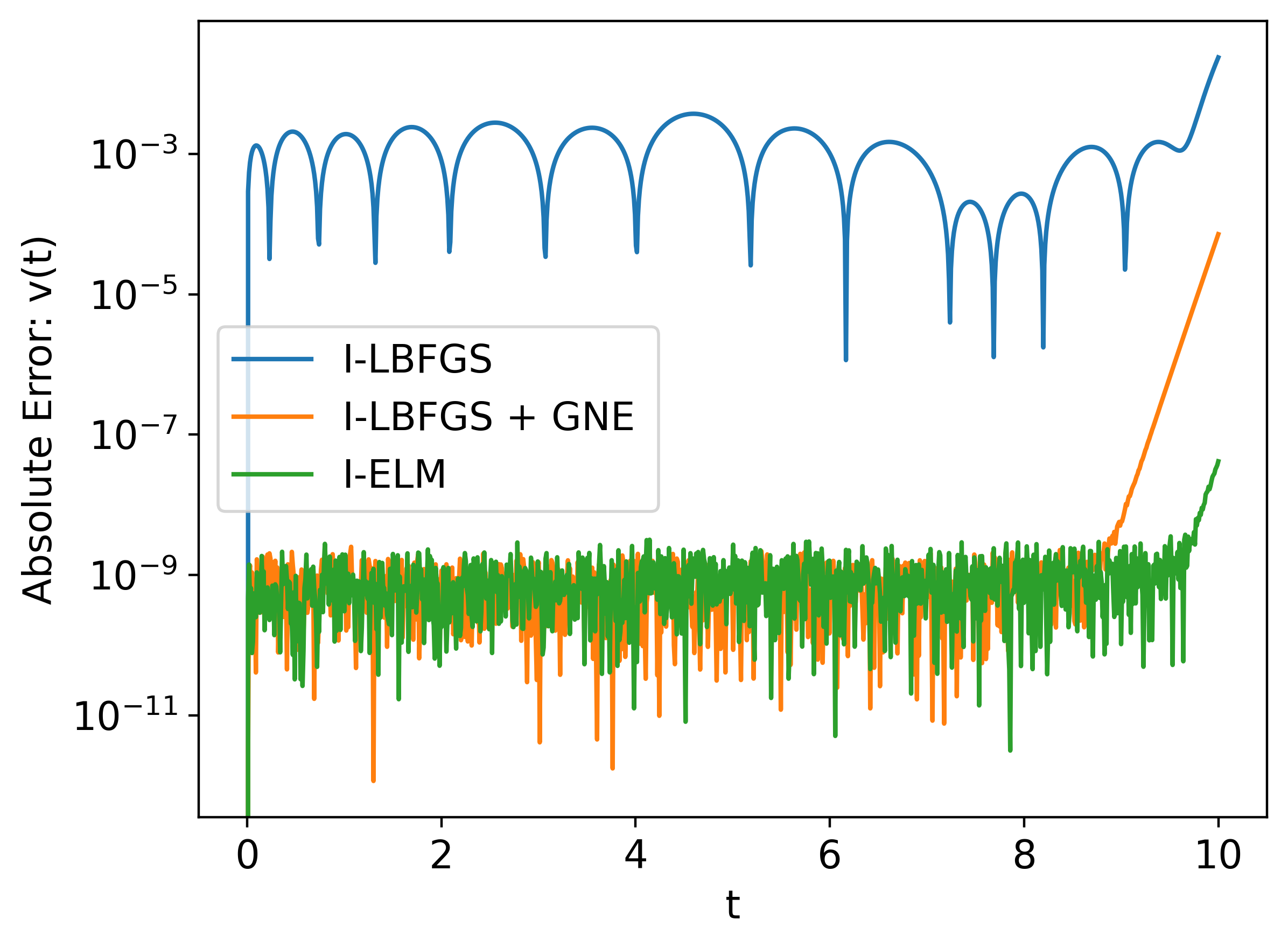}
    \caption{The absolute error in the solution found using different incremental training methods on the domain $t \in [0,10]$. ELM had a single hidden layer with 400 neurons. L-BFGS optimized a neural network with 2 hidden layers with (32, 400) neurons.}
    \label{fig:stiff_error}
\end{figure}

\begin{figure}[h]
    \centering
    \includegraphics[width = 0.49 \linewidth]{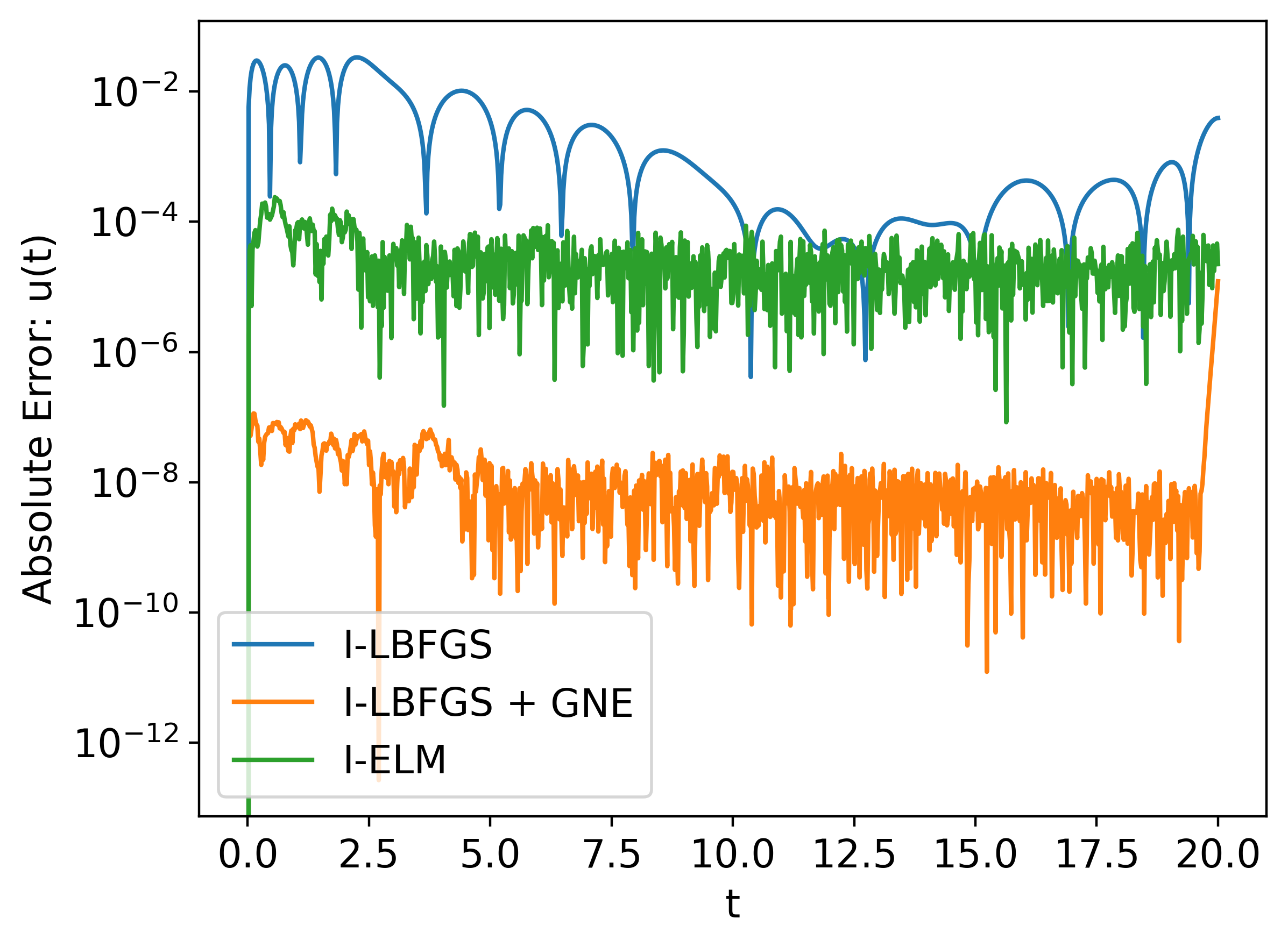}
    \includegraphics[width = 0.49 \linewidth]{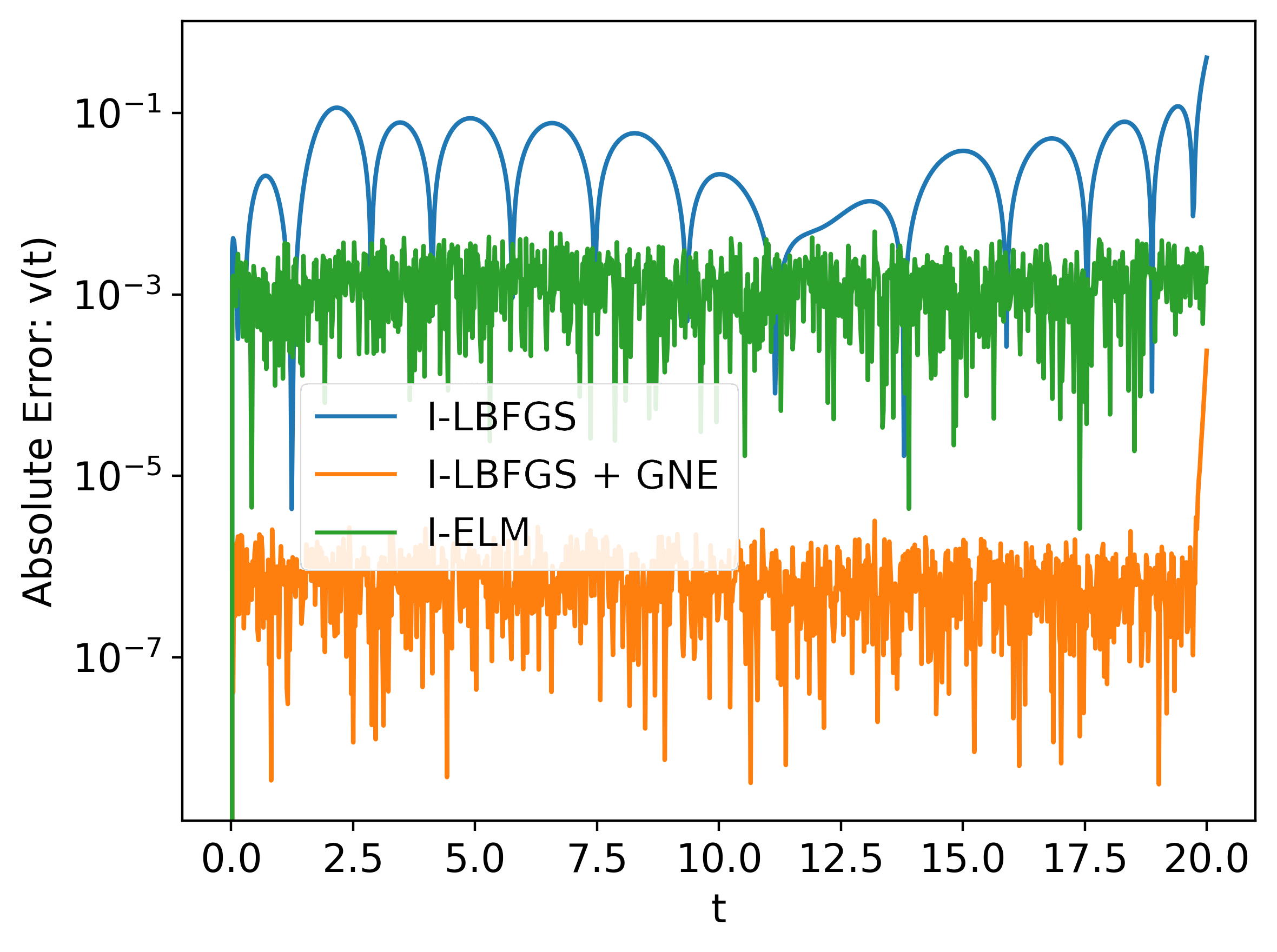}
    \caption{The absolute error in the solution found using different incremental training methods on the domain $t \in [0,20]$. ELM had a single hidden layer with 400 neurons. L-BFGS optimized a neural network with 3 hidden layers with (32, 32, 400) neurons.}
    \label{fig:stiff_error_t_20}
\end{figure}

Fig. \ref{fig:stiff_rmsr} shows the training curve for incremental training with different training methods. The initial fluctuation in the RMSR values are due to change in domain size. The L-BFGS method reaches full domain size faster than ELM, but ELM solution has significantly lower RMSR. The L-BFGS + GNE method takes about twice the amount of time as compared to ELM to achieve marginally lower RMSR value. The corresponding absolute errors in the solutions are shown in Fig. \ref{fig:stiff_error}. Note that the increase in error towards the end of the domain is not a trend that would extrapolate if we increase the domain size. For this specific system of ODEs this trend of increasing error towards the end of the domain is seen irrespective of the domain  $[0,t_{max}]$, where $t_{max} \gtrsim 8$. Fig. \ref{fig:stiff_error_t_20} shows the error in the solution for neural networks trained on a larger domain $[0,20]$.

Similar to the example given in Sec. \ref{sec:sin_eq}, the advantage of deep network trained with incremental L-BFGS + GNE is seen when we further increase the domain size to $[0,20]$ as shown in Fig. \ref{fig:stiff_error_t_20}. Using more neurons in the hidden layer resulted in higher RMSR for incremental ELM.

\FloatBarrier
\subsection{PDE}
\label{sec:pinn_pde}
\subsubsection{2D Linear and Non-Linear PDE}
\label{sec:l_v_nl}
\FloatBarrier
The linear PDE we use in this subsection is given by
\begin{equation}
    \nabla^2 u(x, y)=\left(2-\pi^2 y^2\right) \sin (\pi x)
\end{equation}
and the non-linear PDE is
\begin{equation}
    \label{eq:nl_pde_1}
    \nabla^2 u(x, y)+u(x, y) \frac{\partial}{\partial y} u(x, y)=\sin (\pi x)\left(2-\pi^2 y^2+2 y^3 \sin (\pi x)\right)
\end{equation}
We use the same set of mixed boundary conditions on both equations 
\begin{align}
u(0, y) &=0 \\
u(1, y) &=0 \\
u(x, 0) &=0 \\
\frac{\partial}{\partial y} u(x, 1) &=2 \sin (\pi x)
\end{align}
and both PDEs have the same analytic solution
\begin{equation}
u(x, y)=y^2 \sin (\pi x).
\end{equation}

For ELM we use a single hidden layer with 200 neurons and for the deep network we use 2 hidden layers with (32, 200) neurons. Decreasing the number of neurons in the last hidden layer decreased the RMSR value while increasing only leads to marginal improvement in RMSR. Increasing the number of hidden layers seemed not to affect the RMSR but increased the computational time. We use 1000 random uniformly sampled points from the domain for each L-BFGS training step and 2000 random uniformly sampled points for the GNE in ELM and after L-BFGS training.

For the linear equation ELM required a single iteration of GNE and the solution took 1.5~s to compute. In the case of L-BFGS + GNE, 2 steps of L-BFGS took around 2~s and an additional 1.5~s for GNE taking a total of 3.5~s to compute the solution. In the case of non-linear equation GNE involves multiple iterations. The ELM solution took around 8~s. For the hybrid training 3 L-BFGS steps took around 3.5~s and the GNE took 19~s taking a total of 22.5~s to compute. For both these cases pure L-BFGS RMSR and mean absolute error were around $10^{-1}$. We need to train for significantly longer duration with pure L-BFGS to attain an absolute error which is orders of magnitude worse than ELM and L-BFGS + GNE.

\begin{table}[h]
\centering
\resizebox{\textwidth}{!}{
\begin{tabular}{|l|l|l|l|l|l|l|} 
\hline
 & \multicolumn{3}{|c|}{\textbf{ELM}} & \multicolumn{3}{c|}{\textbf{L-BFGS + GNE}}\\
\hline
 & RMSR & Mean Abs. Err. & Max. Abs. Err.  & RMSR & Mean Abs. Err. & Max. Abs. Err. \\
 \hline
Linear & $9.2 \times 10^{-10}$ & $2.1 \times 10^{-12}$ & $1.2 \times 10^{-11}$ & $1.0 \times 10^{-9}$ & $1.6 \times 10^{-12}$ & $8.1 \times 10^{-12}$ \\
Non-Linear & $3.7 \times 10^{-7}$ & $6.0 \times 10^{-10}$ & $4.2 \times 10^{-9}$ & $1.5 \times 10^{-9}$ & $4.2 \times 10^{-12}$ & $2.3 \times 10^{-11}$ \\
\hline
\end{tabular}
}
\caption{Statisitcs of numerical solution for linear and non-linear PDE found using ELM and L-BFGS + GNE. ELM had a single hidden layer with 200 neurons. L-BFGS optimized a neural network with 3 hidden layers with (32, 32, 400) neurons.}
\label{tab:l_v_nl}
\end{table}

 Even though the analytic solution which the neural network has to learn is the same for both the PDEs, ELM performed worse with the non-linear PDE compared to the linear PDE according to Table \ref{tab:l_v_nl}. In comparison, for L-BFGS + GNE the solution of linear and non-linear PDEs had similar statistics. In the case of the linear PDE, a LLS problem w.r.t. weights in the last layer of the neural network can be extremized in a single iteration of GNE. However for the non-linear PDE a more complicated optimization landscape is found where combining L-BFGS with GNE is advantageous compared to just using GNE in the case of ELM. Note that in the case of the non-linear PDE solved with ELM the LLS algorithm used to perform the Gauss-Newton iteration is seen to significantly affect the numerical solution. This is discussed further in \ref{app:lstsq}.
\FloatBarrier
\subsubsection{1+1 D Burgers' Equation}
\label{sec:burgers_eq}
\FloatBarrier

\begin{figure}[h]
    \centering
    \includegraphics[width = 0.32 \linewidth]{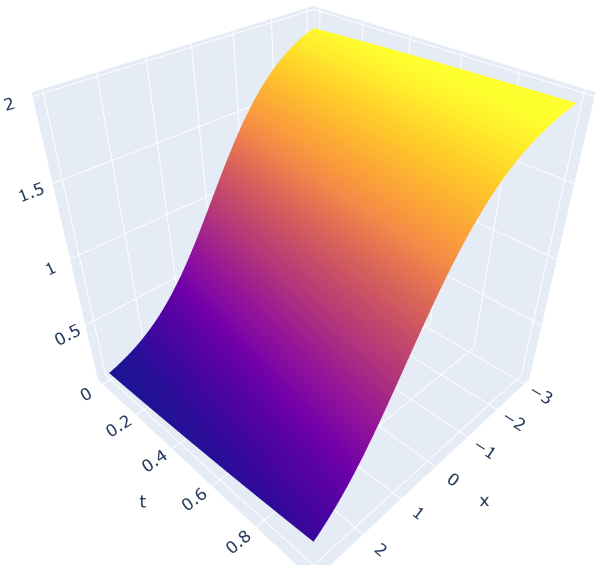}
    \includegraphics[width = 0.32 \linewidth]{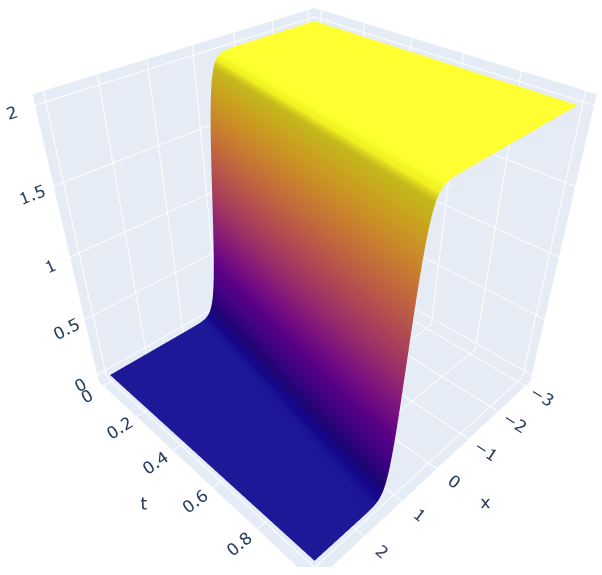}
    \includegraphics[width = 0.32 \linewidth]{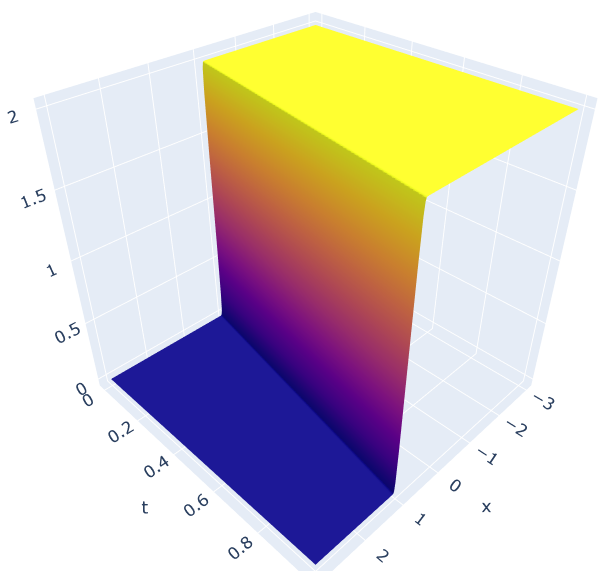}
    \caption{Plot of analytic solution (Eq. \ref{eq:burgers_analytic}) of Burgers' equation for \textbf{(left)} $\nu = 1$, \textbf{(center)} $\nu = 0.1$ and \textbf{(right)} $\nu = 0.01$}
    \label{fig:burgers_nu_grad}
\end{figure}

The (viscous) Burgers' equation is a non-linear PDE given by the following equation:
\begin{equation}
\label{eq:burgers}
    \frac{\partial u}{\partial t}+\alpha u \frac{\partial u}{\partial x}=\nu \frac{\partial^2 u}{\partial x^2}.
\end{equation}

A previous study \cite{schiassi2021extreme} addressed solving Burgers' equation using ELM and DNN. In order to compare with their results, we solve the PDE on the same domain $(x,t) \in [-3,3] \times [0,1]$ using the same boundary conditions:
\begin{align}
u(-3, t)&=\frac{c}{\alpha}-\frac{c}{\alpha} \tanh \left(\frac{c}{2 \nu}(-3-c t)\right) \\
u(3, t)&=\frac{c}{\alpha}-\frac{c}{\alpha} \tanh \left(\frac{c}{2 \nu}(3-c t)\right) \\
u(x, 0)&=\frac{c}{\alpha}-\frac{c}{\alpha} \tanh \left(\frac{c}{2 \nu} x\right).
\end{align}
The exact analytic solution for this boundary value problem is given by
\begin{equation}
\label{eq:burgers_analytic}
    u(x, t)=\frac{c}{\alpha}-\frac{c}{\alpha} \tanh \left(\frac{c}{2 \nu}(x-c t)\right).
\end{equation}
In this case the viscosity parameter $\nu$ plays a role in determining the sharpness of the gradients in the exact solution:

\begin{align}
    & \nabla u(x, t)=\left( \frac{c^2}{\alpha \nu + \alpha \nu \cosh \left(\frac{c}{2 \nu}(x-c t)\right)}, -\frac{c^3}{\alpha \nu + \alpha \nu \cosh \left(\frac{c}{2 \nu}(x-c t)\right)}   \right)\\
    \implies & \left. \nabla u(x, t)\right|_{x=ct} = \left( \frac{c^2}{2 \alpha \nu }, -\frac{c^3}{2 \alpha \nu }   \right).
\end{align}

The magnitude of the gradient at $x = c t $ is inversely proportional to $\nu$ and as $\nu \to 0$ the analytic solution becomes discontinuous. This can be readily seen in Fig. \ref{fig:burgers_nu_grad}. Since all the optimization methods we use in this work rely on computing the gradient terms in Eq. \ref{eq:burgers} and further computing gradients of the residual of Eq. \ref{eq:burgers} w.r.t. the weights in the neural network, we expect PINN to perform worse as $\nu$ becomes small. In this subsection we will look at how robust the proposed method is to large gradients when training.

\begin{figure}[h]
    \centering
    \includegraphics[width = 0.49 \linewidth]{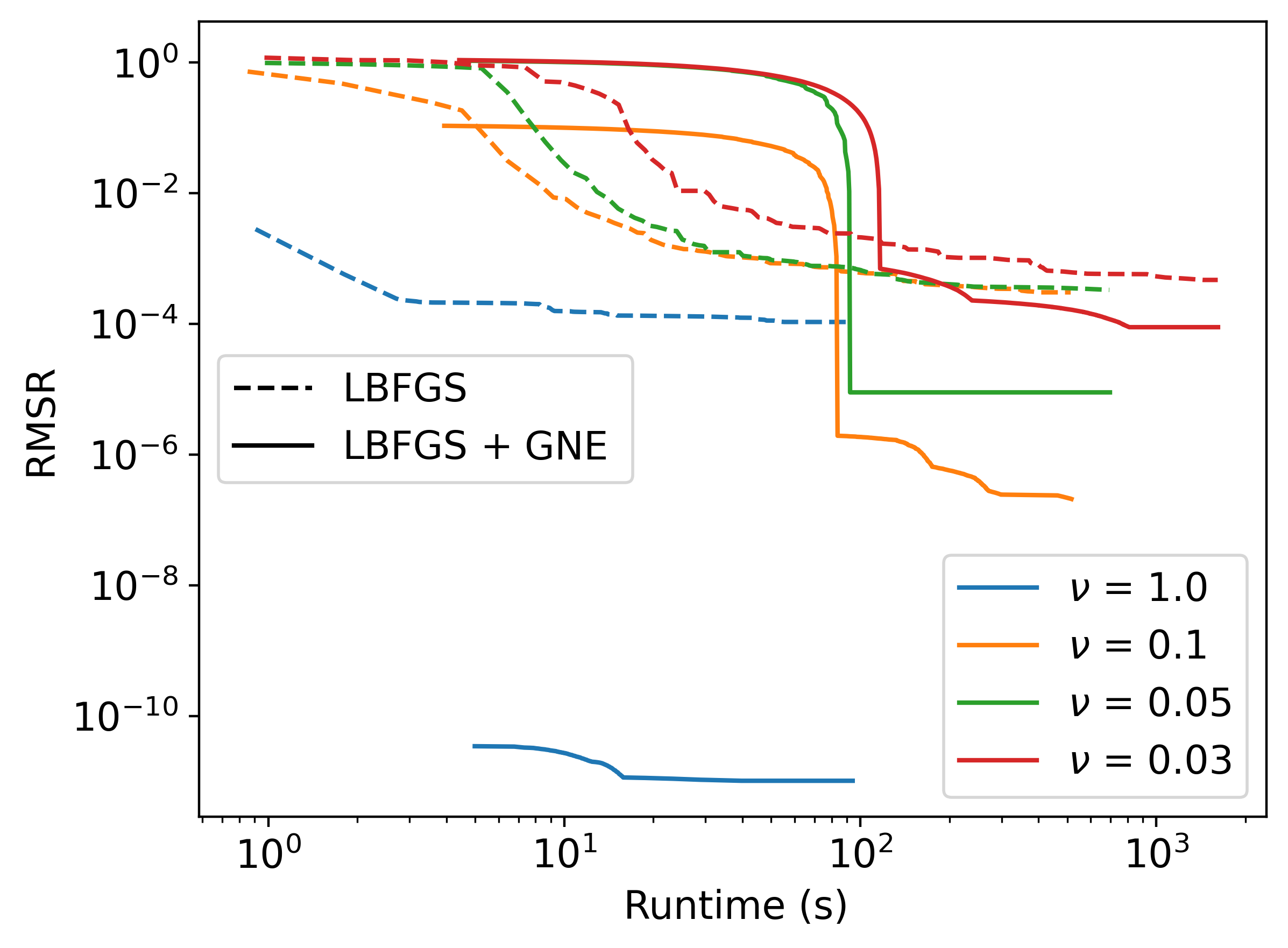}
    \caption{Learning curve of neural networks using L-BFGS (dashed curves) and L-BFGS + GNE (solid curves) for the Burgers' equation with different values of $\nu$. All the neural networks were trained for 15,000 L-BFGS steps. ELM took between 15~s to 25~s depending on the value of $\nu$.}
    \label{fig:burgers_lc}
\end{figure}

We use similar architecture as used in \cite{schiassi2021extreme} and set $\alpha = c = 1$. ELM had a single hidden layer with 601 neurons. The deep network had 4 hidden layers with (32, 32, 32, 200) neurons. Each L-BFGS step was trained on 1000 random uniformly sampled points from the domain and GNE in ELM and after L-BFGS was done with 2000 random uniformly sampled points from the domain. All L-BFGS training in this subsection with and without GNE was done for 15,000 steps for consistency. Most of the neural networks converged to the lowest RMSR well before this, as shown in Fig. \ref{fig:burgers_lc}.

\begin{figure}[h]
    \centering
    \includegraphics[width = 0.49 \linewidth]{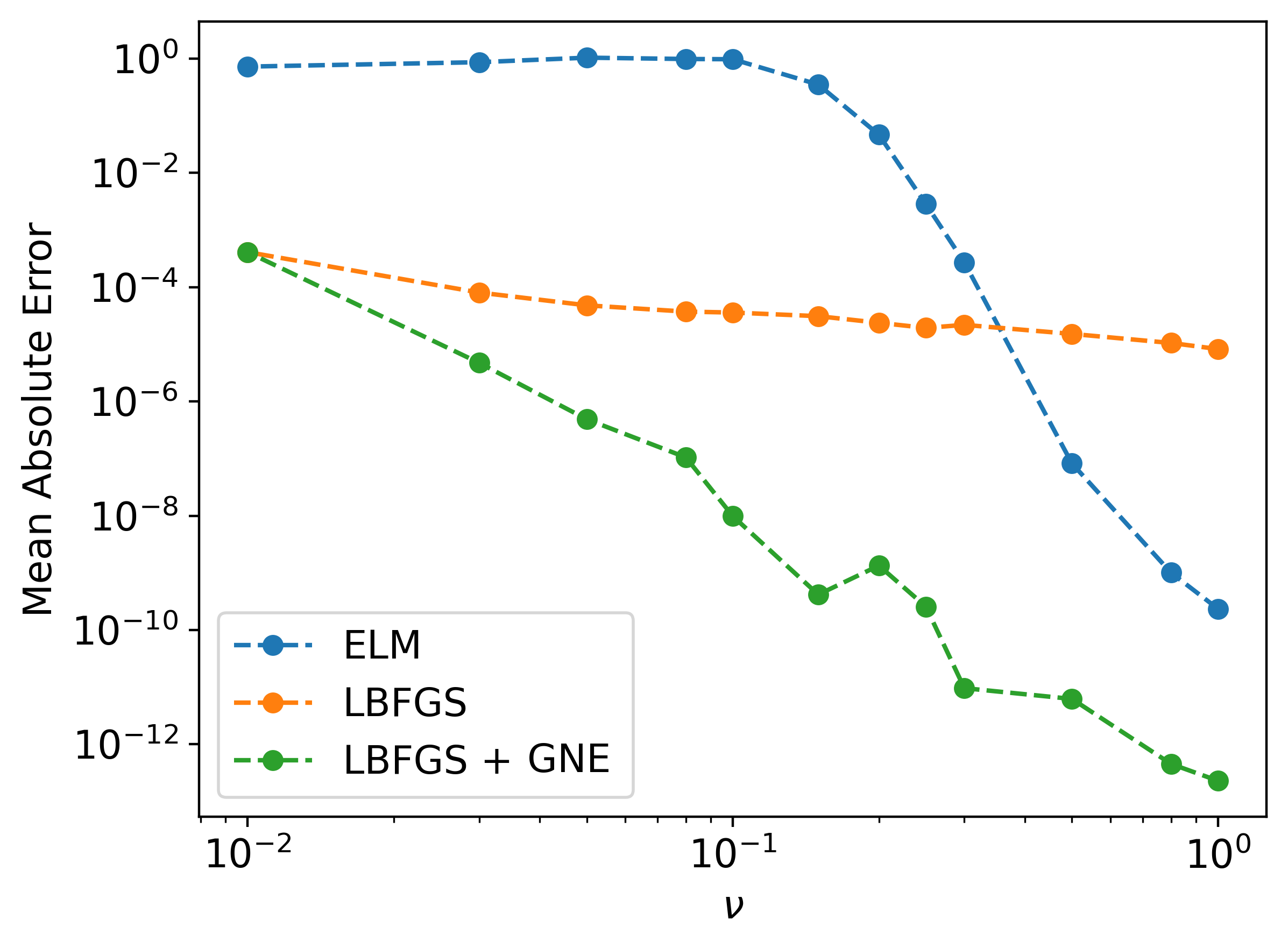}
    \includegraphics[width = 0.49 \linewidth]{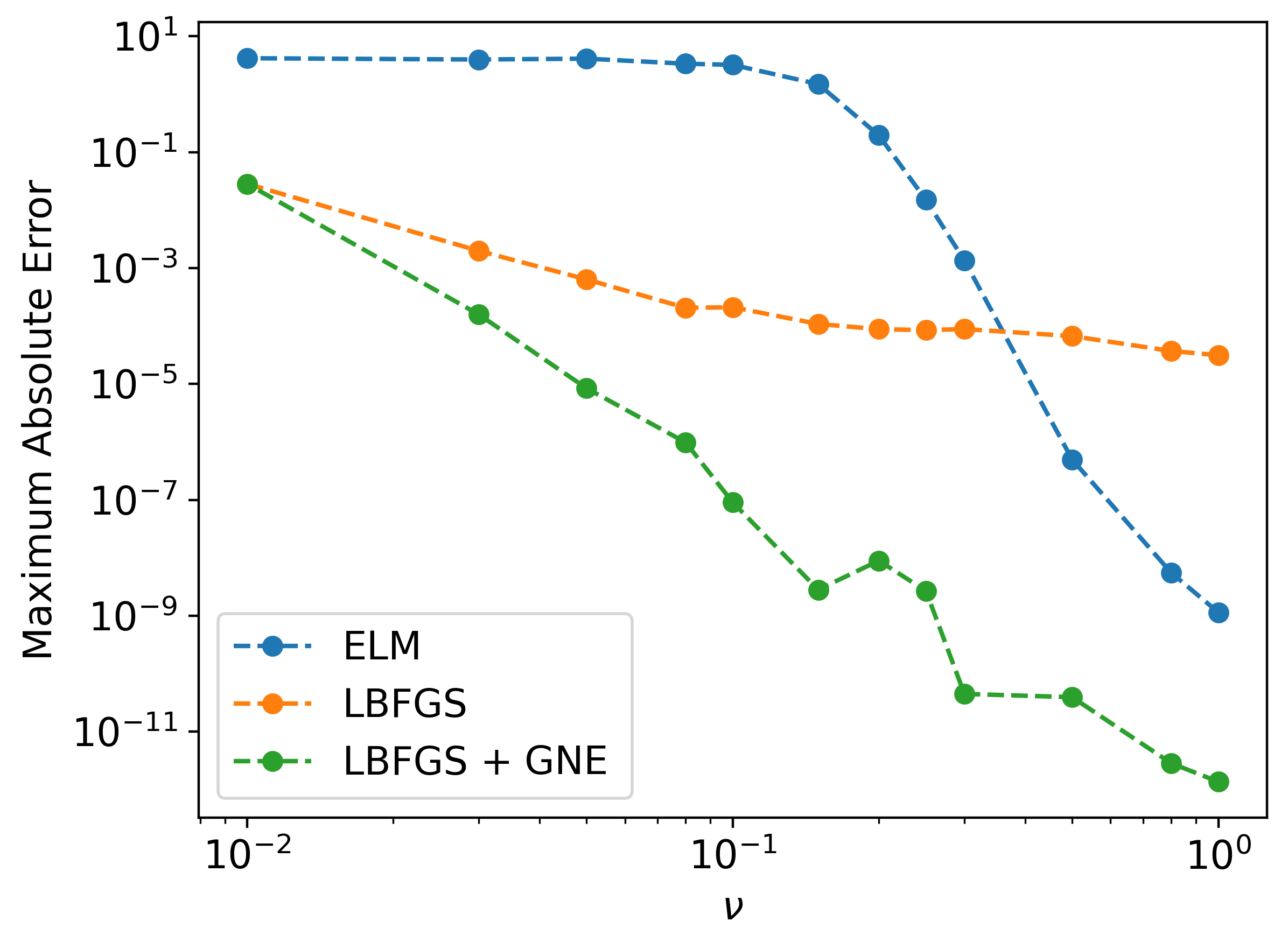}
    \caption{\textbf{(left)} Mean absolute error and \textbf{(right)} maximum absolute error for Burgers' equation solved with different training algorithms.}
    \label{fig:burgers_err}
\end{figure}

From Fig. \ref{fig:burgers_err} it can be seen that for $\nu \lesssim 0.2$ ELM starts performing worse than DNN trained with L-BFGS. In comparison L-BFGS + GNE consistently performs better than both ELM and L-BFGS. For $\nu \lesssim 0.01$ it was observed that applying GNE on a model trained with L-BFGS led to similar or worse solution error. This is likely due to GNE being sensitive to large gradients. 

\begin{figure}[h]
    \centering
    \includegraphics[width = 0.32 \linewidth]{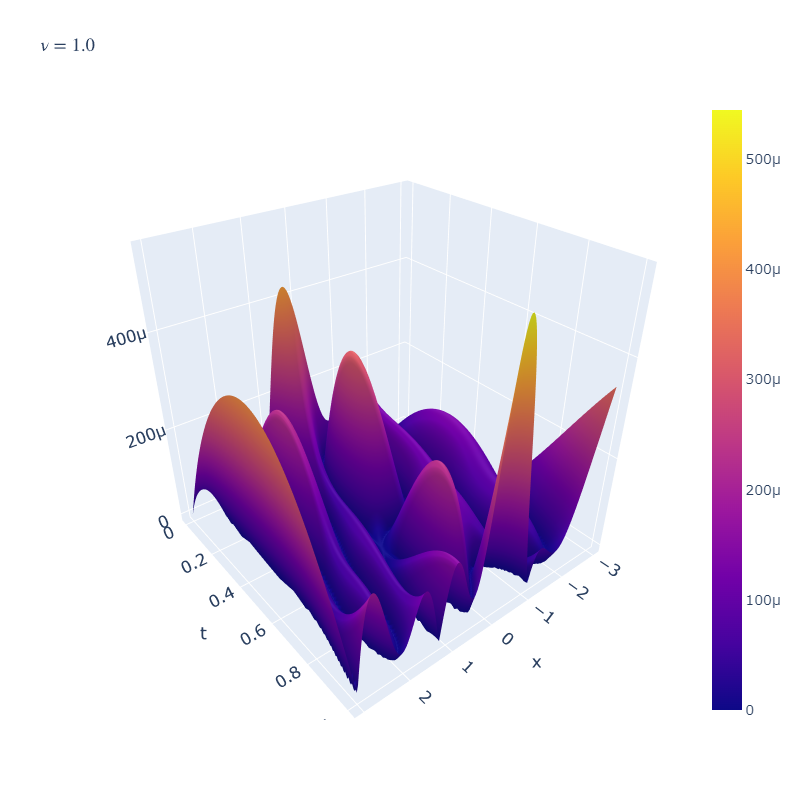}
    \includegraphics[width = 0.32 \linewidth]{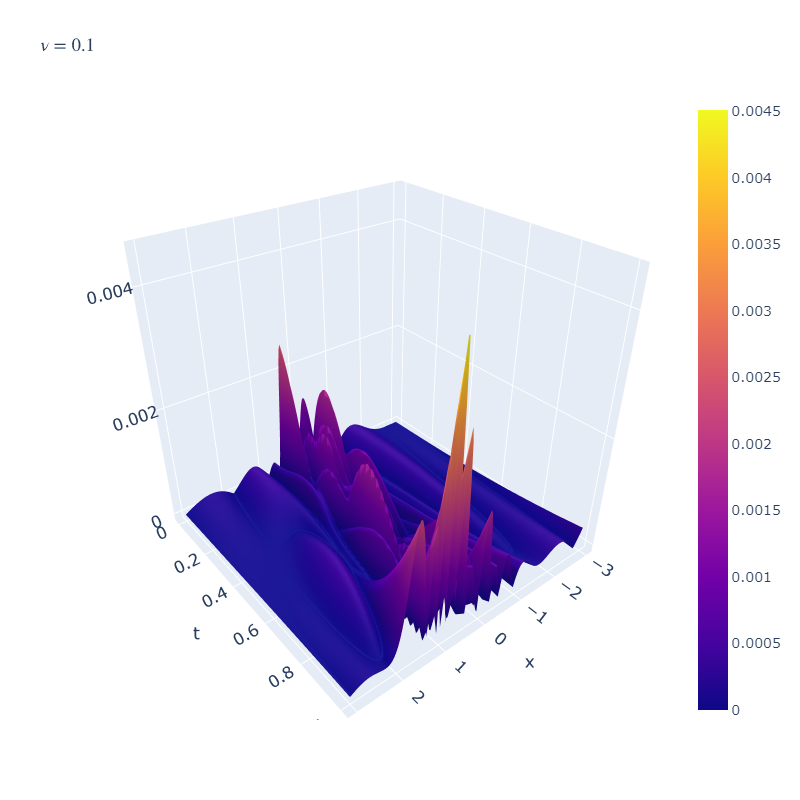}
    \includegraphics[width = 0.32 \linewidth]{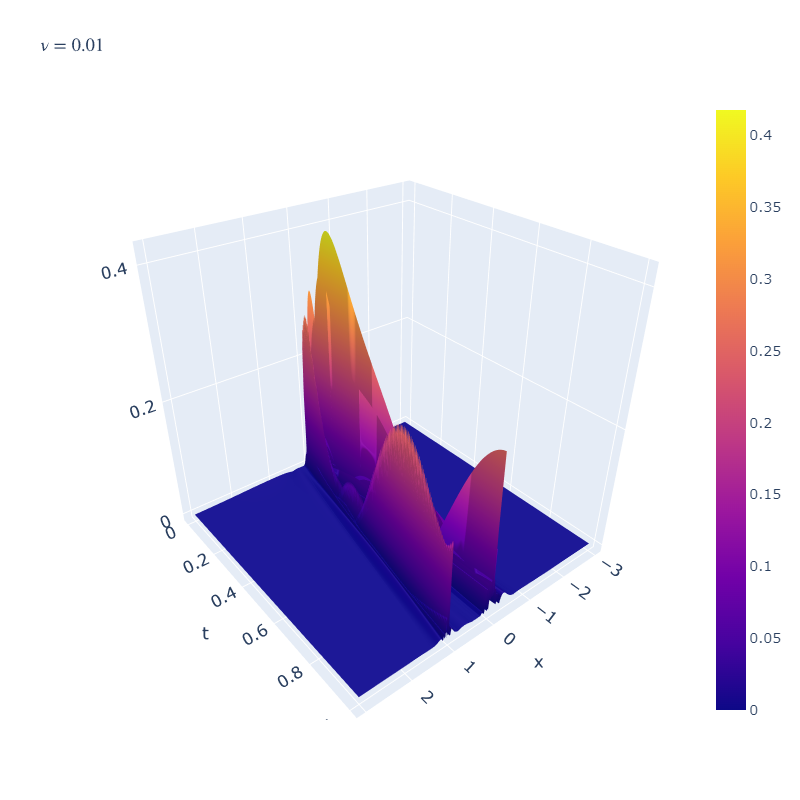}
    \caption{Absolute value of residual of Burgers' equation for \textbf{(left)} $\nu = 1$, \textbf{(center)} $\nu = 0.1$ and \textbf{(right)} $\nu = 0.01$ after training with L-BFGS for 15,000 steps.}
    \label{fig:burgers_res}
\end{figure}

\begin{figure}[h]
    \centering
    \includegraphics[width = 0.49 \linewidth]{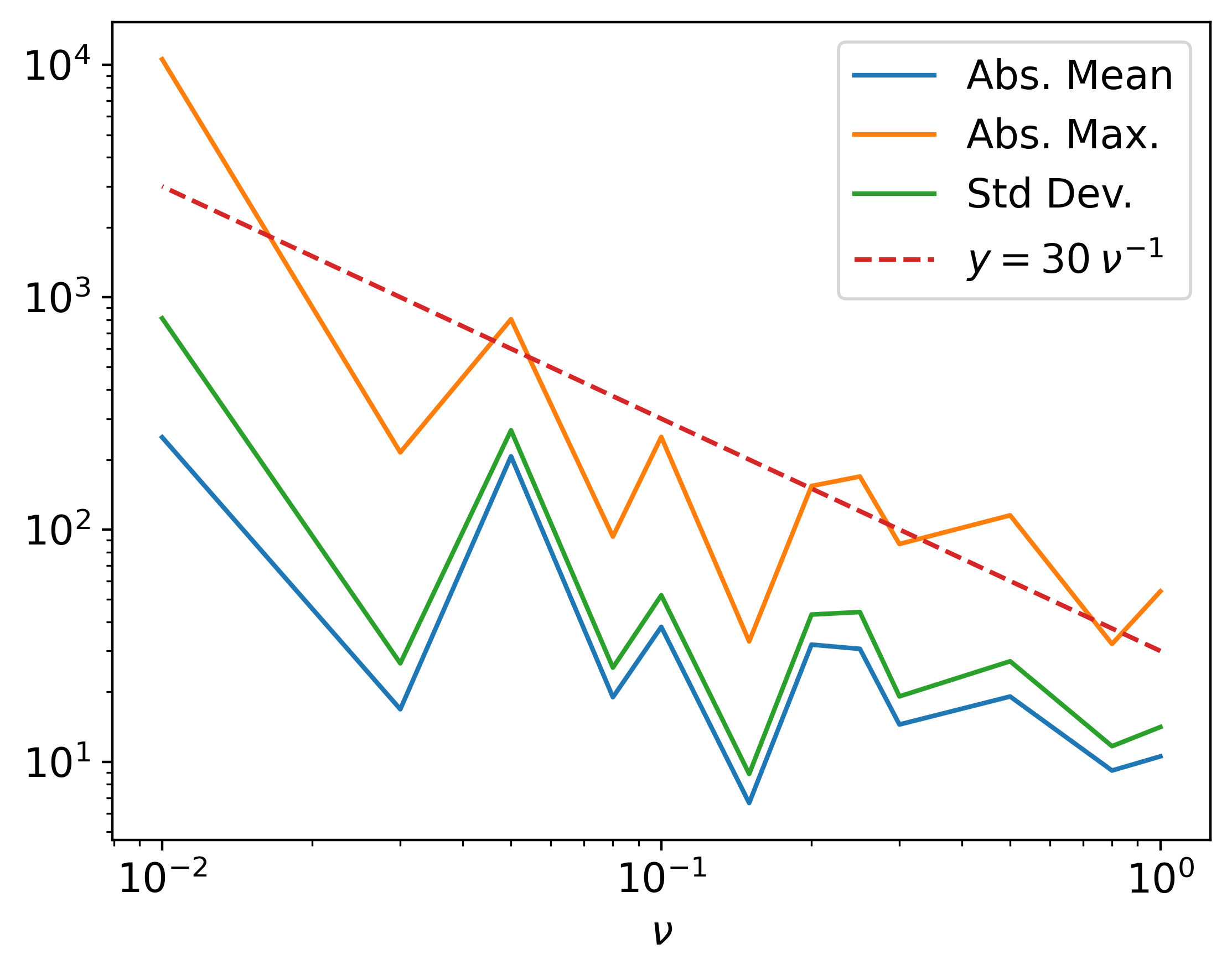}
    \caption{Statistics of the updates proposed to the weights in the last layer of the DNN after the first iteration of GNE. Note that the statistics scales approximately as $\nu^{-1}$.}
    \label{fig:burgers_dw}
\end{figure}

Fig. \ref{fig:burgers_res} shows the residual of the Burgers equation for DNNs trained with L-BFGS for different values of $\nu$. For large values of $\nu$ the residuals are small and the values are evenly distributed throughout the domain. As $\nu$ gets smaller the residuals start to assume large values around the lines $x=0$ and $x=ct$. Here $x=ct$ is the location of the discontinuity when $\nu \to 0$ and the large value of residual at $x=0$ line is due to the presence of $\tanh \left(\frac{c}{2 \nu} x\right)$ term in the TFC constrained expression Eq. \ref{eq:tfc_bruger_disc}. This is a fundamental limitation of TFC when imposing discontinuous boundary conditions or boundary conditions with sharp gradients. In this case, if we impose the boundary condition as a loss function (Eq. \ref{eq:bvp_loss}), we can get rid of the large residual values along the $x=0$ line, but we will still be left with large residual value along $x=ct$. During GNE, these larger residuals localized to a small area of the domain will result in a Jacobian matrix mostly populated by small values with sparsely occuring large values. This in turn will result in large updates (Eq. \ref{eq:gauss_newton}) for the weights. This can be seen in Fig. \ref{fig:burgers_dw}. All the statistics of the proposed updates scales approximately as $\nu^{-1}$. For very small values of $\nu$ ($\nu<0.01$), these large updates result in the solution becoming worse after GNE. For small enough values of $\nu$ all gradient based methods will fail due to exploding gradients, but GNE fails before methods like Adam and L-BFGS since it is more sensitive to large gradients. 

The method described in \cite{liu2022discontinuity} was found to be helpful for extremely small values of $\nu$, even up to $\nu=0$. Nevertheless, the results were not always consistent, occasionally producing solutions with the discontinuity on a line other than $x=ct$.

\FloatBarrier
\subsubsection{2+1 D Heat Equation}
\FloatBarrier
\label{sec:heat_pde}
In this subsection we will solve the 2D heat equation
\begin{align}
\frac{\partial^2}{\partial x^2} u(x, y, t) + \frac{\partial^2}{\partial y^2} u(x, y, t)= \kappa  \frac{\partial}{\partial t} u(x, y, t)
\end{align}
subject to the following Dirichlet boundary conditions:
\begin{align}
u(0, y, t)&=0 \\
u(L, y, t)&=0 \\
u(x, 0, t)&=0 \\
u(x, H, t)&=0 \\
u(x, y, 0)&=\sin \left(\frac{\pi x}{L}\right) \sin \left(\frac{\pi y}{H}\right)
\end{align}
The analytic solution is given by:
\begin{equation}
u(x, y, t)=\sin \left(\frac{\pi x}{L}\right) \sin \left(\frac{\pi y}{H}\right) e^{-\left(\frac{\pi^2}{L^2}+\frac{\pi^2}{H^2}\right) t}
\end{equation}

The equation is solved in the domain $x, y, t \in[0, L] \times[0, H] \times[0,1]$. In this study we use the values $L=2, \ H=1, \ \kappa = 1$. The ELM network had a single hidden layer with 400 neurons. Further increase in the number of neurons did not improve the numerical solution. The deep network had 3 hidden layers with (32, 32, 400) neurons. At each L-BFGS step the network was trained on 2000 random uniformly sampled points from the domain and GNE in ELM and after L-BFGS was done with 2000 random uniformly sampled points from the domain.

\begin{figure}[h]
    \centering
    \includegraphics[width = 0.49 \linewidth]{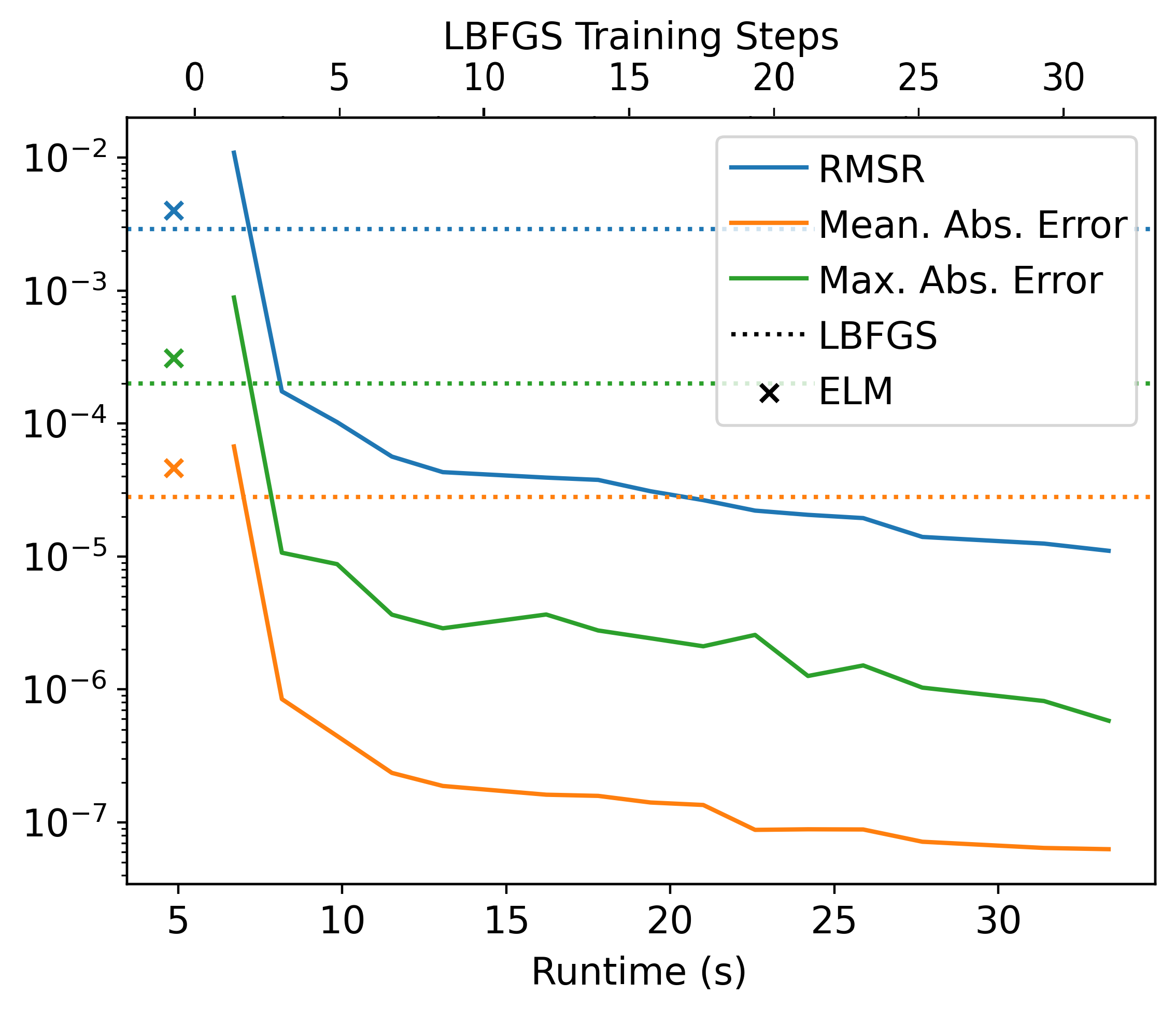}
    \caption{Statistics of the solution to heat equation computed with different methods and using Reduced TFC to impose constraints. Solid lines denote L-BFGS + GNE, dotted lines denote L-BFGS and x marks ELM. The time taken by L-BFGS is not shown in the figure since it takes significantly longer than other methods. See Table. \ref{tab:heat_eq} for exact values.}
    \label{fig:heat_eq}
\end{figure}

Fig. \ref{fig:heat_eq} shows that the solution error from L-BFGS is slightly better than ELM but L-BFGS takes significantly longer to compute. L-BFGS + GNE on the other hand takes slightly longer than ELM to compute the solution but achieves lower error by more than 2 orders of magnitude. From Table \ref{tab:heat_eq} it can be seen that using Reduced TFC we are able to compute solutions around 4 times faster with also a marginal improvement in error. This improvement is due to fact that with Reduced TFC we require only a single evaluation of the neural network compared to multiple evaluations in the case of TFC.

\begin{table}[h]
\centering
\begin{tabular}{|l|l|l|l|l|} 
\hline
 & RMSR & Mean Abs. Err. & Max. Abs. Err.  & Time \\
 \hline
\multicolumn{5}{|c|}{Reduced TFC} \\
\hline
ELM & $4.0 \times 10^{-3}$ & $4.6 \times 10^{-5}$ & $3.1 \times 10^{-4}$ & 4.9 s\\
L-BFGS & $2.9 \times 10^{-3}$ & $2.8 \times 10^{-5}$ & $2.0 \times 10^{-4}$ & 264 s\\
L-BFGS + GNE & $1.1 \times 10^{-5}$ & $6.2 \times 10^{-8}$ & $5.9 \times 10^{-7}$ & 32 s\\
\hline
\multicolumn{5}{|c|}{TFC} \\
\hline
ELM & $4.4 \times 10^{-3}$ & $6.0 \times 10^{-5}$ & $3.6 \times 10^{-4}$ & 19.5 s\\
L-BFGS & $4.2 \times 10^{-3}$ & $3.8 \times 10^{-5}$ & $2.5 \times 10^{-4}$ & 955 s\\
L-BFGS + GNE & $3.2 \times 10^{-5}$ & $1.3 \times 10^{-7}$ & $1.6 \times 10^{-6}$ & 136 s\\
\hline
\end{tabular}
\caption{Statistics for different training algorithms. The shallow ELM had a single hidden layer with 400 neurons. The deep network had 3 hidden layers with (32, 32, 400) neurons. The number of L-BFGS training steps used were 400 for pure L-BFGS and 30 for L-BFGS + GNE.}
\label{tab:heat_eq}
\end{table}

\FloatBarrier
\subsubsection{3+1 D Non-linear PDE}
\label{sec:3p1_nl_pde}
In this subsection we will showcase an example where Reduced TFC provides significant advantage over TFC. We will solve the following 3+1 dimensional non-linear PDE.
\begin{equation}
\begin{aligned}
\partial_xu(x, y, z, t)\, \partial_y u(x, y, z, t)\, \partial_z u(x, y, z, t)+&\partial_t^2u(x, y, z, t)\\
=&\left((t-1) t x(z-1)+x^2 \cos \left(x^2 y\right)+\frac{3}{2} x \sqrt{y} z\right) \\
&\left((t-1) t y(z-1)+2 x y \cos \left(x^2 y\right)+y^{3 / 2} z\right) \\
&\left(2 \pi t^2 \cos (2 \pi z)+(t-1) t x y+x y^{3 / 2}\right)\\
&+2 x y(z-1)+2 \sin (2 \pi z)
\end{aligned}
\end{equation}
subject to the following Dirichlet boundary conditions: 
\begin{align}
&u(0, y, z, t)=t^2 \sin (2 \pi z) \\
&u(x, 0, z, t)=t^2 \sin (2 \pi z) \\
&u(x, y, 1, t)=\sin \left(x^2 y\right)+x y^{3 / 2} \\
&u(x, y, z, 0)=\sin \left(x^2 y\right)+x y^{3 / 2} z \\
&u(x, y, z, 1)=\sin \left(x^2 y\right)+x y^{3 / 2} z+\sin (2 \pi z).
\end{align}
The exact analytic solution for this boundary value problem is given by:
\begin{equation}
    u(x, y, z, t)=t^2 \sin (2 \pi z)+\sin \left(x^2 y\right)+x y^{3 / 2} z+x y t(z-1)(t-1).
\end{equation}
The PDE was solved in the domain $(x,y,z,t) \in [0,1] \times [0,1] \times [0,1] \times [0,1]$. The ELM used in this case had 1 single hidden layer with 400 neurons. The deep network had 3 hidden layers with (32, 32, 400) neurons. At each L-BFGS step the network was trained on 2000 random uniformly sampled points from the domain and GNE in ELM and after L-BFGS was done with 2000 random uniformly sampled points from the domain.

Table \ref{tab:3p1_pde} shows that L-BFGS + GNE provides only a marginal improvement in accuracy over ELM. The important thing to note here is that, using Reduced TFC we are able to achieve more than 20 times speedup in the case of ELM and L-BFGS + GNE, and more than 40 times speedup for L-BFGS.
\begin{table}[H]
\centering
\begin{tabular}{|l|l|l|l|l|} 
\hline
 & RMSR & Mean Abs. Err. & Max. Abs. Err.  & Time \\
 \hline
\multicolumn{5}{|c|}{Reduced TFC} \\
\hline
ELM & $2.8 \times 10^{-9}$ & $5.9 \times 10^{-11}$ & $5.4 \times 10^{-9}$ & 7s\\
L-BFGS & $1.5 \times 10^{-4}$ & $3.9 \times 10^{-6}$ & $6.2 \times 10^{-5}$ & 47s\\
L-BFGS + GNE & $6.6 \times 10^{-10}$ & $1.8 \times 10^{-11}$ & $2.0 \times 10^{-9}$ & 10s \\
\hline
\multicolumn{5}{|c|}{TFC} \\
\hline
ELM & $3.3 \times 10^{-6}$ & $1.0 \times 10^{-7}$ & $1.2 \times 10^{-5}$ & 150s\\
L-BFGS & $3.4 \times 10^{-4}$ & $7.3 \times 10^{-6}$ & $2.0 \times 10^{-4}$ & 1899 s\\
L-BFGS + GNE & $9.0 \times 10^{-7}$ & $2.2 \times 10^{-8}$ & $1.9 \times 10^{-6}$ & 245 s\\
\hline
\end{tabular}
\caption{Statistics for different training algorithms. The ELM had 1 single hidden layer with 400 neurons. The deep network had 3 hidden layers with (32,32,400) neurons. The number of training steps used were 800 for pure L-BFGS and 10 for L-BFGS + GNE.}
\label{tab:3p1_pde}
\end{table}

\subsection{Coupled PDE}
\label{sec:pinn_cpde}
\subsubsection{Kovasznay Flow Solution}
\label{sec:kovas}
\FloatBarrier
The 2D incompressible stationary Navier-Stokes equation given by
\begin{align}
\label{eq:kov_1}
u \frac{\partial u}{\partial x}+v \frac{\partial u}{\partial y}&=-\frac{1}{\rho} \frac{\partial p}{\partial x}+\nu\left(\frac{\partial^2 u}{\partial x^2}+\frac{\partial^2 u}{\partial y^2}\right) \\
\label{eq:kov_2}
u \frac{\partial v}{\partial x}+v \frac{\partial v}{\partial y}&=-\frac{1}{\rho} \frac{\partial p}{\partial y}+\nu\left(\frac{\partial^2 v}{\partial x^2}+\frac{\partial^2 v}{\partial y^2}\right)\\
\label{eq:kov_3}
\frac{\partial u}{\partial x}+\frac{\partial v}{\partial y}&=0, 
\end{align}
where $u$ and $v$ are $x$ and $y$ components of the velocity respectively, $p$ the pressure, $\nu$ the viscosity and $\rho$ the density. The Kovasznay flow \cite{kovasznay1948laminar} is an exact solution to the above coupled system of PDEs given as:
\begin{align}
&u(x,y)=1-e^{\lambda x} \cos (2 \pi y) \\
&v(x,y)=\frac{\lambda}{2 \pi} e^{\lambda x} \sin (2 \pi y)\\
&p(x,y)=p_0-\frac{1}{2} e^{2 \lambda x}, \ \text{where $p_0$ is an arbitrary constant}\\
&\lambda=\frac{1}{2 \nu}-\sqrt{\frac{1}{4 \nu^2}+4 \pi^2}.\\
\end{align}

\begin{figure}[h]
    \centering
    \includegraphics[width = 0.32 \linewidth]{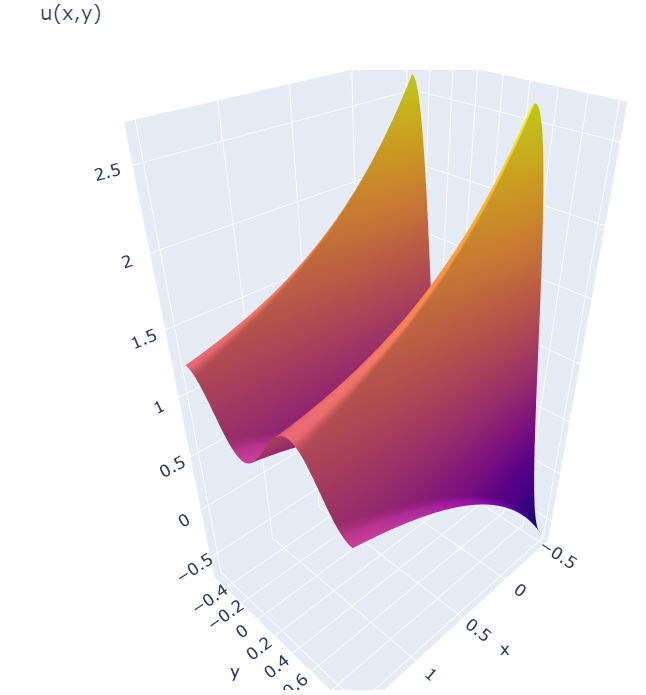}
    \includegraphics[width = 0.32 \linewidth]{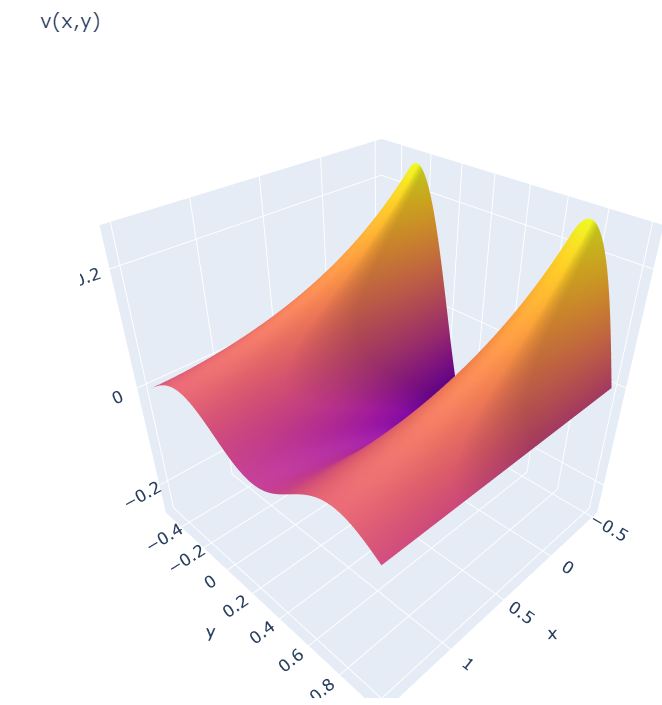}
    \includegraphics[width = 0.32 \linewidth]{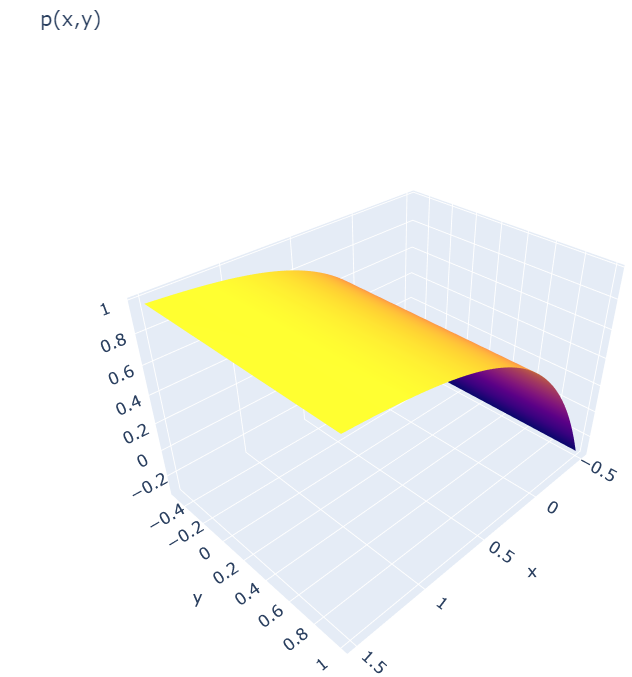}
    \caption{Plot of analytic Kovasznay flow solution with $\nu=0.025$, $\rho=1$ and $p_0 = 1$.
    }
    \label{fig:kavos_exact}
\end{figure}

The Kovasznay flow solution shown in Fig. \ref{fig:kavos_exact} is often used to benchmark traditional \cite{lee2018reconstruction, wijayanta2021numerical} as well as neural network-based \cite{wang2020multi, lou2021physics} numerical solvers. Compared to these methods, in this subsection we show that using L-BFGS + GNE we are able to achieve orders of magnitude more accurate numerical solutions in a fraction of the time. For this test we set the values $\nu=0.025$, $\rho=1$ and $p_0 = 1$. The domain is chosen as $(x,y) \in [0,2]\times[0,2]$ and we apply Dirichlet boundary conditions on $u(x,y)$ and $v(x,y)$ at all 4 boundaries of the domain based on the analytic solution. In principle without applying any boundary condition on $p(x,y)$, we can still solve for $u(x,y)$ and $v(x,y)$. In practice when no boundary condition is applied on $p(x,y)$ we have an extra degree of freedom, related to the pressure $p_0$ which is found to become extremely large ($\sim 10^7$) when solved using neural networks. This in turn leads to a loss in precision as the computations are limited by double precision arithmetic. Therefore we apply Dirichlet boundary condition for the pressure $p(x,y)$ on any one of the boundaries.

The shallow network used for ELM had 400 neurons in the single hidden layer. Increasing the number of neurons further did not decrease the error. The deep network had 3 hidden layers with (32, 32, 400) neurons. At each L-BFGS step the network was trained on 2000 random uniformly sampled points from the domain and GNE in ELM and after L-BFGS was done with 3000 random uniformly sampled points from the domain.

\begin{figure}[h]
    \centering
    \includegraphics[width = 0.32 \linewidth]{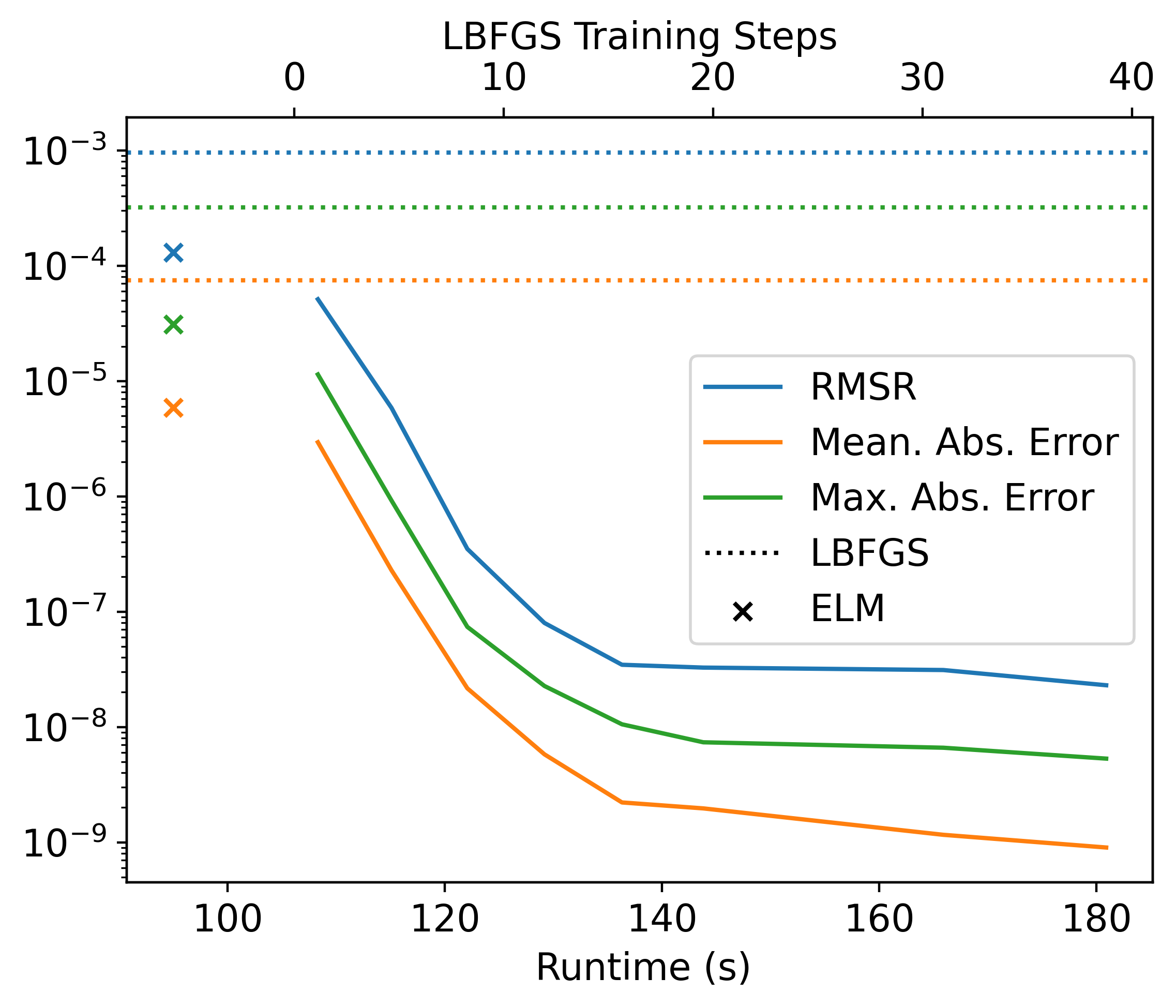}
    \includegraphics[width = 0.32 \linewidth]{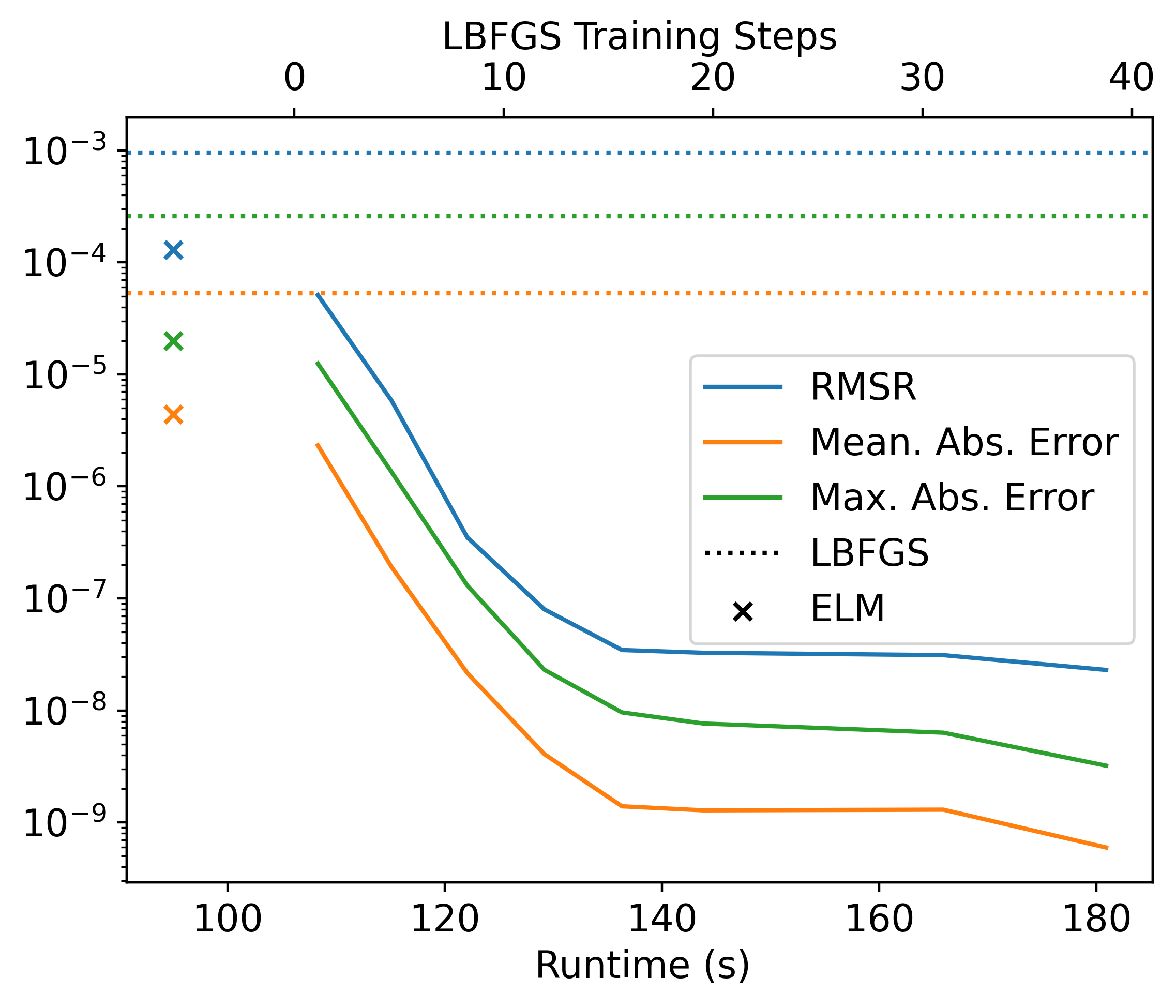}
    \includegraphics[width = 0.32 \linewidth]{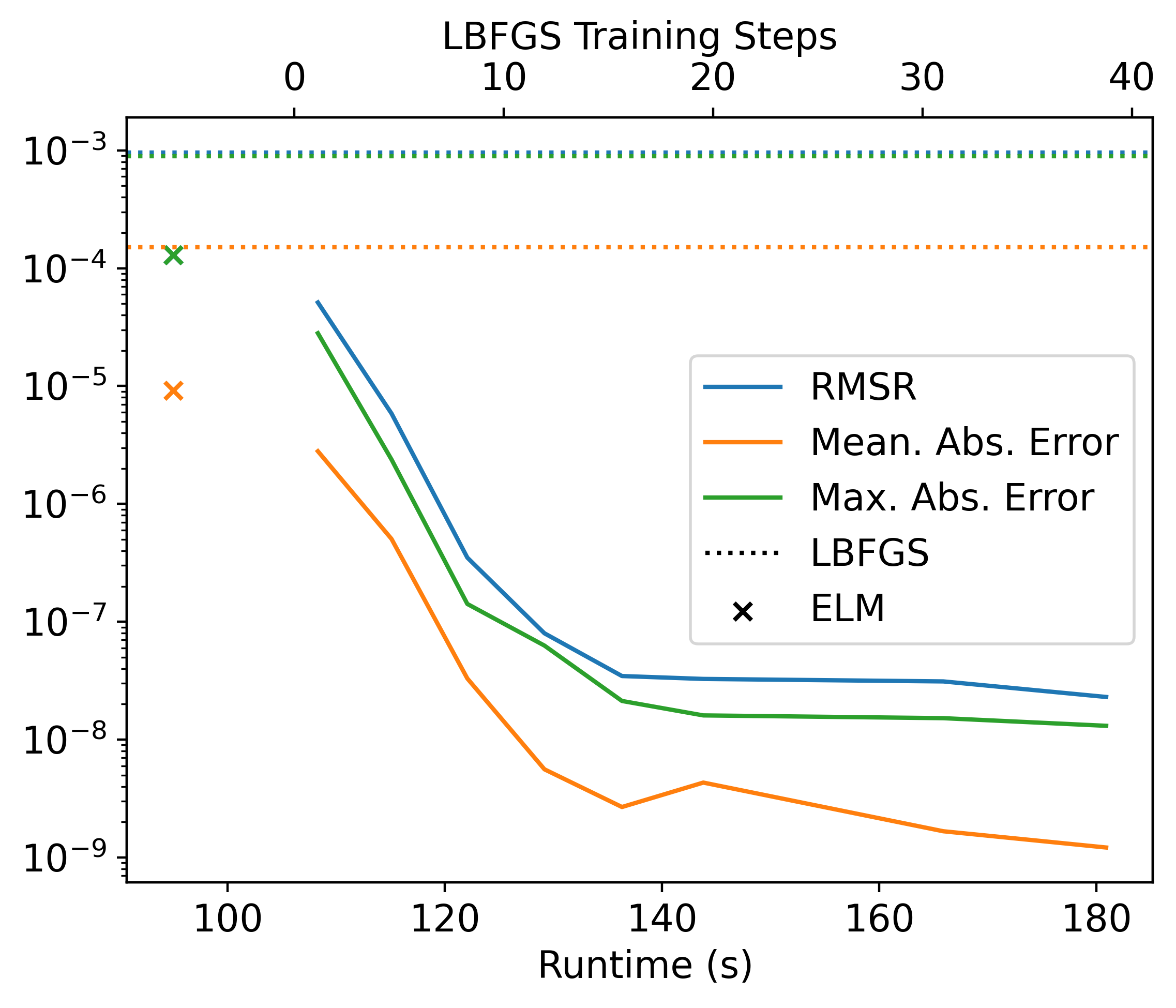}
    \caption{Statistics of the Kovasznay flow solution computed with different methods using Reduced TFC to impose constraints for \textbf{(left)} $u(x,y)$, \textbf{(center)} $v(x,y)$ and \textbf{(right)} $p(x,y)$. Solid lines denote L-BFGS + GNE, dotted lines denote L-BFGS and x marks ELM. The time taken by L-BFGS is not shown in the figure since it takes significantly longer than other methods. See Table. \ref{tab:kovas} for exact values.
    }
    \label{fig:kavos_err}
\end{figure}

As illustrated in Fig. \ref{fig:kavos_err}, even though L-BFGS + GNE takes longer to compute the solution compared to ELM, the LBGS + GNE solution has several orders of magnitude lower error.  L-BFGS, even after running for 600 training steps, still lagged behind ELM in terms of accuracy while taking significantly longer to compute. Table \ref{tab:kovas} shows the advantage of Reduced TFC over TFC. Reduced TFC provided a 50\% improvement in runtime. This pales in comparison to Sec. \ref{sec:heat_pde} and Sec. \ref{sec:3p1_nl_pde} where we achieved a speedup of 4 times and 20 times respectively. This is mostly due to the fact that for this boundary value problem, the computational time is dominated by computing the various derivatives in the PDE using auto-diffferentiation rather than evaluating the neural network multiple times to impose constraints using TFC.

\begin{table}[h]
\centering
\resizebox{\textwidth}{!}{
\begin{tabular}{|l|c|c|c|c|c|c|c|c|c|c|}
\hline
& \multicolumn{3}{|c|}{RMSR} & \multicolumn{3}{|c|}{Mean Abs. Err.} & \multicolumn{3}{|c|}{Max. Abs. Err.} & Time\\
\hline
 & Eq. \ref{eq:kov_1} & Eq. \ref{eq:kov_2} & Eq. \ref{eq:kov_3} & $u(x,y)$ & $v(x,y)$ & $p(x,y)$ & $u(x,y)$ & $v(x,y)$ & $p(x,y)$ & \\
 \hline
\multicolumn{11}{|c|}{Reduced TFC} \\
\hline
%ELM & $1.5 \times 10^{-4}$ & $1.3 \times 10^{-4}$ & $1.1 \times 10^{-4}$ & $5.9 \times 10^{-6}$ & $4.4 \times 10^{-6}$ & $9.2 \times 10^{-6}$ & $3.1 \times 10^{-5}$ & $2.0 \times 10^{-5}$ & $1.3 \times 10^{-4}$ & 16.2 s\\ %llayer_200
ELM & $2.8 \times 10^{-4}$ & $2.6 \times 10^{-4}$ & $2.4 \times 10^{-4}$ & $1.8 \times 10^{-5}$ & $1.3 \times 10^{-5}$ & $1.9 \times 10^{-5}$ & $6.4 \times 10^{-5}$ & $7.3 \times 10^{-5}$ & $1.1 \times 10^{-4}$ & 95 s \\
L-BFGS & $1.4 \times 10^{-3}$ & $7.9 \times 10^{-4}$ & $6.9 \times 10^{-4}$ & $7.5 \times 10^{-5}$ & $5.3 \times 10^{-5}$ & $1.5 \times 10^{-4}$ & $3.2 \times 10^{-4}$ & $2.6 \times 10^{-4}$ & $8.9 \times 10^{-4}$ & 857 s\\
%L-BFGS+GNE & $8.1 \times 10^{-7}$ & $3.2 \times 10^{-7}$ & $4.4 \times 10^{-7}$ & $2.3 \times 10^{-8}$ & $1.9 \times 10^{-8}$ & $3.7 \times 10^{-8}$ & $1.7 \times 10^{-7}$ & $1.1 \times 10^{-7}$ & $3.9 \times 10^{-7}$ & 140 s\\ %llayer_200
L-BFGS + GNE & $2.9 \times 10^{-8}$ & $1.9 \times 10^{-8}$ & $2.1 \times 10^{-8}$ & $9.0 \times 10^{-10}$ & $6.0 \times 10^{-10}$ & $1.2 \times 10^{-9}$ & $5.3 \times 10^{-9}$ & $3.2 \times 10^{-9}$ & $1.1 \times 10^{-8}$ & 182 s \\
\hline
\multicolumn{11}{|c|}{TFC} \\
\hline
ELM & $1.4 \times 10^{-4}$ & $1.3 \times 10^{-4}$ & $1.2 \times 10^{-4}$ & $7.0 \times 10^{-6}$ & $5.1 \times 10^{-6}$ & $1.5 \times 10^{-5}$ & $3.7 \times 10^{-5}$ & $2.8 \times 10^{-5}$ & $1.0 \times 10^{-4}$ & 147 s \\
L-BFGS & $1.3 \times 10^{-3}$ & $9.3 \times 10^{-4}$ & $1.0 \times 10^{-3}$ & $8.8 \times 10^{-5}$ & $8.2 \times 10^{-5}$ & $1.1 \times 10^{-4}$ & $3.1 \times 10^{-4}$ & $4.7 \times 10^{-4}$ & $5.9 \times 10^{-4}$ & 1307 s \\
L-BFGS + GNE & $6.4 \times 10^{-8}$ & $2.5 \times 10^{-8}$ & $3.8 \times 10^{-8}$ & $2.4 \times 10^{-9}$ & $1.4 \times 10^{-9}$ & $5.2 \times 10^{-9}$ & $1.6 \times 10^{-8}$ & $7.2 \times 10^{-9}$ & $4.5 \times 10^{-8}$ & 318 s\\
\hline
\end{tabular}
}
\caption{Statistics of the Kovasznay flow solution computed with different training methods using TFC and Reduced TFC. The number of L-BFGS training steps used were 600 for pure L-BFGS and 40 for L-BFGS + GNE.}
\label{tab:kovas}
\end{table}
Note that the slowest step in the GNE process is the computation of the Jacobian matrix. Since GPUs are equipped with a limited amount of VRAM, this computation is often done in batches. For instance in the case of the neural network used in this subsection, for each Gauss-Newton iteration, Jacobian matrix of 3000 sample outputs w.r.t. 1200 weights was computed in batches of 100 and took around 25s. Once the Jacobian matrix was computed, the LLS took on average 560ms per Gauss-Newton iteration. Since we use random sampling from the domain to train, multiple GPUs can readily be used to parallelize the batch-wise computation of the Jacobian. For multi-dimensional PDEs the synchronization time between all the GPUs and time for computing the LLS will be negligible compared to the time it takes to compute the Jacobian matrix, resulting in an almost linear speedup. Another option discussed in Sec. \ref{sec:training} is to derive an analytic expression for Jacobian.
\FloatBarrier
\subsubsection{Taylor-Green Vortex Solution}
\label{sec:taylor_pde}
\FloatBarrier

The 2D incompressible Navier-Stokes equations are written as
\begin{align}
&\frac{\partial u}{\partial t}+u \frac{\partial u}{\partial x}+v \frac{\partial u}{\partial y}=-\frac{1}{\rho} \frac{\partial p}{\partial x}+\nu\left(\frac{\partial^2 u}{\partial x^2}+\frac{\partial^2 u}{\partial y^2}\right) \\
&\frac{\partial v}{\partial t}+u \frac{\partial v}{\partial x}+v \frac{\partial v}{\partial y}=-\frac{1}{\rho} \frac{\partial p}{\partial y}+\nu\left(\frac{\partial^2 v}{\partial x^2}+\frac{\partial^2 v}{\partial y^2}\right)\\
&\frac{\partial u}{\partial x}+\frac{\partial v}{\partial y}=0 
\end{align}

The Taylor-Green vortex solution \cite{taylor1937mechanism} to the Navier-Stokes equations describes an unsteady flow with decaying vortex and can be written as follows:

\begin{align}
\label{eq:taylor_e1}
&u(x,y,t)=\sin x \cos y e^{-2 \nu t} \\
\label{eq:taylor_e2}
&v(x,y,t)=-\cos x \sin y e^{-2 \nu t} \\
\label{eq:taylor_e3}
&p(x,y,t)=\frac{\rho}{4}(\cos 2 x+\sin 2 y) e^{-4 \nu t}.
\end{align}

In this case we will solve the Navier-Stokes equations by applying periodic boundary conditions and initial conditions which match the analytic solution for $u(x,y,t)$, $v(x,y,t)$ and $p(x,y,t)$ at $t=0$. As we discussed in Sec. \ref{sec:kovas}, the initial condition on $p(x,y,t)$ is not required in principle but necessary in practice.

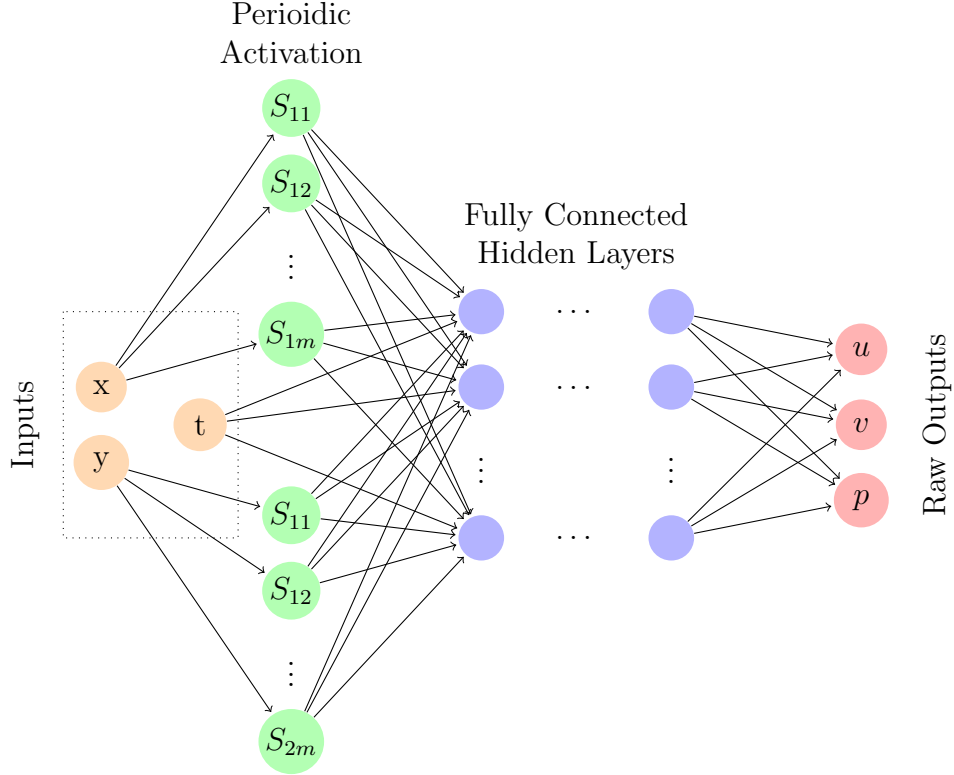
\begin{figure}[h]
\centering
\begin{tikzpicture}

\foreach \i [count=\s] in {x,y}
{
    \node[circle, minimum size = 6mm,fill=orange!30] (Input-\s) at (0,-\s) {\i};
}

\node[circle, minimum size = 7mm,fill=orange!30, yshift=(9-2)*5mm] (Input-3) at (1.3,-5) {t};

\draw[black, dotted] (-0.5,0) rectangle (1.8,-3);
\node [text centered, rotate=90] at (-1,-1.5) {Inputs};

\foreach \s [count=\i] in {$S_{11}$, $S_{12}$, $S_{1n}$, $S_{1m}$}
{
    \ifnum \i=3
    \node[yshift=(9-2)*5mm] at (2.5,-\i+0.2)  {\vdots};
    \else
    \node[circle, minimum size = 7mm,fill=green!30, yshift=(9-2)*5mm, inner sep = 1pt] (Sinx-\i) at (2.5,-\i+0.2) {\s};
    \fi
}

\foreach \s [count=\i] in {$S_{11}$, $S_{12}$, $S^_{2n}$, $S_{2m}$}
{
    \ifnum \i=3
    \node[yshift=(9-2)*5mm]  at (2.5,-\i-5-0.2) {$\vdots$};
    \else
    \node[circle, minimum size = 6mm,fill=green!30, yshift=(9-2)*5mm, inner sep = 1pt] (Siny-\i) at (2.5,-\i-5-0.2) {\s};
    \fi
}

\node [text width=6em, text centered, yshift=(9-2)*5mm] at (2.5,0.2) {Perioidic Activation};

\foreach \j in {1,2,4}
    {
        \draw[->, shorten >=1pt] (Input-1) -- (Sinx-\j);
        \draw[->, shorten >=1pt] (Input-2) -- (Siny-\j);
    }

\foreach \i in {1,2,3,4}
{
    \ifnum \i=3
    \node[yshift=(4-2)*5mm] at (5,-\i)  {\vdots};
    \else
    \node[circle, minimum size = 6mm,fill=blue!30, yshift=(4-2)*5mm] (Hidden-\i) at (5,-\i) {};
    \fi
}

\foreach \i in {1,2,4}
    {
        \draw[->, shorten >=1pt]  (Input-3) -- (Hidden-\i);
        \foreach \j in {1,2,4}
    {
        \draw[->, shorten >=1pt]  (Sinx-\j) -- (Hidden-\i);
        \draw[->, shorten >=1pt]  (Siny-\j) -- (Hidden-\i);
    }
    }
    
\foreach \i in {1,2,3,4}
{
    \ifnum \i=3
    \node[yshift=(4-2)*5mm] at (7.5,-\i)  {\vdots};
    \else
    \node[circle, minimum size = 6mm,fill=blue!30, yshift=(4-2)*5mm] (Hidden2-\i) at (7.5,-\i) {};
    \fi
}
\node[yshift=(4-2)*5mm] at (6.25,-1)  {\dots};
\node[yshift=(4-2)*5mm] at (6.25,-2)  {\dots};
\node[yshift=(4-2)*5mm] at (6.25,-4)  {\dots};

\node [text width=8em, text centered, yshift=(4-2)*5mm] at (6.25,0) {Fully Connected Hidden Layers}; 

\foreach \s [count=\i] in {$u$, $v$, $p$}
{
    \node[circle, minimum size = 6mm,fill=red!30, yshift=(3-2)*5mm] (Output-\i) at (10,-\i) {\s};
}
\node [text centered, rotate=90] at (11,-1.5) {Raw Outputs}; 
\foreach \i in {1,2,3}
    {
        \foreach \j in {1,2,4}
    {
        \draw[->, shorten >=1pt]  (Hidden2-\j) -- (Output-\i);
    }
    }

\end{tikzpicture}
    \caption{Structure of the neural network imposing periodic boundary conditions along the $x$ and $y$ axes. Raw outputs are the neural network outputs before applying TFC constraints.}
    \label{fig:periodic_nn}
\end{figure}

Based on \cite{dong2021method}, the periodic boundary conditions were exactly imposed using an additional layer with a periodic activation function right after the input layer as shown in Fig. \ref{fig:periodic_nn}. The other constraints are imposed using Reduced TFC. The variables $x$ and $y$ of the periodic dimensions are first passed to a set of $m$ periodic activation functions defined as follows:
\begin{align}
    \label{eq:periodic_activation_1}
    S_{1i}(x)=A_{1i} \sin \left(\frac{2 \pi}{L_x} x+\phi_{1i}\right), \quad 1 \leq i \leq m,\\
    \label{eq:periodic_activation_2}
    S_{2i}(y)=A_{2i} \sin \left(\frac{2 \pi}{L_y} y+\phi_{2i}\right), \quad 1 \leq i \leq m.
\end{align}
Here $L_x$ and $L_y$ are the periods of the $x$ and $y$ variables respectively, which have the values $L_x=L_y=2 \pi$ in this example. $A_{1i},\, A_{2i},\, \phi_{1i} \text{ and } \phi_{2i}$ are weights which are initialized randomly with Xavier normal initialization \cite{glorot2010understanding} and remain constant for ELM but are trainable parameters in the case of DNN trained with L-BFGS. $S_{1i}(x)$, $S_{2i}(y)$ and the non-periodic variable $t$ now form the new input variables which are then passed to a FCNN. Note here that Eqs. \ref{eq:periodic_activation_1} and \ref{eq:periodic_activation_2} do not exactly follow the prescription in \cite{dong2021method}. The current form of Eqs. \ref{eq:periodic_activation_1} and \ref{eq:periodic_activation_2} was found to be optimal for ELM. L-BFGS and L-BFGS + GNE was found to be agnostic to the specific form, as long as the periodic activation was present, since they have the additional freedom to train the parameters.

\begin{figure}
    \centering
    \includegraphics[width = 0.5 \linewidth]{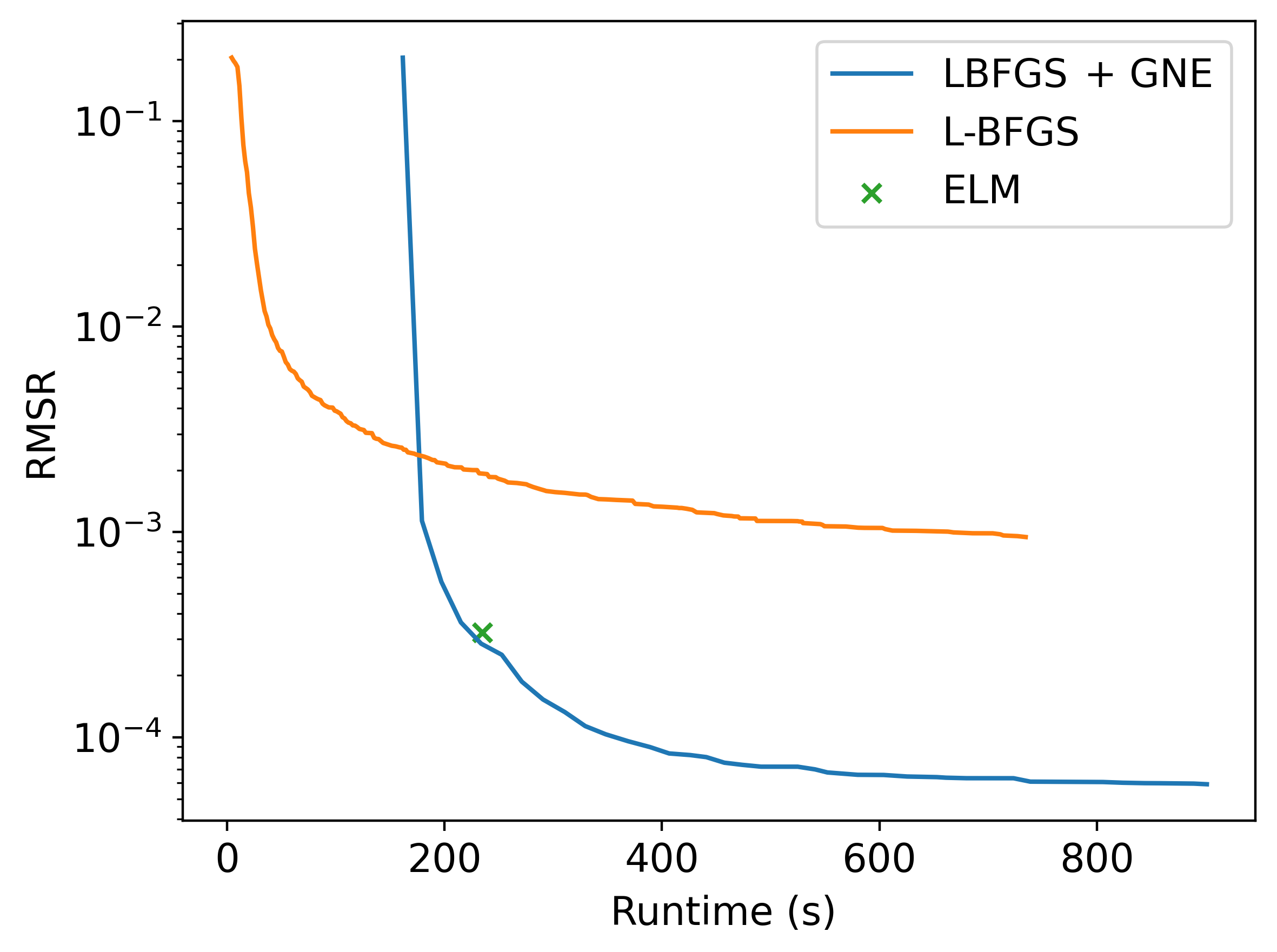}
    \caption{Learning curve of the different training algorithms used for solution of the Taylor-Green vortex problem. For L-BFGS with and without GNE 700 training steps were used.}
    \label{fig:taylor_rmsr}
\end{figure}

The shallow network used for ELM has 15 neurons ($m=15$) each for $x$ and $y$ variables in the periodic activation layer and 400 neurons in the single hidden layer. Increasing the number of neurons in the hidden layer was found to decrease the RMSR but took significantly longer to train due to computational constraints. The deep network had the same configuration for the periodic activation layer and had 3 hidden layers with (32, 32, 400) neurons. At each L-BFGS step the network was trained on 2000 random uniformly sampled points from the domain and GNE in ELM and after L-BFGS was done with 3000 random uniformly sampled points from the domain.

\begin{figure}[h]
    \centering
    \includegraphics[width = 0.49 \linewidth]{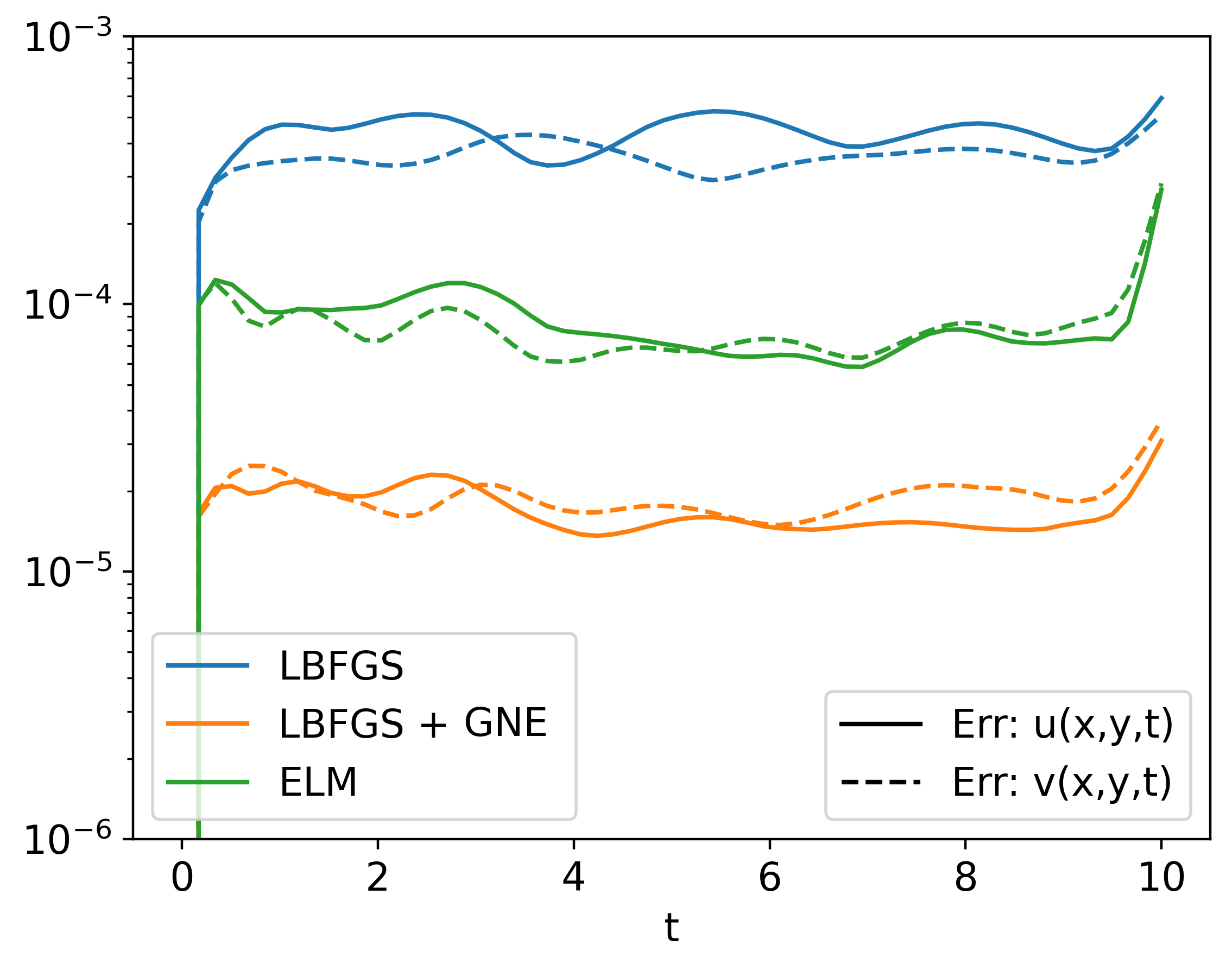}
    \includegraphics[width = 0.49 \linewidth]{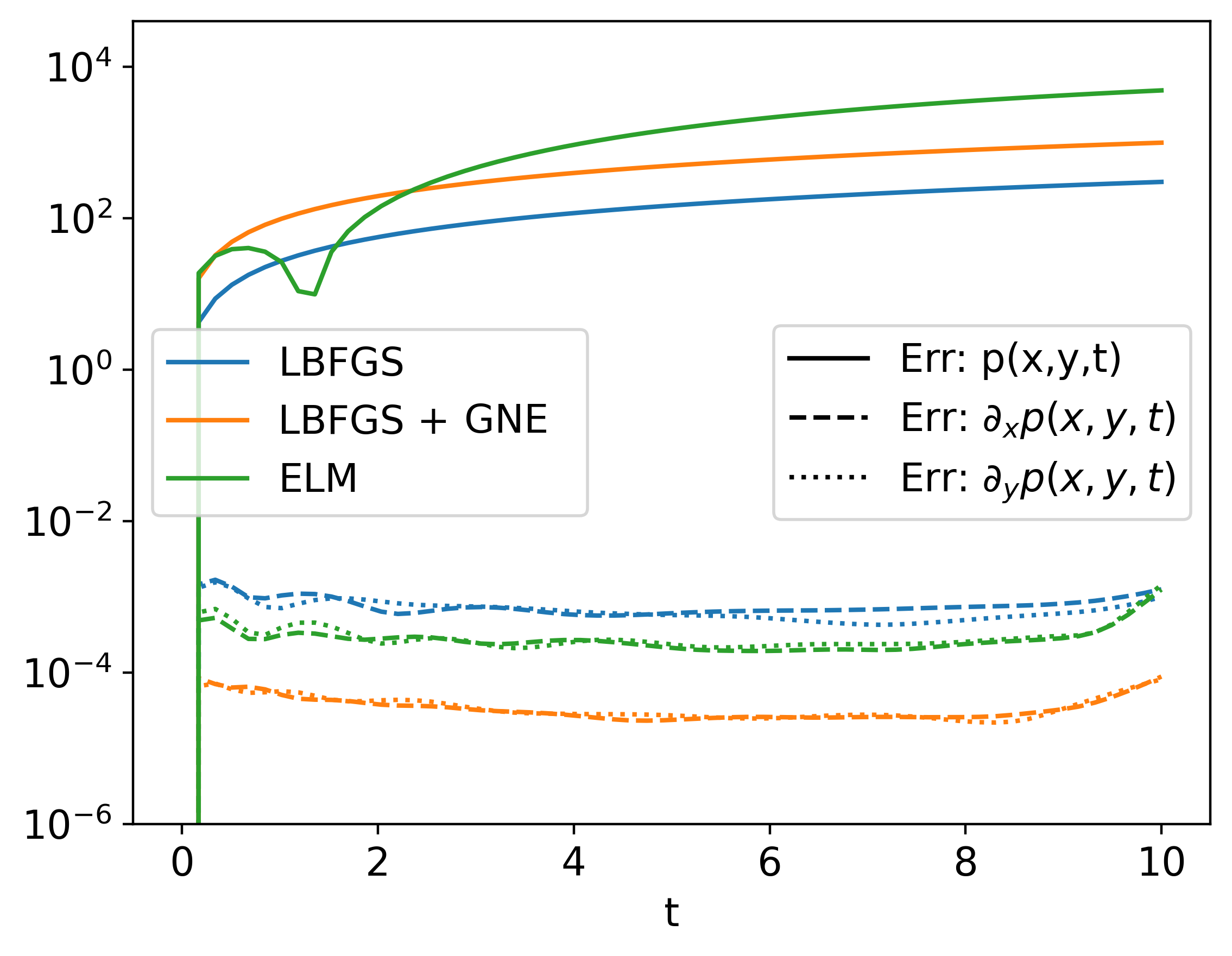}
    \caption{Mean absolute error at different time slices $t$ for the Taylor-Green vortex solution computed using different training methods.}
    \label{fig:taylor_uvp_err}
\end{figure}

Fig. \ref{fig:taylor_rmsr} shows the learning curve for the various training methods used. ELM and L-BFGS + GNE outperform pure L-BFGS. L-BFGS + GNE can acheive the same RMSR value as ELM slighlty faster, and being trained for a longer duration is able to achieve a better RMSR. This corresponds to decreased error in the L-BFGS + GNE solution compared to ELM as shown in Fig. \ref{fig:taylor_uvp_err}. Note that in Fig. \ref{fig:taylor_uvp_err}, the absolute error in $p(x,y,t)$ is extremely large for all the training methods, even though we applied boundary conditions as we did in Sec. \ref{sec:kovas}. This is because $p(x,y,t)$ in this section has an additional gauge freedom where $p(x,y,t) \to p(x,y,t) + g(t)$ will also satisfy the Navier-Stokes equations since $p(x,y,t)$ only appears as a gradient w.r.t. $x$ and $y$ variables in the equation. Imposing any kind of boundary condition at $t=0$ will only constrain the value of $g(t)$ at $t=0$. Even though it is not possible to unambiguously learn the function $p(x,y,t)$, the value of its derivatives can be determined. This is evident in Fig. \ref{fig:taylor_uvp_err} where the errors in $\partial_x p(x,y,t)$ $\partial_y p(x,y,t)$ are small.

\begin{figure}[h]
    \centering
    \includegraphics[width = 0.49 \linewidth]{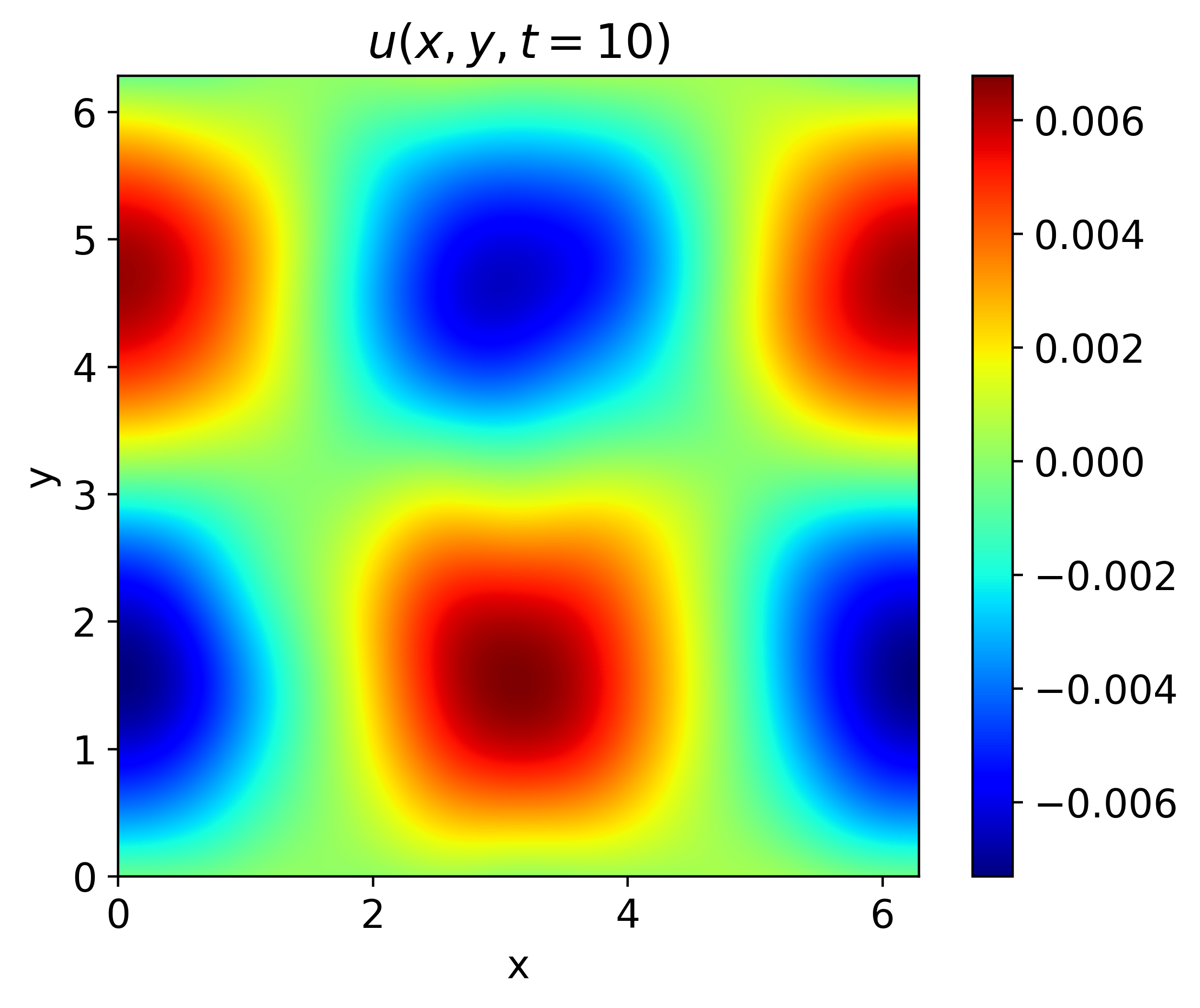}
    \includegraphics[width = 0.49 \linewidth]{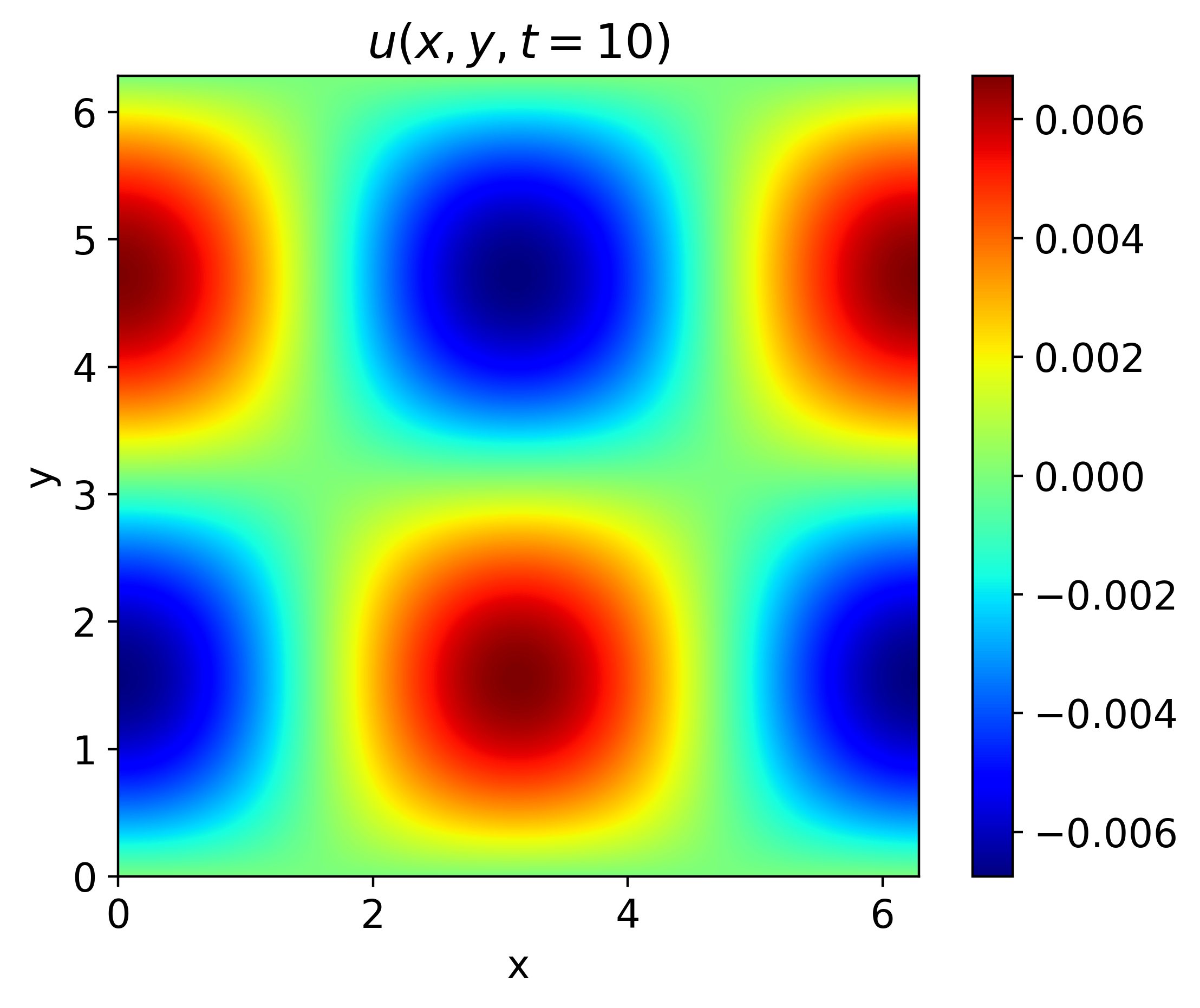}
    \caption{Heat map of numerical solution $u(x,y,t)$ to Taylor-Green vortex problem  at $t=10$ found using \textbf{(left)} ELM and \textbf{(right)} L-BFGS + GNE.}
    \label{fig:taylor_u}
\end{figure}

\begin{figure}[h]
    \centering
    \includegraphics[width = 0.49 \linewidth]{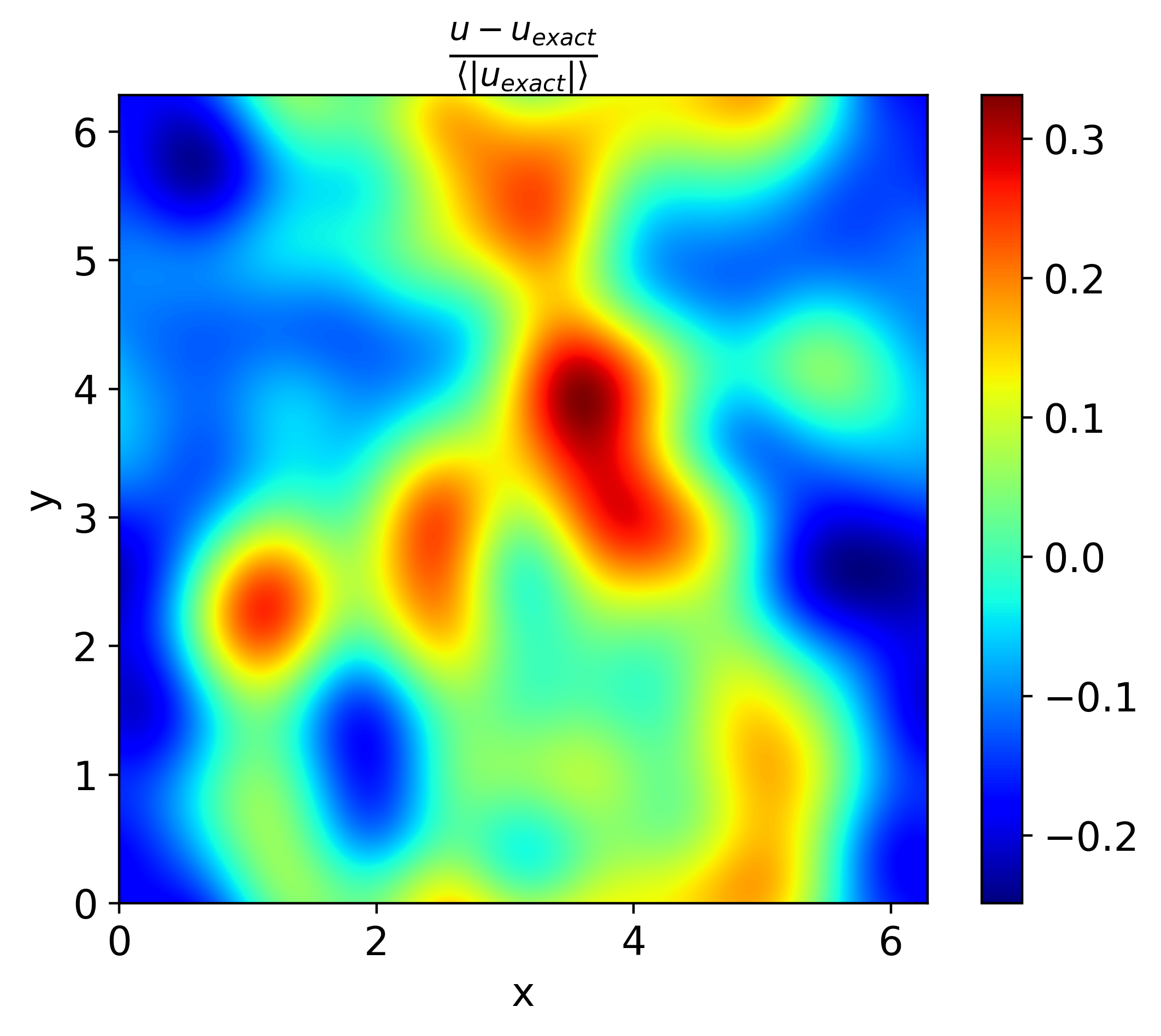}
    \includegraphics[width = 0.49 \linewidth]{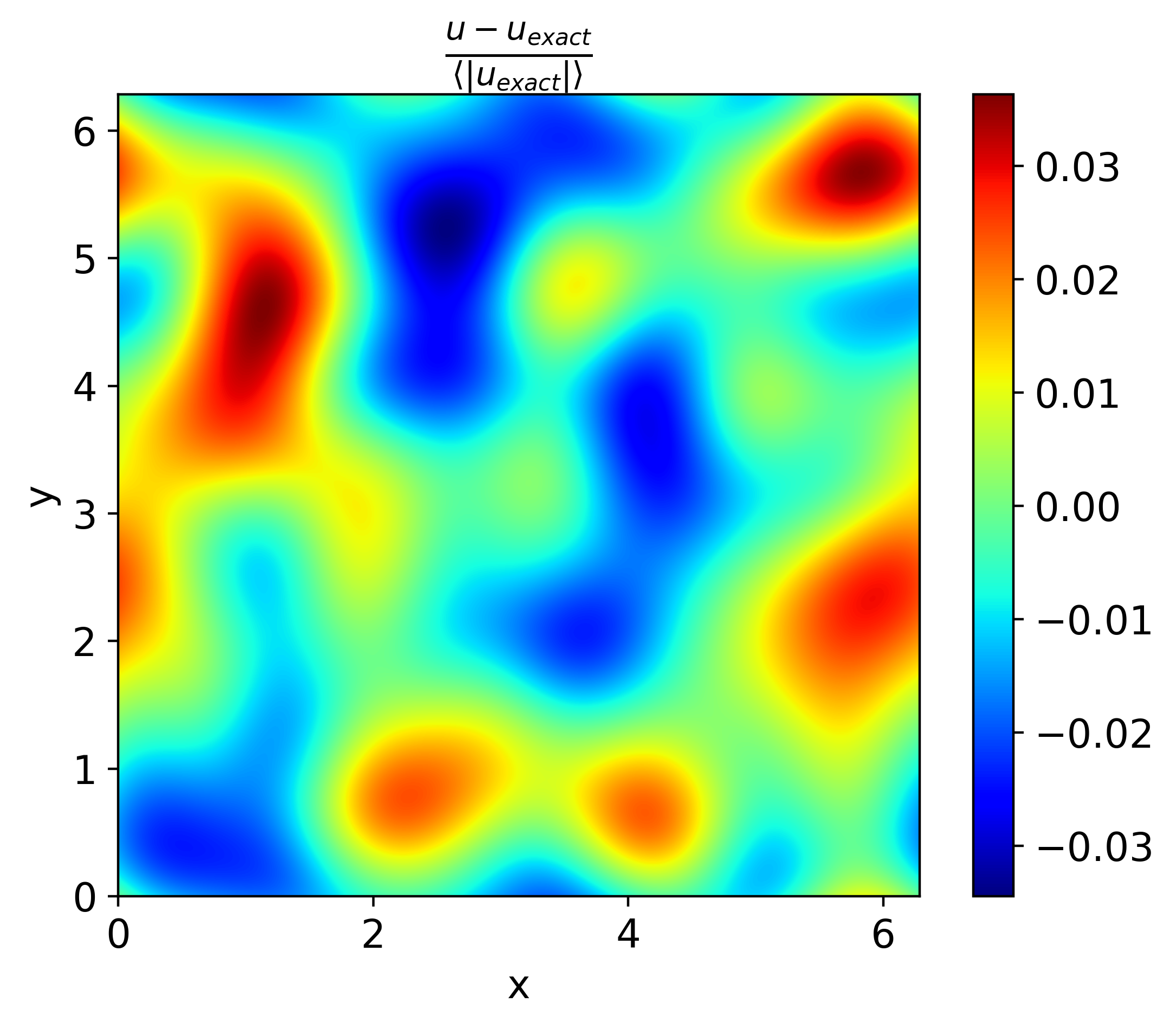}
    \caption{Heat map of relative error of numerical solution $u(x,y,t)$ to Taylor-Green vortex problem  at $t=10$ found using \textbf{(left)} ELM and \textbf{(right)} L-BFGS + GNE.}
    \label{fig:taylor_u_rel_err}
\end{figure}

An important thing to consider here is that the error in  $u$, $v$ and $\nabla p$ are almost a constant with a slight increase towards the end of the $t$ domain. However, the exact solutions given by Eqs. \ref{eq:taylor_e1}, \ref{eq:taylor_e2} and \ref{eq:taylor_e3} are exponentially decaying in time. This means that the relative error in the solution will be exponentially increasing. This increase in relative error translates to visible artifacts in the ELM solution at $t=10$ as shown in Fig. \ref{fig:taylor_u}. L-BFGS + GNE on the other hand yields a solution which is visibly indistinguishable from the true solution. This is even more evident in Fig. \ref{fig:taylor_u_rel_err} which shows that the maximum relative error in the ELM solution is around 30\% whereas for L-BFGS + GNE the maximum relative error is around 3\%.

In order to better learn long-time solutions to time-dependent differential equations, a method, named block time marching (BTM), was introduced in \cite{dong2021local}. With BTM, the time domain is decomposed into sub-domains represented by independent neural networks. Starting from the sub-domain containing the initial condition, the neural networks are trained one at a time and successively, with value from the previous neural network being used as the initial condition for the current network. For this example we decompose the domain into two sub-domains $[0,5.25]$ and $[5,10.25]$. The 5\% extension of the domain was to take into account the slight increase in error towards the end of the domain boundary. The first neural network is trained with the initial condition at $t=0$ and the second network is trained with initial condition at $t=5$, the values of which are evaluated from the previously trained neural network. 

\begin{figure}[h]
    \centering
    \includegraphics[width = 0.49 \linewidth]{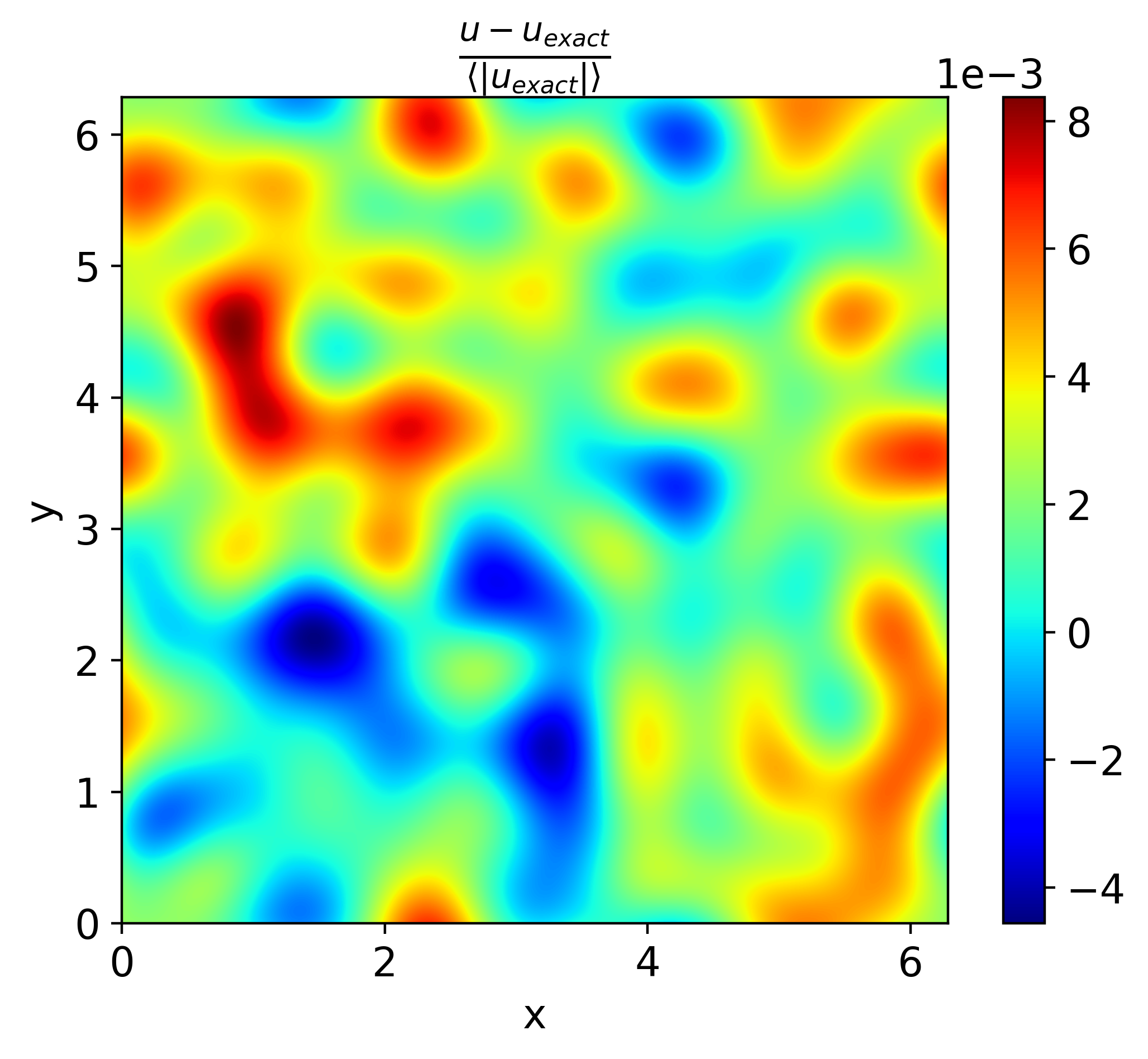}
    \includegraphics[width = 0.49 \linewidth]{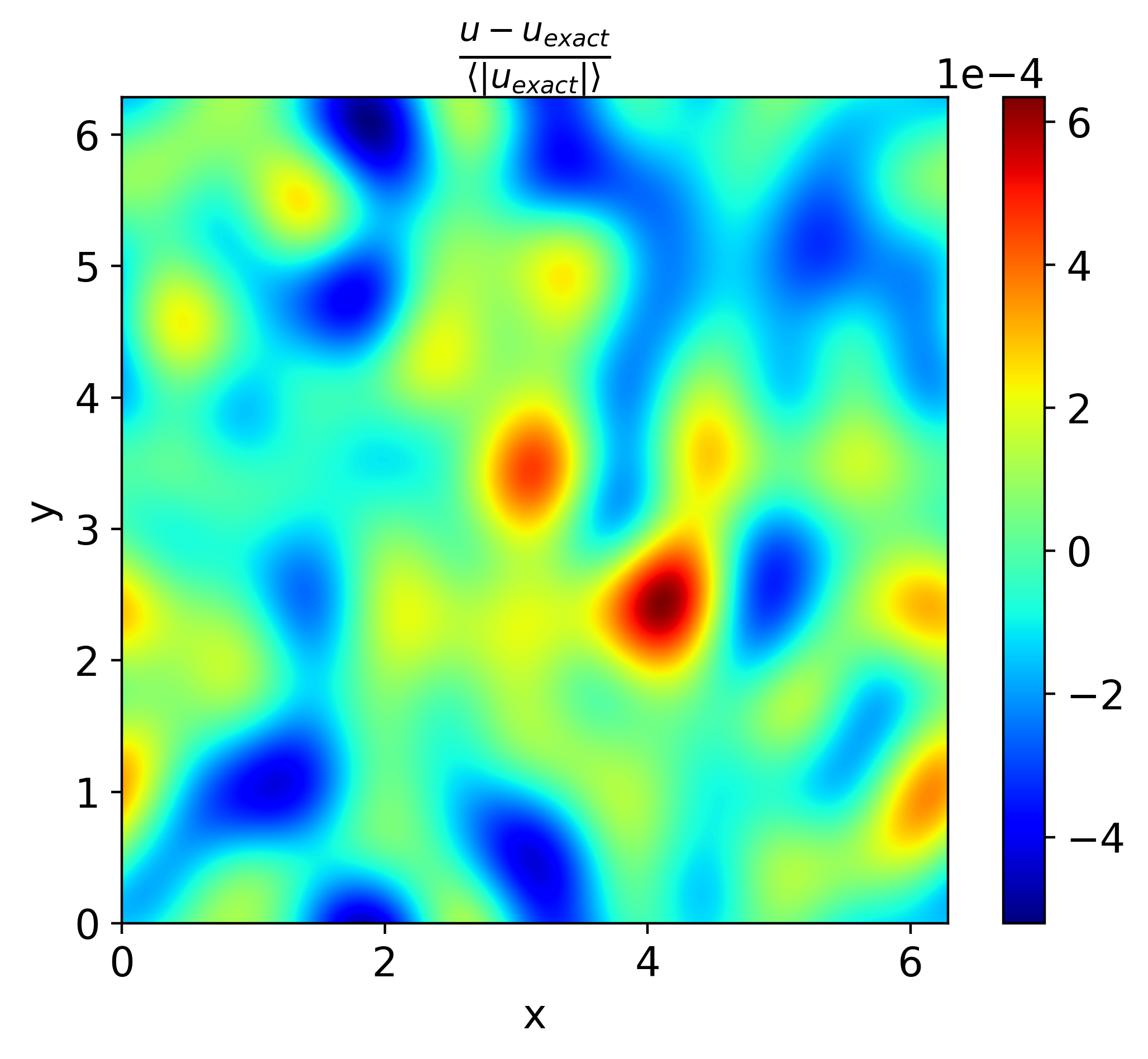}
    \caption{Heat map of relative error of numerical solution $u(x,y,t)$ to Taylor-Green vortex problem at $t=10$ found using a combination of BTM with\textbf{(left)} ELM and \textbf{(right)} L-BFGS + GNE.}
    \label{fig:taylor_u_rel_err_2}
\end{figure}

Fig. \ref{fig:taylor_u_rel_err_2} shows the relative error in the solution found using BTM. The accuracy of both ELM and L-BFGS + GNE improved with BTM, resulting in a 0.8\% maximum relative error in the case of ELM and 0.06\% in the case of L-BFGS + GNE at $t=10$. Note that it is in principle possible to further improve the solution accuracy by decomposing the domain into smaller sub-domains and applying BTM. BTM is but an example of a general class of domain decomposition PINN methods \cite{dong2021local, jagtap2021extended} which may be necessary to find accurate solutions on large domains due to computational hardware limitations. This example serves as an illustration of the applicability of extremization applied on top of domain decomposition to improve solution accuracy. Further in-depth study is required into this topic and is not undertaken at this point.

\FloatBarrier
\subsubsection{Pure Advection in Compressible Flow}
\FloatBarrier
\label{sec:pure_advection}
The compressible 1-dimensional Euler equations are given as
\begin{align}
&\frac{\partial \rho}{\partial t} + \frac{\partial}{\partial x} \left(\rho u\right) = 0\\
&\frac{\partial}{\partial t}  \left(\rho u\right) + \frac{\partial}{\partial x}\left(\rho u^2 + p\right) = 0\\
&\frac{\partial E}{\partial t}+ \frac{\partial}{\partial x} \left(\left(E+p\right) u\right) =0,
\end{align}
where the total (kinetic + internal) energy $E$ is defined as:
\begin{equation}
    E = \frac{1}{2}\rho u^2 + \frac{p}{\gamma-1}.
\end{equation}
We also assume $\gamma = 1.4$. For this system of PDEs, if we start with a constant pressure $p(x,0) = p_0$ and constant velocity $u(x,0) = u_0$, the result is pure advection in density, where we have $\rho(x,t) = \rho(x-u_0 t, 0)$. The pressure and velocity remain constant in time. In this example we start with the ICs 
\begin{align}
    p(x,0) &= 1,\\
    u(x,0) &= 1,\\
    \label{eq:rho_ini}
    \rho(x,0) &= \exp \left(- \frac{\left(x-\mu\right)^2}{2 \sigma^2}\right), \quad\mu=0.5,\, \sigma=0.1.
\end{align}

We solve this system of PDEs on the domain $(x,t) \in [0,1]\times[0,1]$. We use the BC that the density vanishes at infinity, $\lim_{x \to \pm \infty} \rho(x,t) = 0$. The ICs on $p$ and $u$ are imposed using Reduced TFC. In order to impose the BCs on $\rho$ we modify the Reduced TFC constrained expression as follows:
\begin{align}
    f_p^c(x,t) &= \mathcal{N}_p(x,t)\, t + 1\\
    f_u^c(x,t) &= \mathcal{N}_u(x,t)\, t + 1\\
    \label{eq:bc_at_infinity}
    f_{\rho}^c(x,t) &= \exp \left(\mathcal{N}_{\rho}(x,t)\, t -\frac{\left(x-\mu\right)^2}{2 \sigma^2}\right).
\end{align}

Here $\mathcal{N}_p,~\mathcal{N}_u,~\mathcal{N}_{\rho}$ are the neural network outputs before applying constraints. The proof that Eq. \ref{eq:bc_at_infinity} satisfies the BC at infinity is given in \ref{app:bc_at_infinity}.

\begin{figure}[h]
    \centering
    \includegraphics[width = 0.5 \linewidth]{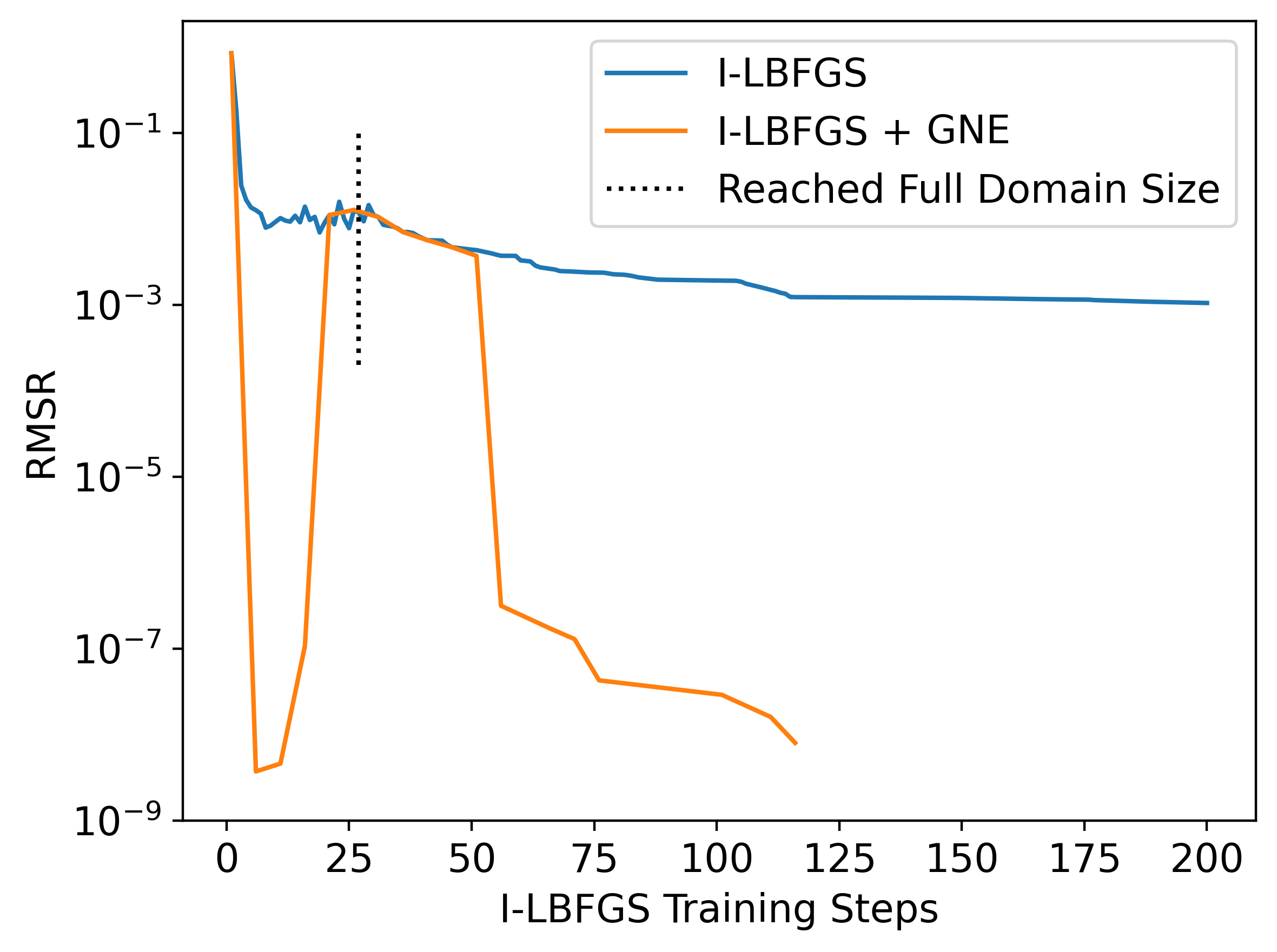}
    \caption{Learning curves of incremental L-BFGS and incremental L-BFGS + GNE. The incremental training reached the full domain size in 27 steps. The curves are not plotted against training time to show the overlap in RMSR values. The 200 L-BFGS steps took 152~s. The GNE took on average 50~s.
    }
    \label{fig:pure_advection_rmsr}
\end{figure}

In order to train the neural network we use the incremental training approach we used in Section \ref{sec:stiff_eq}. The neural network was first trained on the $[0,0.01]$ time domain. After a threshold RMSR of $10^{-3}$ was reached, the domain size was increased by 50\%. This process was repeated until the full time domain size of $[0,1.0]$ was reached. Then the training is continued till the desired RMSR value is reached. For this example it is possible to train the neural network without incremental training, but it was found to be significantly slower. Note that the PyTorch implementation of L-BFGS while training, occasionally encountered large gradients and resulted in $NaN$ values in the neural network parameters. When this happens the neural network is reverted back to the previous best parameters (lowest RMSR) and training was resumed.

We tried different number of output neurons $\{100, 200, 300, 400, 500, 600\}$ for the ELM, with and without incremental training. Directly training on the full domain always resulted in $NaN$ values of the parameters. With incremental training ELM initially learns well when domain size is small but later fails and network parameters become $NaN$ when the domain size is increased. A similar behavior can be seen in Fig. \ref{fig:pure_advection_rmsr} when GNE initially yields better RMSR, but as the domain size increases, it fails and overlaps the learning curve of pure L-BFGS. After reaching the full domain size, it takes further L-BFGS training for GNE to start improving the RMSR. The DNN we used in this example had 3 hidden layers with (32, 32, 400) neurons.

\begin{figure}[h]
    \centering
    \includegraphics[width = 0.32 \linewidth]{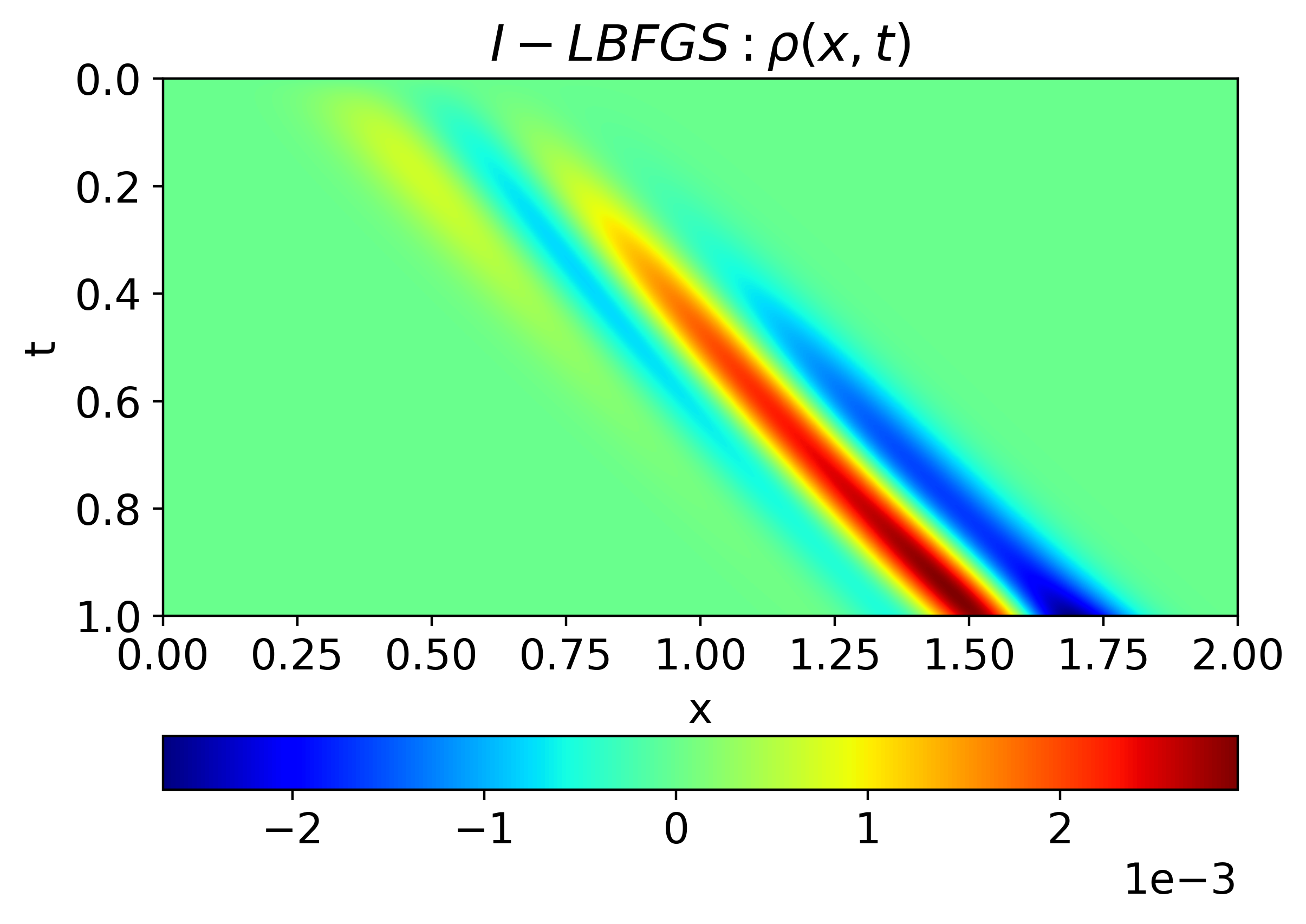}
    \includegraphics[width = 0.32 \linewidth]{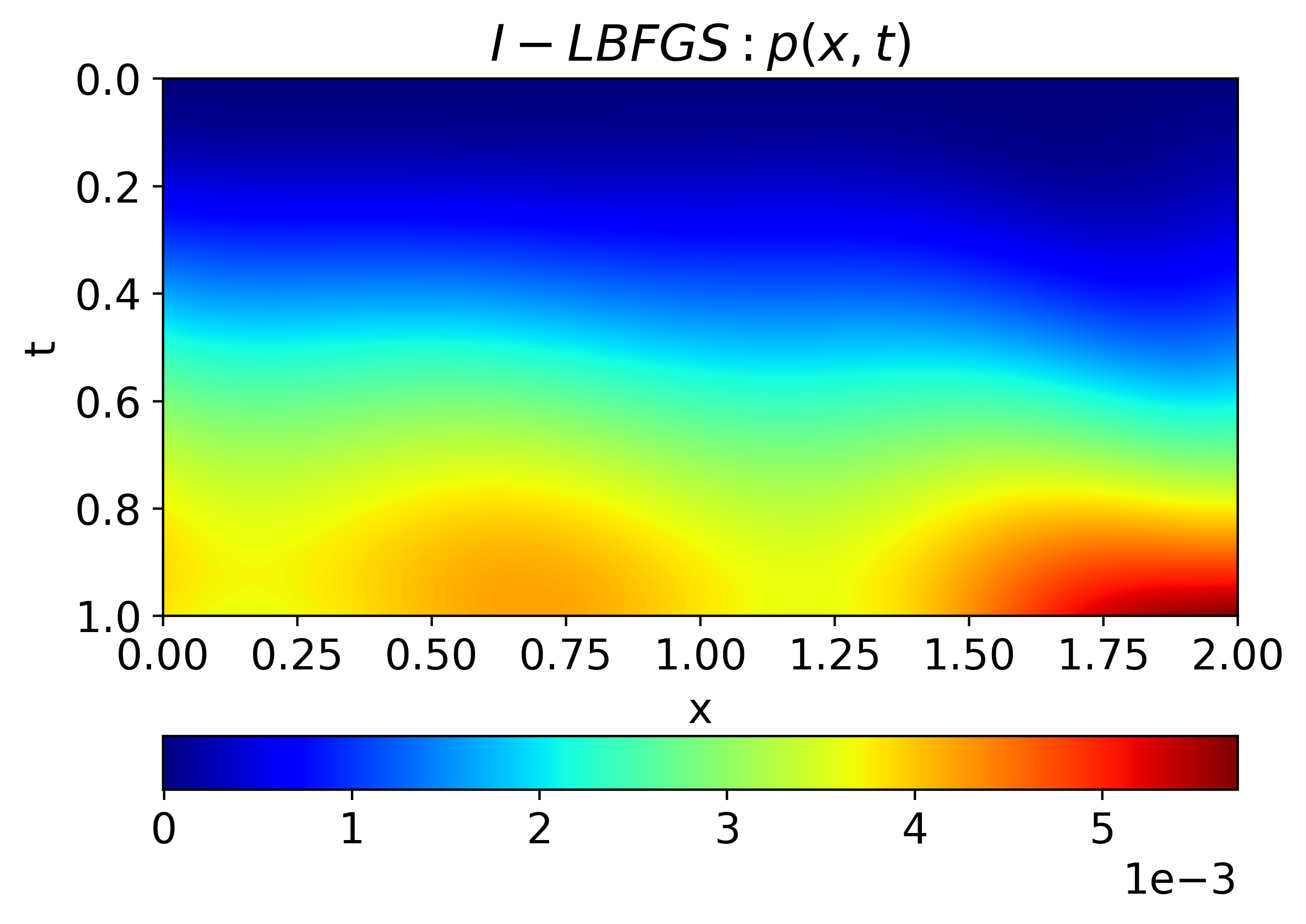}
    \includegraphics[width = 0.32 \linewidth]{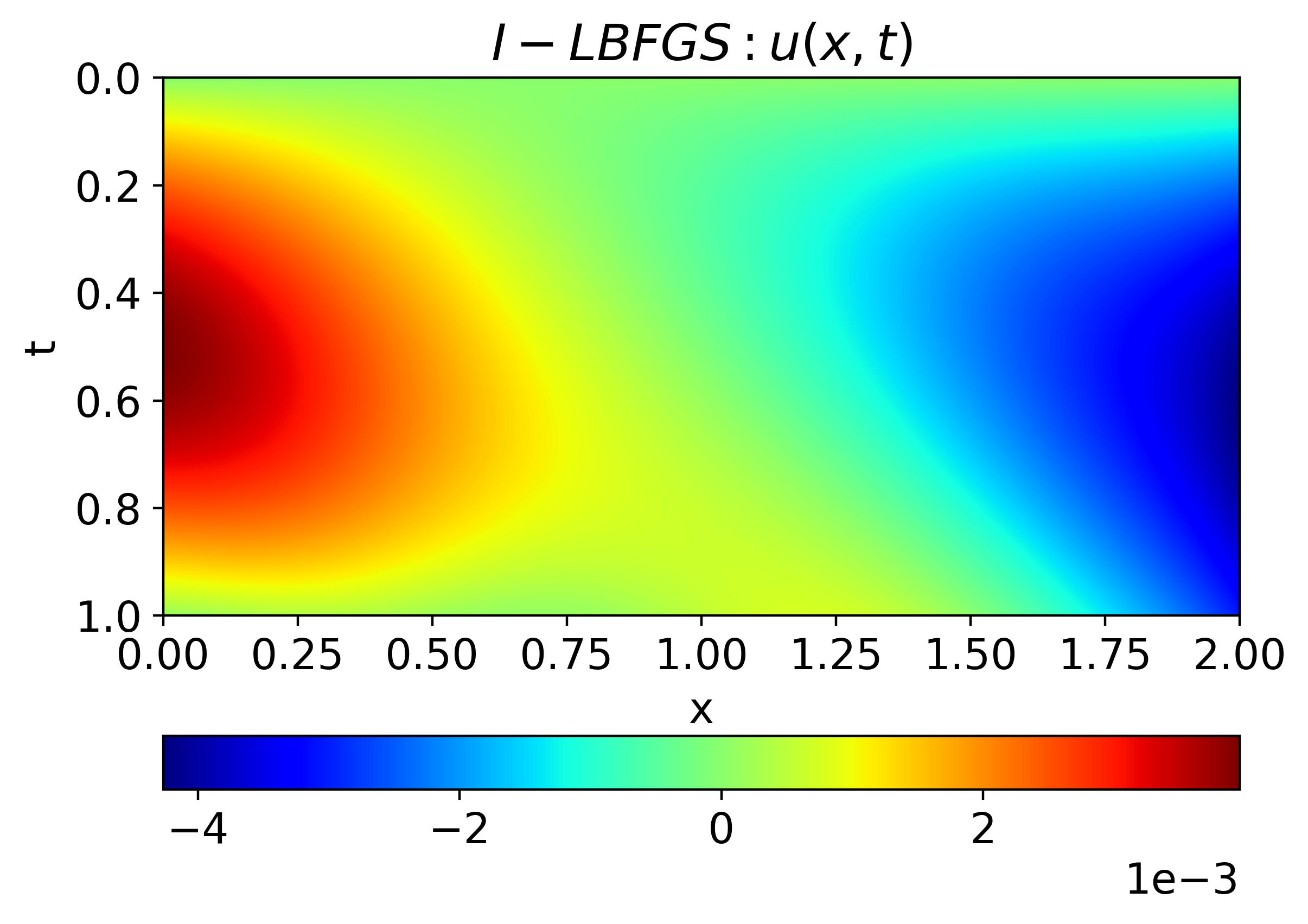}
    \includegraphics[width = 0.32 \linewidth]{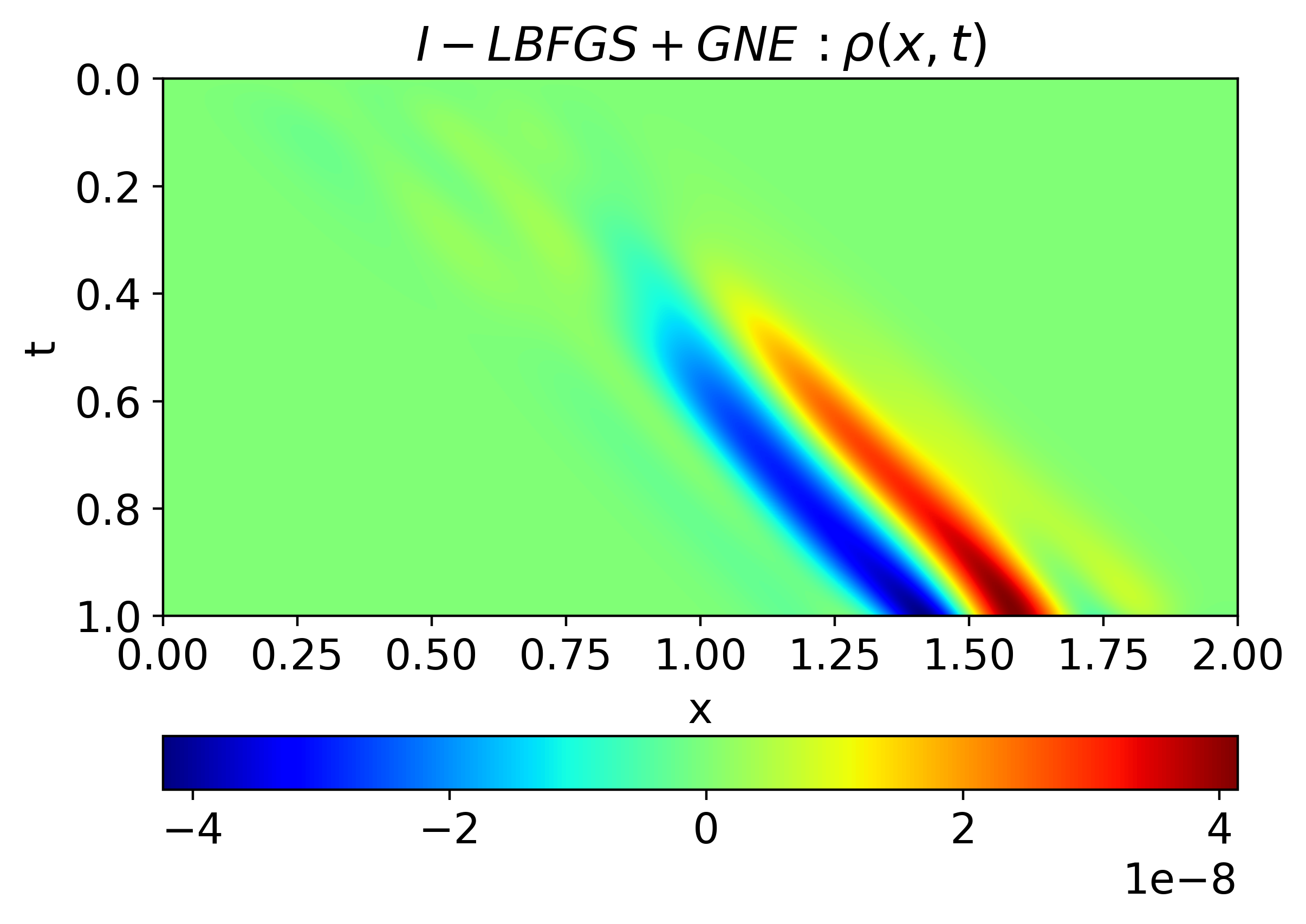}
    \includegraphics[width = 0.32 \linewidth]{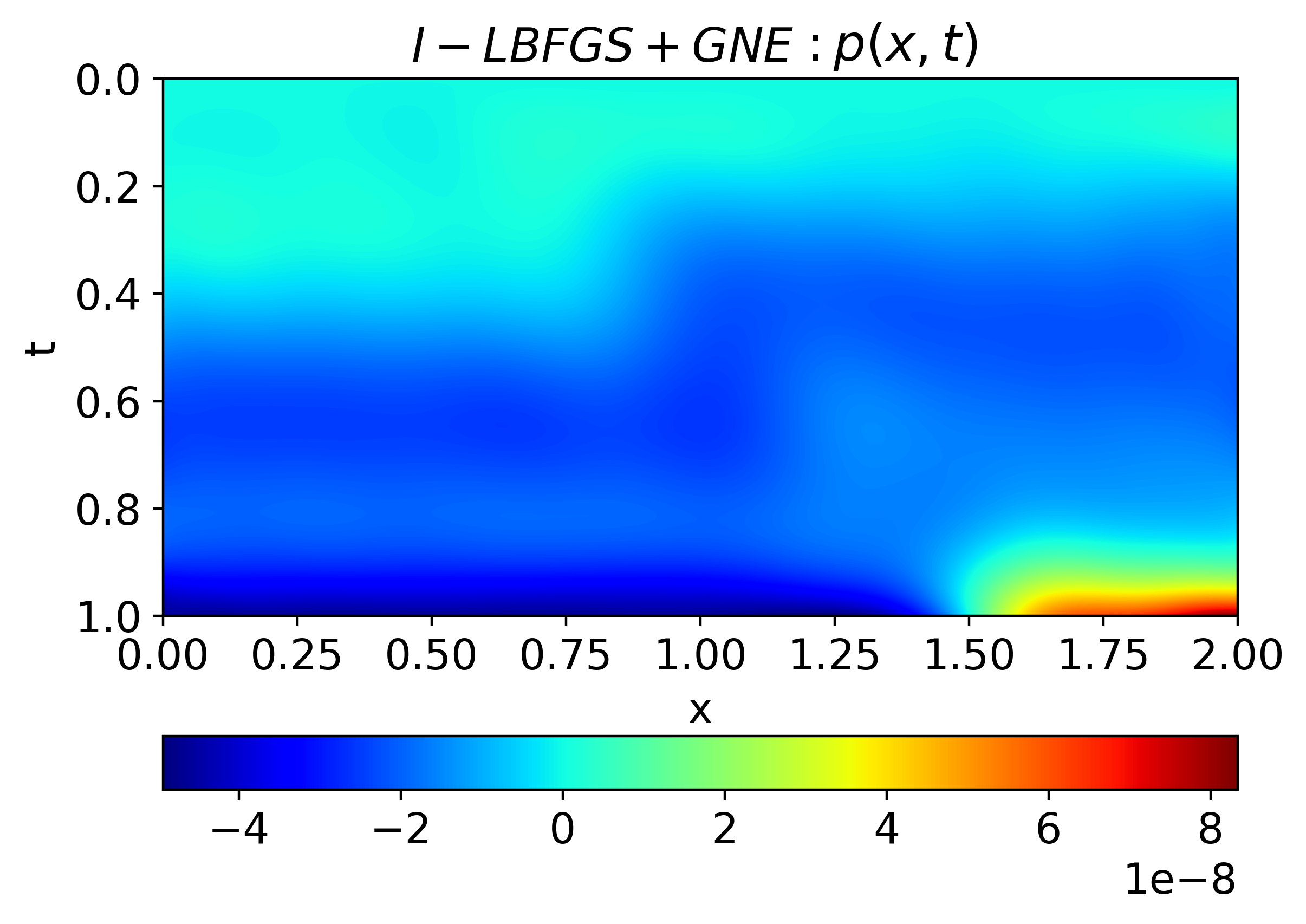}
    \includegraphics[width = 0.32 \linewidth]{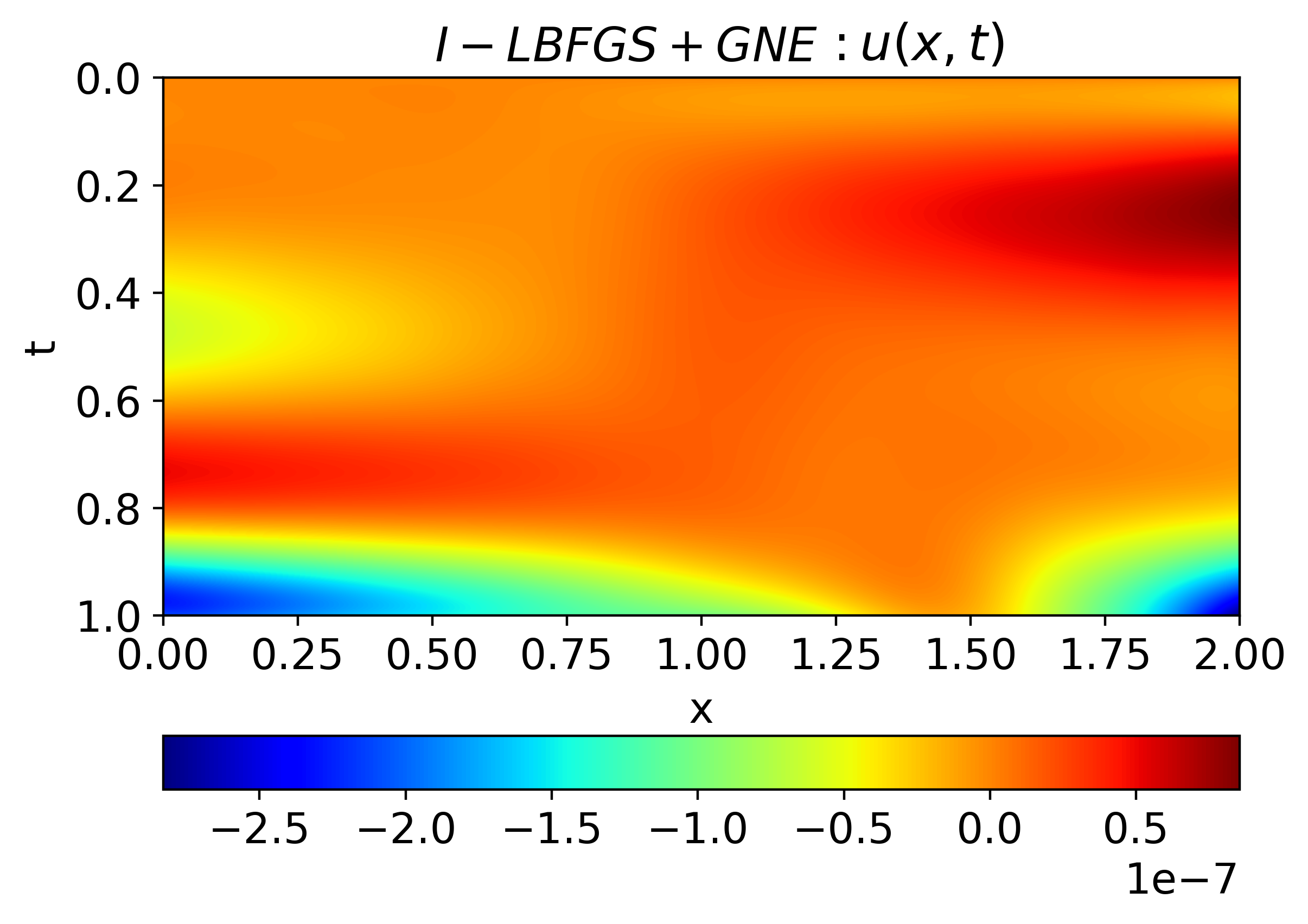}
    \caption{Errors in the solution of pure advection computed using \textbf{(top)} incremental L-BFGS and \textbf{(bottom)} incremental L-BFGS + GNE.
    }
    \label{fig:pure_advection_err}
\end{figure}

Fig. \ref{fig:pure_advection_err} shows the errors for the solutions of $u$, $p$ and $\rho$. The use of GNE has resulted in a 5 orders of magnitude improvement in the solution accuracy. However it is to be noted that similar to the case of the Burgers' equation in Sec. \ref{sec:burgers_eq}, the improvement with GNE decreases with sharper gradients, which in this case means smaller $\sigma$ in Eq. \ref{eq:rho_ini}. This results in difficulties in solving shock solutions to the Euler equations \cite{sod1978survey}. Even though crude solutions can be computed using PINN trained with Adam and L-BFGS \cite{mao2020physics, liu2022discontinuity, papadossolving}, GNE always fails due to exploding gradients as discussed in Sec. \ref{sec:burgers_eq}.

\FloatBarrier
\section{Discussion}
\label{sec:pinn_discussion}
In this work we proposed a novel extremization method for fast and accurate training of PINNs for solving (I)BVPs by combining SGD or L-BFGS method for training DNNs, with GNE. By combining the ideas from training DNNs and ELMs, our proposed method could retain the expressive power of DNNs while benefiting from the fine tuning ability of ELMs. Using TFC to exactly satisfy ICs and BCs, in Sec. \ref{sec:pinn_results} we showed the superiority of our proposed method compared to traditional training methods for DNNs and ELMs in solving BVPs of various ODEs, PDEs and coupled PDEs. In all cases the extremization method was found to produce solutions which are orders of magnitude better than DNNs trained with SGD or L-BFGS method alone. It also performed better than ELMs and depending on the complexity of the solution resulted in marginal to significant improvement in solution accuracy. The extremization method was shown to work in cases where ELM failed (Sec. \ref{sec:sin_eq}, Sec. \ref{sec:burgers_eq}, Sec. \ref{sec:pure_advection}) and further improve the accuracy of solutions computed using domain decomposition methods (Sec. \ref{sec:taylor_pde}). The utility of this method is yet to be investigated in the case of data-driven solutions and data-driven parameter discovery of PDEs. The simplicity of this algorithm and the ease of implementation suggest that this method can be readily used in other contexts where overfitting is not a concern or adequate measures can be taken to prevent overfitting. 

The extremization method is in general significantly faster than SGD or L-BFGS training methods since a lesser number of SGD or L-BFGS training iterations need to be performed before applying GNE, compared to using SGD or L-BFGS alone, to achieve similar or better solution accuracy. Our proposed method is slower than ELM since it requires additional SGD or L-BFGS training iterations before performing GNE, whereas ELM only uses GNE.

We also proposed a modification to the TFC framework called Reduced TFC (Sec. \ref{sec:rtfc}). Compared to TFC, Reduced TFC was shown in Sec. \ref{sec:heat_pde}, \ref{sec:3p1_nl_pde} \& \ref{sec:kovas} to provide up to 40x speed-up in computational time. In Sec. \ref{sec:tfc_remarks} we discussed the limitation of TFC in imposing boundary conditions on complex boundary geometries and how Reduced TFC can be used to solve this in principle. More work is needed in this direction to derive analytic expression for $\mathcal{G}(\boldsymbol{x})$ in Eq. \ref{eq:rtfc_g}. 

In Sec. \ref{sec:kovas} a significant amount of computational time was spend on GNE and we suggested a multi-GPU implementation of GNE to speed up this computation. Such an implementation is expected to provide an almost linear speed-up of GNE computation with the number of GPUs used. Another problem-specific method of speeding up the computation is to derive an analytic expression for the Jacobian. This can in principle be done by first deriving a TFC or Reduced TFC-constrained expression $f^c$ of Eq. \ref{eq:fcnn_eq}, plugging it into the differential equation to get the expression for the residual $\mathcal{R}$ and then using Eq. \ref{eq:jacobian} to compute the analytic expression for the Jacobian. The expression for the Jacobian will often be complicated and the use of a symbolic computation package like Mathematica is recommended. The analytic expression for the Jacobian will enable us to compute the Jacobian in a single forward pass with little memory overhead to evaluate the analytic expression.

In Sec. \ref{sec:burgers_eq} through the example of Burgers' equation we showed the failure of GNE when large gradients are present in the solution. GNE is not a unique choice in performing extremization, and other methods like the Levenberg-Marquardt \cite{more1978levenberg} and robust Gauss-Newton \cite{qin2018robust} need to be investigated. For large enough gradients, since all gradient-based methods fail, there is a need to explore non-gradient-based optimization methods \cite{larson2019derivative, aly2019derivative, chen2019zo} for PINN training. Even with non-gradient-based optimization, the residuals of the differential equations still contain derivatives which can lead to convergence failure of the optimization algorithm when large enough gradients or discontinuities are present. Some training algorithms have been proposed \cite{han2020derivative, lv2021hybrid} that partly solve this problem by using residuals of the differential equations while training the neural network. Such methods require careful treatment of the problem on a case-by-case basis. Another fundamental limitation of commonly used neural network architectures is that they can only represent continuous functions and approximating discontinuous solutions leads to exploding gradients during training. This issue is addressed in \cite{della2023discontinuous} where a neural network architecture with trainable discontinuous activation functions is proposed, which can represent arbitrary discontinuous functions in solutions. Even though such a network can represent an arbitrary discontinuous function, new training methods need to be developed to train it in the context of PINNs.

\section{Software Availability}
The Python source code for the methods introduced in this work is available from \cite{pinngithub}.

\appendix

\section{Least Squares}
\label{app:lstsq}
\begin{figure}[h]
    \centering
    \includegraphics[width = 0.32 \linewidth]{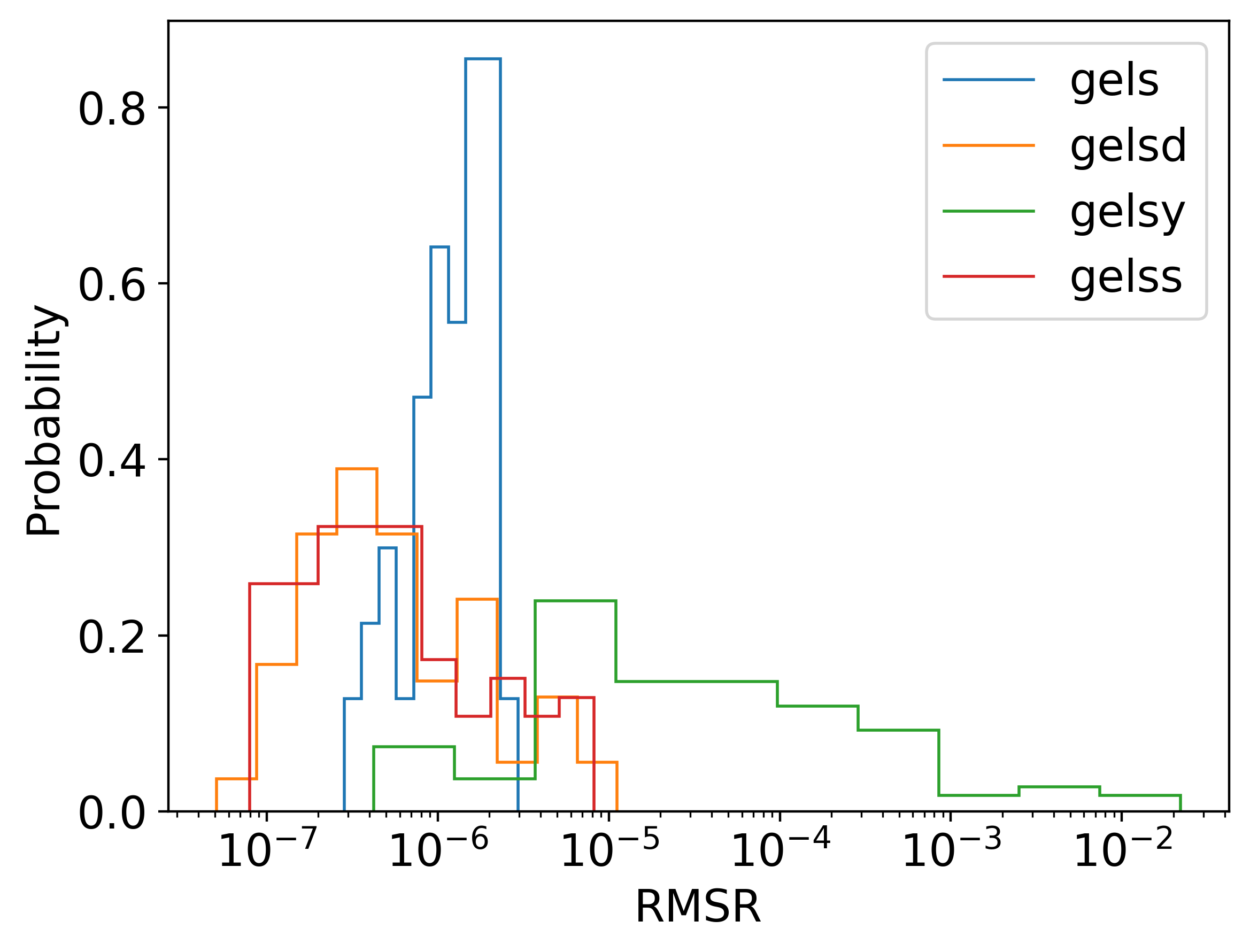}
    \includegraphics[width = 0.32 \linewidth]{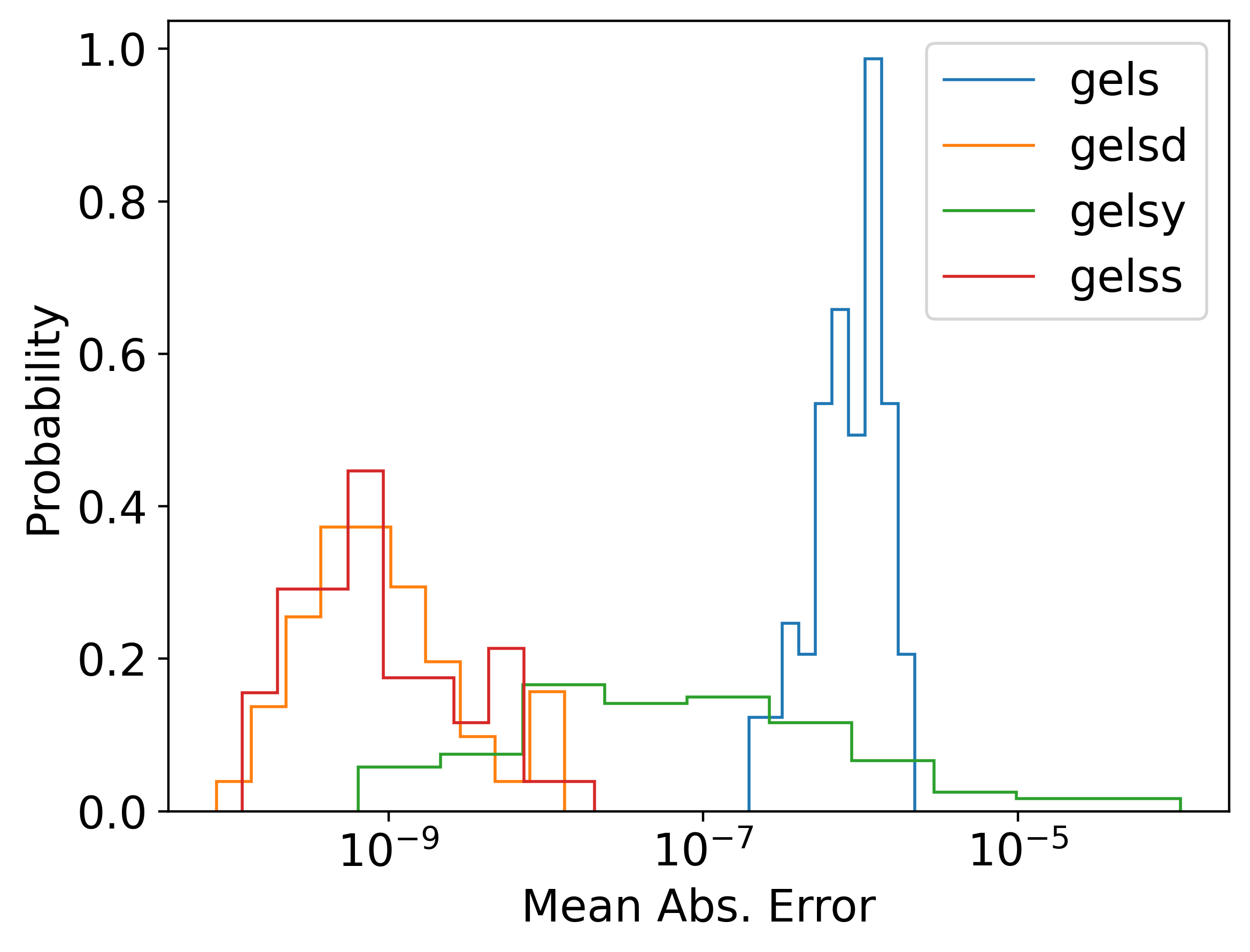}
    \includegraphics[width = 0.32 \linewidth]{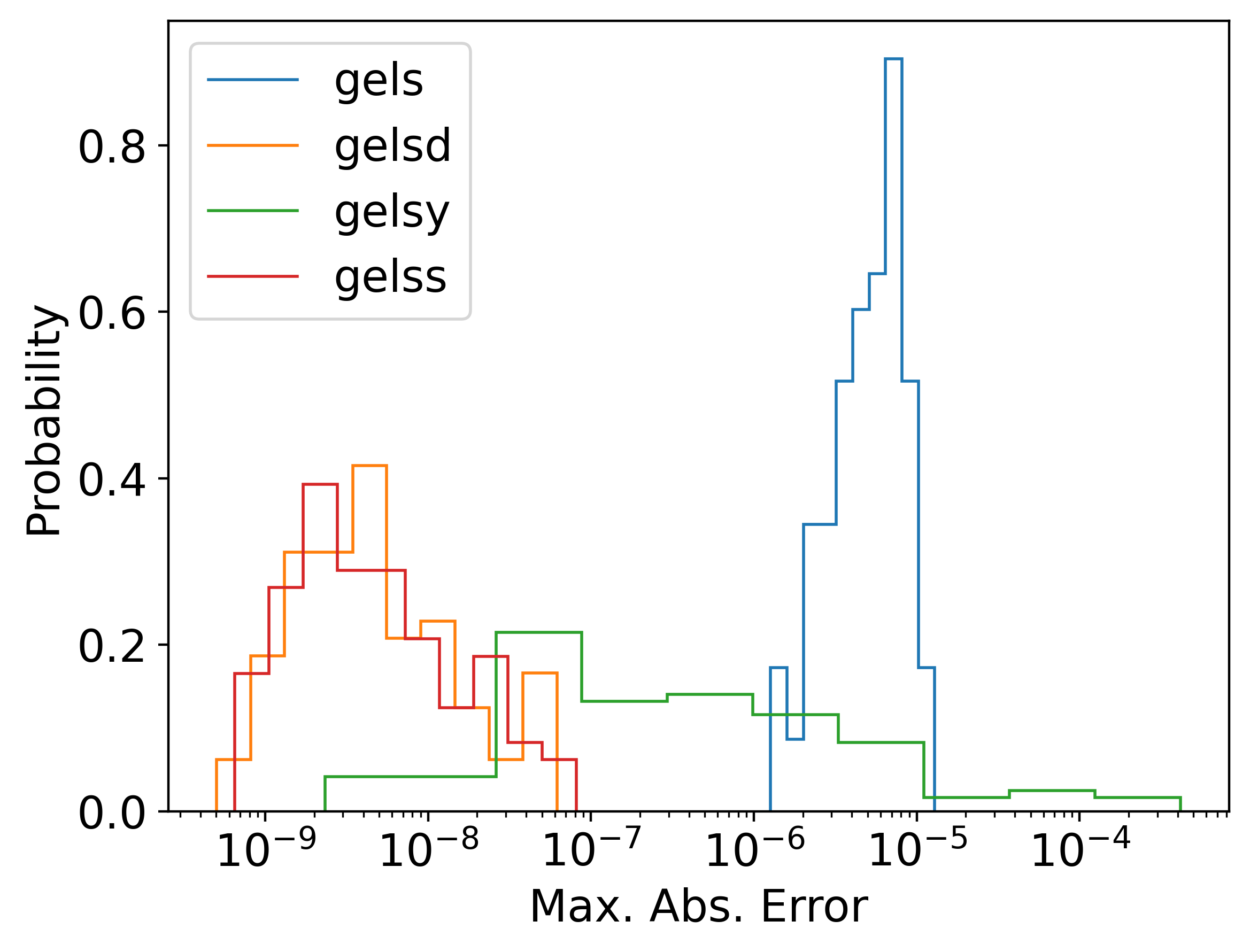}
    \includegraphics[width = 0.32 \linewidth]{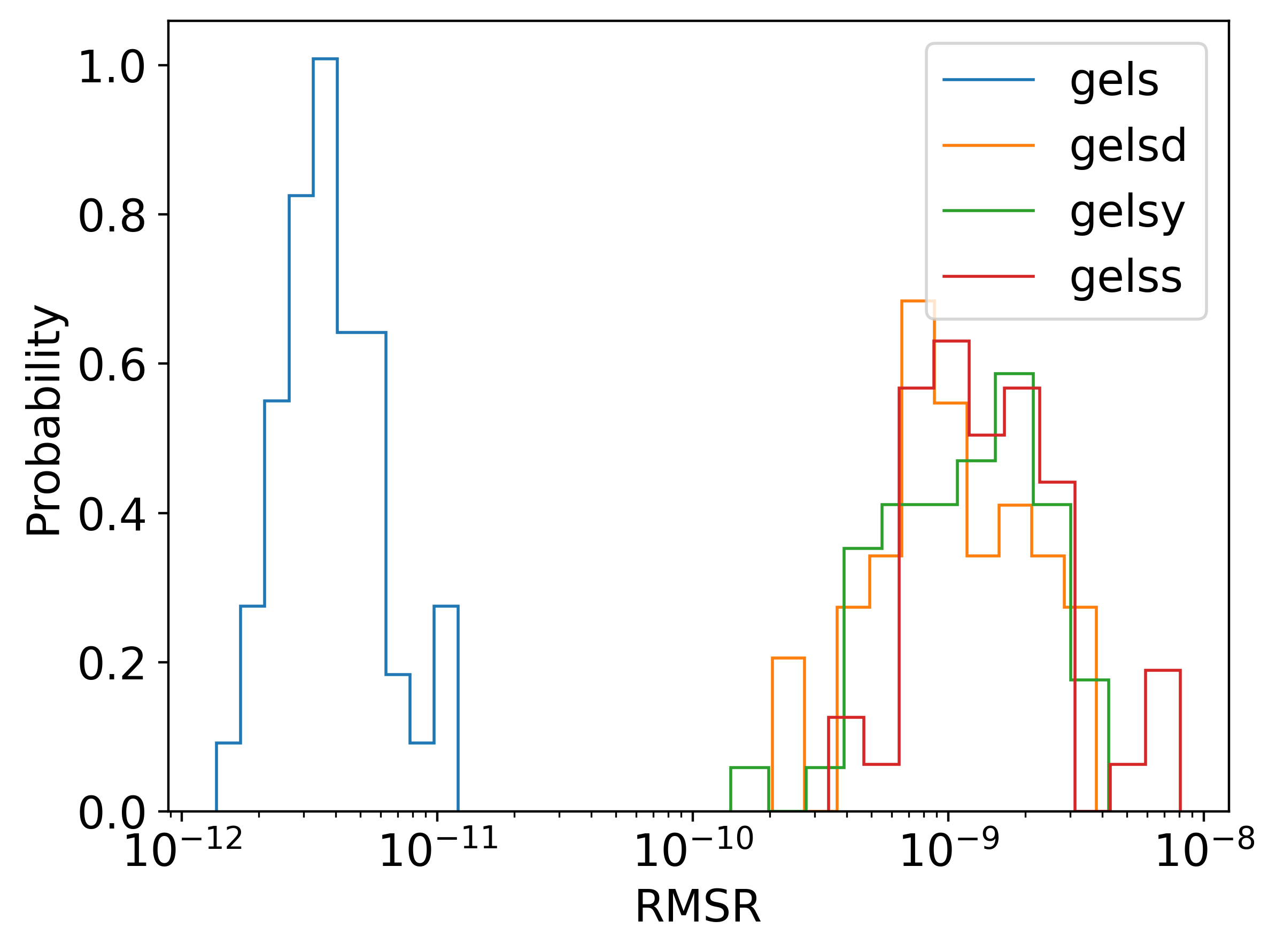}
    \includegraphics[width = 0.32 \linewidth]{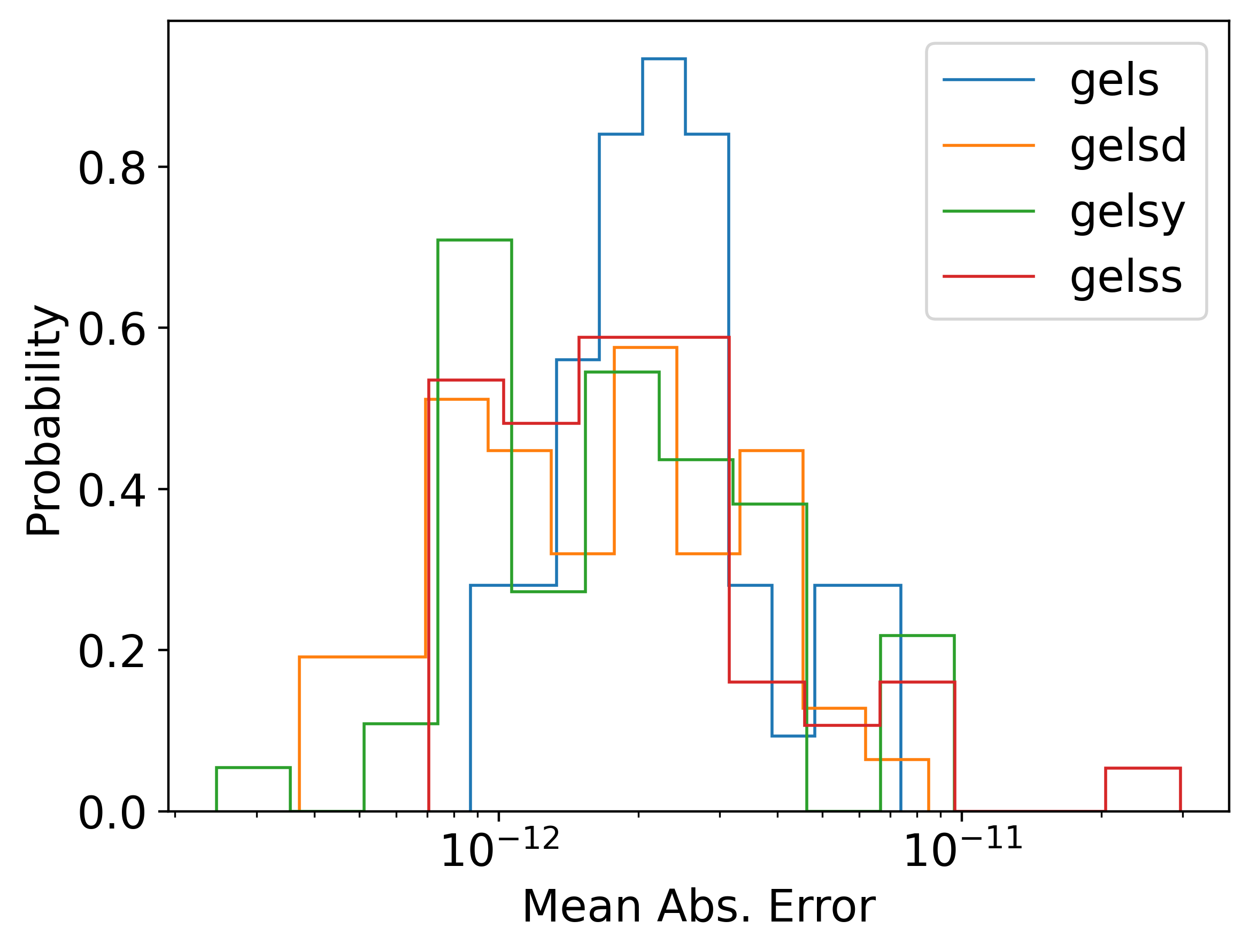}
    \includegraphics[width = 0.32 \linewidth]{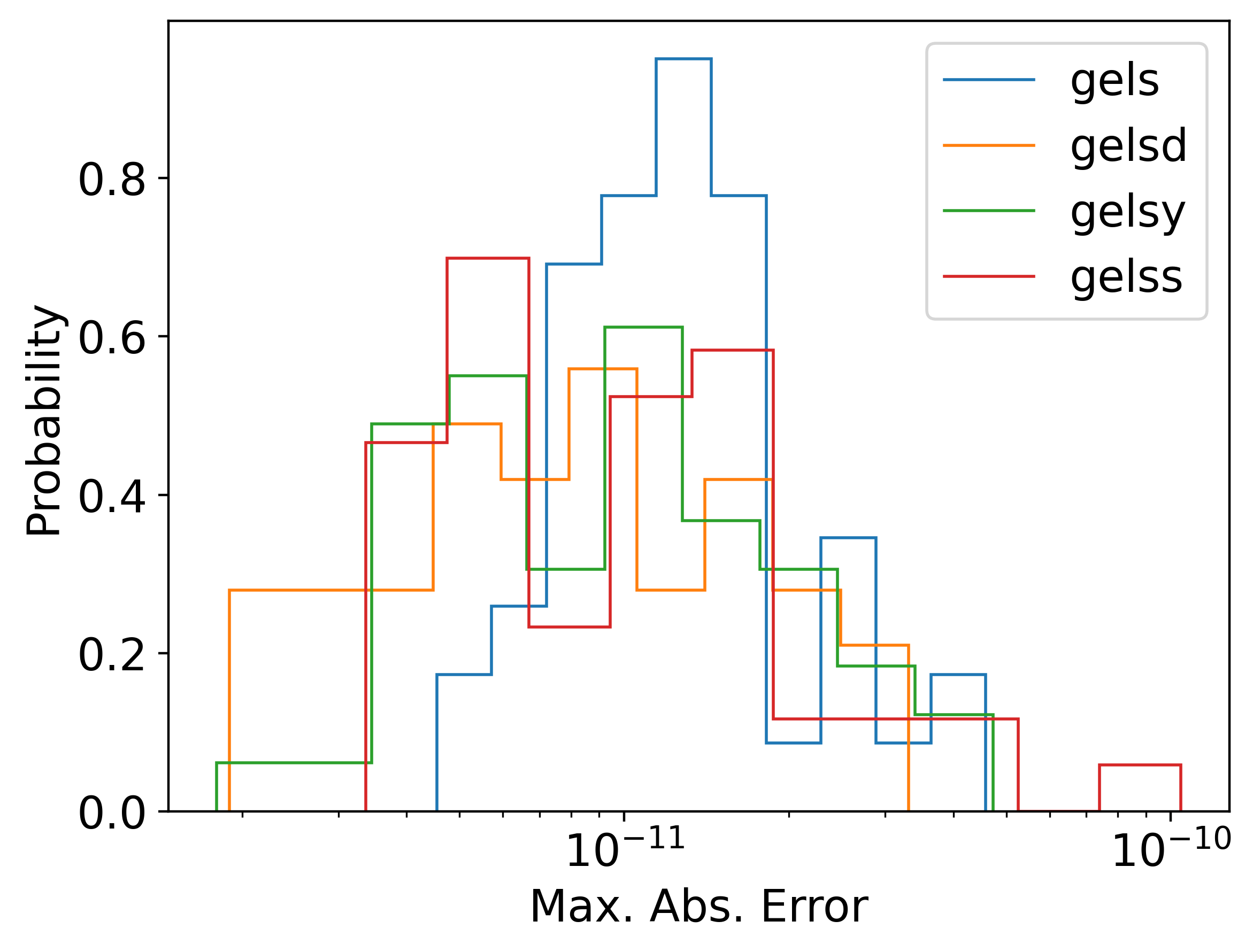}
    \caption{Statistics of the solution to the non-linear PDE given by Eq. \ref{eq:nl_pde_1} computed by \textbf{(Top)} ELM and \textbf{(bottom)} L-BFGS + GNE using different LLS algorithms provided in the LAPACK library \cite{lapack}. See Tab. \ref{tab:gels} for details on the algorithms.
    }
    \label{fig:lstsq_distrib}
\end{figure}

\FloatBarrier
\begin{table}[h]
\begin{center}
\begin{tabular}{ |l|l| }
\hline
 Algorithm Name & Operation \\
 \hline
 gels & solve LLS using QR or LQ factorization \\
 gelsd & solve LLS using divide-and-conquer SVD \\
 gelsy & solve LLS using complete orthogonal factorization \\
 gelss & solve LLS using SVD \\
 \hline
\end{tabular}
\end{center}
\caption{LAPACK \cite{lapack} algorithm names and the specific operation it performs.}
\label{tab:gels}
\end{table}
In this section we look at the effect of specific LLS algorithm on the solution accuracy achieved from GNE. We use the non-linear equation (Eq. \ref{eq:nl_pde_1}) in Sec. \ref{sec:l_v_nl} as a case study. We use the same neural network architecture and sample size as in Sec. \ref{sec:l_v_nl}. ELM was trained with GNE. L-BFGS + GNE was trained with 3 steps of L-BFGS and then used GNE. The training process was repeated 100 times for each network architecture with each LLS algorithm to generate the statistics shown in Fig. \ref{fig:lstsq_distrib}.

In Fig. \ref{fig:lstsq_distrib}, it can be seen that \textit{gelsd} and \textit{gelss} which use SVD decomposition (Tab. \ref{tab:gels}) consistently perform better than the other methods. An anomalous behavior can be seen when using \textit{gels} to perform GNE after L-BFGS. In this case, even though the RMSR value was better than for the other algorithms, the mean and maximum absolute errors were in line with the other algorithms. The explanation for these trends is beyond the scope of this work and requires further investigation. Throughout this work we use \textit{gelsd} to perform LLS.

\FloatBarrier
\section{Boundary Condition at Infinity}
\label{app:bc_at_infinity}
Consider a neural network with multiple inputs $\boldsymbol{x}$ and a bounded activation function in at least one of the layers, say, layer $k$. Then the outputs from the layer $k$, represented by $h^{(k)}_j(\boldsymbol{x})$ will be bounded:
\begin{equation}
    \left|h^{(k)}_j(\boldsymbol{x})\right| < C_1 \ \forall \boldsymbol{x}, \ 0<C_1<\infty.
\end{equation}
Now let $\Phi$ denote the output(s) of the neural network after the forward pass through the rest of the layers. Then the output(s) will also be bounded if there are no singularities in the activation functions in the rest of the layers:
\begin{equation}
    \left|\Phi\left(h^{(k)}_j(\boldsymbol{x})\right)\right| < C_2 \ \forall \boldsymbol{x}, \ 0<C_2<\infty.
\end{equation}

In order to apply the BC at infinity consider a function $g(x)$ which vanishes at infinity, $\lim_{|x| \to \infty} g(x) = 0$. Then the constrained expression for vanishing BC at infinity can be written as:
\begin{align}
\label{eq:inf_fc_1}
f^c &= \tilde{g}(\Phi) g(x) \ \text{or}\\
\label{eq:inf_fc_2}
f^c &= g\left(\tilde{g}(\Phi) + x\right),
\end{align}
where $\tilde{g}(x)$ represents any function without singularities in the domain $|\Phi| < C_2$. By considering the asymptotic behavior of $g$ and $\tilde{g}$ similar expressions can be derived for the case of unbounded activation functions. A BC with finite value at infinity can also be imposed by adding constants to the constrained expression Eq. \ref{eq:inf_fc_1} or Eq. \ref{eq:inf_fc_2}.

In Sec. \ref{sec:pure_advection} the constrained expression we used (Eq. \ref{eq:rho_ini}) can be derived using both Eq. \ref{eq:inf_fc_1} and Eq. \ref{eq:inf_fc_2}. In terms of Eq. \ref{eq:inf_fc_1}, $g(x)$ is the initial condition $\rho(x,0)$ and $\tilde{g}(x) = \exp(t \ x)$. The specific form of $\tilde{g}$ was chosen so that the density $\rho>=0$ and $\tilde{g} \to 1$ when $t \to 0$ to satisfy the initial condition.

\section{Constrained Expressions}
\label{app:tfc_expressions}
The explicit form of the TFC and Reduced TFC constrained expressions we use in this work are as follows:

\begin{itemize}
    \item \textbf{Sec. \ref{sec:sin_eq}}\\
    \begin{equation}
        f^c_y(t) = \mathcal{N}_y(t) - \mathcal{N}_y(0) + 1
    \end{equation}
    
    \item \textbf{Sec. \ref{sec:stiff_eq}}\\
    \begin{align}
        f^c_u(t) &= \mathcal{N}_u(t) - \mathcal{N}_u(0) \\
        f^c_v(t) &= \mathcal{N}_v(t) - \mathcal{N}_v(0) + 1
    \end{align}
    
    \item \textbf{Sec. \ref{sec:l_v_nl}}\\
    \begin{align}
        &f^c_{u \: 1}(x,y) = \mathcal{N}_u(x,y) - (1-x) \mathcal{N}_u(0, y) + x \mathcal{N}_u(1, y)\\
        &f^c_u(x,y) = f^c_{u \: 1}(x,y) - f^c_{u \: 1}(x,0) + y \left(2 \sin(\pi x) - \left. \partial_y f^c_{u \: 1}(x,y) \right|_{y = 1} \right)
    \end{align}
    
    \item \textbf{Sec. \ref{sec:burgers_eq}}\\
    \begin{align}
        &\label{eq:tfc_bruger_disc}
        f^c_{u \: 1}(x,t) = \mathcal{N}_u(x,t) - \mathcal{N}_u(x, 0) + \frac{c}{\alpha}-\frac{c}{\alpha} \tanh \left(\frac{c}{2 \nu} x\right)\\
        &\begin{aligned}
        f^c_u(x,y) = f^c_{u \: 1}(x,t) & + \frac{3+x}{6}\left(\frac{c}{\alpha}-\frac{c}{\alpha} \tanh \left(\frac{c}{2 \nu}(3-c t)\right) -  f^c_{u \: 1}(3,t)\right)\\ & + \frac{3-x}{6}\left(\frac{c}{\alpha}-\frac{c}{\alpha} \tanh \left(\frac{c}{2 \nu}(-3-c t)\right) -  f^c_{u \: 1}(-3,t)\right)
        \end{aligned}
    \end{align}
    
    \item \textbf{Sec. \ref{sec:heat_pde}}\\
    TFC constraints:
    \begin{align}
        &f^c_{u \: 1}(x,y,t) = \mathcal{N}_u(x,y,t) - \frac{1}{L}\left((L-x) \mathcal{N}_u(0, y,t) + x \mathcal{N}_u(L, y, t)\right)\\
        &f^c_{u \: 2}(x,y,t) = f^c_{u \: 1}(x,y,t) - \frac{1}{H} \left((H-y) f^c_{u \: 1}(x,0,t) + y f^c_{u \: 1}(x,H,t)\right)\\
        &f^c_{u}(x,y,t) = f^c_{u \: 2}(x,y,t) - f^c_{u \: 2}(x,y,0) + \sin \left(\frac{\pi x}{L}\right) \sin \left(\frac{\pi y}{H}\right)
    \end{align}
    Reduced TFC constraints:
    \begin{align}
        f^c_{u}(x,y,t) = \mathcal{N}_u(x,y,t) \, (x-L)\, x\, (y-H)\, y\, t + \sin \left(\frac{\pi x}{L}\right) \sin \left(\frac{\pi y}{H}\right) 
    \end{align}
    
    \item \textbf{Sec. \ref{sec:3p1_nl_pde}}\\
    TFC constraints:
    \begin{align}
        &f^c_{u\:1}(x, y, z, t)=\mathcal{N}_u(x,y,z,t) - \mathcal{N}_u(0,y,z,t)t^2 + \sin (2 \pi z) \\
        &f^c_{u\:2}(x, y, z, t)=f^c_{u\:1}(x, y, z, t) - f^c_{u\:1}(x, 0, z, t)t^2 + \sin (2 \pi z) \\
        &f^c_{u\:3}(x, y, z, t)=f^c_{u\:2}(x, y, z, t) - f^c_{u\:2}(x, y, 1, t) + \sin \left(x^2 y\right)+x y^{3 / 2} \\
        &\begin{aligned}
            f^c_{u}(x, y, z, t) = f^c_{u\:3}(x, y, z, t) &+ (1-t) \left(\sin \left(x^2 y\right)+x y^{3 / 2} z - f^c_{u\:3}(x, y, z, 0) \right)\\
            &+ t \left(\sin \left(x^2 y\right)+x y^{3 / 2} z+\sin (2 \pi z) -  f^c_{u\:3}(x, y, z, 0)\right)
        \end{aligned}
    \end{align}
    
    Reduced TFC constraints:
    \begin{align}
        f^c_{u}(x, y, z, t) = \mathcal{N}_u(x,y,z,t)\, x \, y \, (z-1)\, t\, (t-1) + x y^{3/2} z + \sin\left(x^2 y \right) + t^2 \sin\left(2 \pi z \right)
    \end{align}
    
    \item \textbf{Sec. \ref{sec:kovas}}\\
    TFC constraints:
    \begin{align}
        &\begin{aligned}
            f^c_{u \: 1}(x,y) = \mathcal{N}_u(x,y) 
            &+ \frac{x_2-x}{x_2-x_1}\left( 1-e^{\lambda x_1} \cos (2 \pi y) -\mathcal{N}_u(x_1,y) \right)\\
            & + \frac{x-x_1}{x_2-x_1}\left( 1-e^{\lambda x_2} \cos (2 \pi y) -\mathcal{N}_u(x_2,y) \right)
        \end{aligned}\\
        &\begin{aligned}
            f^c_{u}(x,y) = f^c_{u \: 1}(x,y)
            &+ \frac{y_2-y}{y_2-y_1}\left( 1-e^{\lambda x} \cos (2 \pi y_1) -f^c_{u \: 1}(x,y_1) \right)\\
            & + \frac{y-y_1}{y_2-y_1}\left( 1-e^{\lambda x} \cos (2 \pi y_2) -f^c_{u \: 1}(x,y_2)\right)
        \end{aligned}
    \end{align}
        \begin{align}
        &\begin{aligned}
            f^c_{v \: 1}(x,y) = \mathcal{N}_v(x,y) 
            &+ \frac{x_2-x}{x_2-x_1}\left( \frac{\lambda}{2 \pi} e^{\lambda x_1} \sin (2 \pi y) -\mathcal{N}_v(x_1,y) \right)\\
            & + \frac{x-x_1}{x_2-x_1}\left( \frac{\lambda}{2 \pi} e^{\lambda x_2} \sin (2 \pi y) -\mathcal{N}_v(x_2,y) \right)
        \end{aligned}\\
        &\begin{aligned}
            f^c_{v}(x,y) = f^c_{v \: 1}(x,y)
            &+ \frac{y_2-y}{y_2-y_1}\left( \frac{\lambda}{2 \pi} e^{\lambda x} \sin (2 \pi y_1) -f^c_{v \: 1}(x,y_1) \right)\\
            & + \frac{y-y_1}{y_2-y_1}\left( \frac{\lambda}{2 \pi} e^{\lambda x} \sin (2 \pi y_2) -f^c_{v \: 1}(x,y_2)\right)
        \end{aligned}\\
        &f^c_{p}(x,y) = \mathcal{N}_p(x,y) - \mathcal{N}_p(x_1,y_1) + p_0 - \frac{1}{2} e^{2 \lambda x_1}
    \end{align}
    
    Reduced TFC constraints:
     \begin{align}
        &\begin{aligned}
            f^c_{u \: 1}(x,y) = 
             \frac{x_2-x}{x_2-x_1}\left( 1-e^{\lambda x_1} \cos (2 \pi y) \right)
            + \frac{x-x_1}{x_2-x_1}\left( 1-e^{\lambda x_2} \cos (2 \pi y) \right)
        \end{aligned}\\
        &\begin{aligned}
            f^c_{u}(x,y) = \: &\mathcal{N}_u(x,y)\, (x-x_1) \, (x-x_2) \, (y-y_1) \, (y-y_2) + f^c_{u \: 1}(x,y) \\
            &+ \frac{y_2-y}{y_2-y_1}\left( 1-e^{\lambda x} \cos (2 \pi y_1) -f^c_{u \: 1}(x,y_1) \right)\\
            & + \frac{y-y_1}{y_2-y_1}\left( 1-e^{\lambda x} \cos (2 \pi y_2) -f^c_{u \: 1}(x,y_2)\right)
        \end{aligned}
    \end{align}
        \begin{align}
        &\begin{aligned}
            f^c_{v \: 1}(x,y) = 
            + \frac{x_2-x}{x_2-x_1}\left( \frac{\lambda}{2 \pi} e^{\lambda x_1} \sin (2 \pi y)\right)
             + \frac{x-x_1}{x_2-x_1}\left( \frac{\lambda}{2 \pi} e^{\lambda x_2} \sin (2 \pi y) \right)
        \end{aligned}\\
        &\begin{aligned}
            f^c_{v}(x,y) = \: &\mathcal{N}_v(x,y)\, (x-x_1) \, (x-x_2) \, (y-y_1) \, (y-y_2) + f^c_{v \: 1}(x,y) \\
            &+ \frac{y_2-y}{y_2-y_1}\left( \frac{\lambda}{2 \pi} e^{\lambda x} \sin (2 \pi y_1) -f^c_{v \: 1}(x,y_1) \right)\\
            & + \frac{y-y_1}{y_2-y_1}\left( \frac{\lambda}{2 \pi} e^{\lambda x} \sin (2 \pi y_2) -f^c_{v \: 1}(x,y_2)\right)
        \end{aligned}\\
        &f^c_{p}(x,y) = \mathcal{N}_p(x,y) \, (x-x1) \, (y-y1) + p_0 - \frac{1}{2} e^{2 \lambda x_1}
    \end{align}
    
    \item \textbf{Sec. \ref{sec:taylor_pde}}\\
    Reduced TFC constraints:
    \begin{align}
        &f^c_{u}(x,y,t) = \mathcal{N}_u(x,y,t)\, t + \sin x \cos y\\
        &f^c_{v}(x,y,t) = \mathcal{N}_v(x,y,t)\, t - \cos x \sin y\\
        &f^c_{p}(x,y,t) = \mathcal{N}_p(x,y,t)\, t + \frac{\rho}{4}(\cos 2 x+\sin 2 y)
    \end{align}
    
    \item \textbf{Sec. \ref{sec:pure_advection}}\\
    Reduced TFC constraints:
    \begin{align}
    f_p^c(x,t) &= \mathcal{N}_p(x,t)\, t + 1\\
    f_u^c(x,t) &= \mathcal{N}_u(x,t)\, t + 1\\
    f_{\rho}^c(x,t) &= \exp \left(\mathcal{N}_{\rho}(x,t)\, t -\frac{\left(x-\mu\right)^2}{2 \sigma^2}\right)
\end{align}
Note that the form of $f_{\rho}^c(x,t)$ is not based on TFC. It imposes boundary condition at infinity as shown in \ref{app:bc_at_infinity}.
\end{itemize}

\section*{Acknowledgements}
This research did not receive any specific grant from funding agencies in the public, commercial, or not-for-profit sectors. 

Abhiram Anand Thiruthummal thanks Dr. Abhishek Kumar for the valuable discussion on exact solutions of Navier-Stokes equations.

\bibliographystyle{unsrt}
\bibliography{main}
\end{document}